\documentclass{article}
\pdftrailerid{}
\usepackage{iclr2027_conference,times}
\usepackage{amsmath,amssymb,mathtools}
\usepackage{booktabs}
\usepackage{tabularx}
\usepackage{longtable}
\usepackage{multirow}
\usepackage{graphicx}
\usepackage{float}
\usepackage{placeins}
\usepackage{xcolor}
\usepackage{tikz}
\usepackage{pgfplots}
\pgfplotsset{compat=1.18}
\usepgfplotslibrary{groupplots,fillbetween}
\usetikzlibrary{positioning,calc,fit,arrows.meta}
\usepackage{algorithm}
\usepackage{algpseudocode}
\usepackage{hyperref}
\usepackage{url}
\hypersetup{colorlinks=true,linkcolor=blue,citecolor=blue,urlcolor=blue}

\title{Backpropagated Output Momentum:\\Relocating Optimizer History\\from Parameters to Task Space}

\newcommand{\bpm}{\textsc{BOM}}
\newcommand{\apm}{\textsc{APM}}
\newcommand{\adamw}{\textsc{AdamW}}

\definecolor{adamgray}{RGB}{45,45,45}
\definecolor{bpmblue}{RGB}{0,114,178}
\definecolor{lionorange}{RGB}{213,94,0}
\definecolor{galoregreen}{RGB}{0,158,115}
\definecolor{adamminisky}{RGB}{86,180,233}
\definecolor{muonpurple}{RGB}{168,87,127}
\definecolor{scaleolive}{RGB}{125,107,30}

\iclrfinalcopy
\author{Yuchen Li, Zongqi Fan, Nguyen H. Tran$^{*}$, Ken-Tye Yong$^{*}$ \\ University of Sydney \\ $^{*}$Corresponding authors}

\begin{document}
\maketitle

\begin{abstract}
Optimizer momentum is usually stored as a parameter-sized moving average of past gradients, which makes history costly and fixes each past signal in the coordinates in which it was computed. We introduce Backpropagated Output Momentum (\bpm{}), which instead stores a compact moving average of prediction errors at the model output and reprojects that history through the current network at every step. A batch-level analysis characterizes the information retained and omitted by this relocation, while the implementation preserves the current supervised gradient and can replace the first-moment component of several adaptive optimizers. As a plug-in for momentum-based optimizers, including ones that already compress their state, \bpm{} reduces parameter-shaped optimizer state by $49.7$--$99.8\%$ in three compositions and, averaged over three language backbones, paired step time by $4.0\%$. It also improves mean validation performance across language and vision fine-tuning, by $1.42$ points in the primary five-task comparison. Language and vision pretraining studies, together with matched mechanism controls, further test the construction across output spaces and model scales.
\end{abstract}

\section{Introduction}
AdamW is a reliable default for modern neural-network training because it combines a smoothed update direction with coordinate-wise adaptive scaling \citep{kingma2015adam,loshchilov2019decoupled}. The smoothing is carried by a \emph{first moment}: a dense exponential moving average (EMA) $m_t$ of the gradients up to optimizer step $t$, stored beside every trainable tensor, next to a second-moment EMA $v_t$ used for adaptive scaling. This design duplicates temporal state at every tensor: the two moments together occupy two parameter-shaped buffers, twice the memory of the parameters themselves and often a dominant part of the training footprint.

Memory is the visible cost; the second cost is structural. For the mini-batch loss at a step $\tau\le t$, $\mathcal L_\tau(\theta)=B^{-1}\sum_{i=1}^{B}\phi\bigl(f(\theta;x_{\tau,i}),y_{\tau,i}\bigr)$, the chain rule gives the batch gradient $g_\tau=B^{-1}\sum_{i=1}^{B}J_{\tau,i}^\top r_{\tau,i}$, where $J_{\tau,i}=\partial f(\theta;x_{\tau,i})/\partial\theta$ is the output Jacobian and $r_{\tau,i}=\partial\phi/\partial f$ the output residual of example $i$, with $\phi$ the per-example loss and $f$ the network output (the logits). This sum admits the exact decomposition
\begin{equation}
\label{eq:batch-decomposition}
    g_\tau=\bar J_\tau^\top s_\tau+c_\tau,\qquad
    s_\tau=\tfrac1B\textstyle\sum_i r_{\tau,i},\quad
    \bar J_\tau=\tfrac1B\textstyle\sum_i J_{\tau,i},\quad
    c_\tau=\tfrac1B\textstyle\sum_i(J_{\tau,i}-\bar J_\tau)^\top(r_{\tau,i}-s_\tau),
\end{equation}
where $s_\tau\in\mathbb R^{d_{\mathrm{out}}}$ is the batch-mean prediction error, $\bar J_\tau$ the batch-mean Jacobian, and $c_\tau$ the within-batch covariance between example-specific Jacobians and residuals. The first term maps the shared output error back to parameter space; the second is the example-specific part lost when errors are averaged before that mapping. AdamW's first moment stores an average of \emph{already-projected} gradients, $m_t=\sum_{\tau\le t} w_{t,\tau}(\bar J_\tau^\top s_\tau+c_\tau)$ with EMA weights $w_{t,\tau}$: each past task signal $s_\tau$ stays tied to the Jacobian $\bar J_\tau$ through which it was first projected, together with its covariance contribution. As training moves the parameters and replaces the samples, these historical projections drift away from the map the current model would apply to the current examples; Section~\ref{sec:reprojection} states this precisely and Appendix~\ref{app:drift-measurement} measures both sources of drift. Momentum factorization and state quantization \citep{park2025smmf,dettmers2022eightbit} reduce storage while keeping the accumulated history in parameter space.

We ask \emph{which parts of past gradients must be retained, and how compact can that history be?} We develop a batch-shared output history for supervised classification and language modeling, replacing dense first-order state with a compact residual EMA. In these tasks, output coordinates have stable meanings even while the model Jacobian changes. Our approach, \emph{Backpropagated Output Momentum} (\bpm{}), therefore keeps the history in task space and defers the projection: it stores one EMA of the task-space error, $q_t=\sum_{\tau\le t}w_{t,\tau}\,s_\tau$, and at each step transports the accumulated history through the current batch-mean Jacobian, $h_t=\bar J_t^\top q_t$. While AdamW averages projected gradients, \bpm{} averages task-space signals and projects them only once, through the current model on the current examples. The complete update retains the current supervised gradient (Section~\ref{sec:formal-update}), and Section~\ref{sec:reprojection} makes the comparison to AdamW exact at batch level.

The relocation changes both the update rule and the stored state: current reprojection omits the historical-projection term, and one task-space vector replaces every dense local first-moment tensor, reducing first-order state from $O(P)$ to $O(d_{\mathrm{out}})$. The decomposition identifies this structural difference, whereas its effect on optimization is evaluated empirically. Only the first-order history is relocated: the second moment stays parameter-local because elementwise squaring does not commute with the vector--Jacobian map, so no exact relocation of the same form exists (Section~\ref{sec:adaptive-scaling}).

\paragraph{Contributions.} Our contributions are summarized as follows.
\begin{enumerate}
    \item \emph{Introduce a composable first-moment replacement.} A compact task-space EMA replaces parameter-space momentum history, with current reprojection implemented in one backward pass. The construction composes with AdamW, Adam-mini and GaLore while retaining their adaptive-scaling rules.
    \item \emph{Explain and verify historical-projection drift.} An exact batch-level decomposition separates historical-projection drift from example-specific covariance. Probe measurements verify the presence of drift, and matched temporal-kernel and own-step controls support the contribution of current reprojection.
    \item \emph{Quantify fine-tuning gains and scale-dependent resource efficiency.} Five-task macros improve across three natural-language processing (NLP) backbones and three optimizer compositions. RoBERTa-base measurements show optimizer-state and peak-memory reductions in all three compositions. Pretraining measurements quantify training-checkpoint savings and, together with a calibrated cost model, characterize the observed efficiency crossover.
\end{enumerate}

\section{Related Work}
\paragraph{Adaptive, sign, and matrix-momentum optimization.}
Adam maintains exponential moving averages of gradients and squared gradients, and AdamW decouples weight decay from the adaptive update \citep{kingma2015adam,loshchilov2019decoupled}. Lion applies sign-based updates from a learned momentum rule \citep{chen2023symbolic}, Muon orthogonalizes matrix-valued momentum for hidden weights \citep{jordan2024muon}, and M+Adam combines additive and multiplicative updates \citep{liang2026madam}. These methods change how a parameter-space update is formed; \bpm{} instead changes where first-order history is stored and which Jacobian expresses it.

\paragraph{Memory-efficient optimizer states.}
Adafactor and CAME factor adaptive statistics \citep{shazeer2018adafactor,luo2023came}; GaLore projects gradients into low-rank subspaces \citep{zhao2024galore}; SMMF and a low-rank momentum approximation factorize momentum tensors \citep{park2025smmf,wang2026lorapre}. SCALE replaces the parameter-local first moment of the backbone matrices with gradient normalization, retaining parameter-space first-order momentum at the output layer and on one-dimensional parameters \citep{glentis2025minimalist}. Adam-mini reduces the number of distinct learning rates by sharing one second-moment scalar per block \citep{zhang2025adammini}. \bpm{} instead removes parameter-local first moments entirely: one $O(d_{\mathrm{out}})$ task-space EMA, stored before backpropagation and reprojected through the current full-network Jacobian, replaces them, while local second-moment scaling is kept.

\paragraph{Transported momentum: manifold and natural-gradient methods.}
Riemannian optimization accounts for the geometry of parameter-space search directions \citep{absil2008manifolds,bonnabel2013riemannian}. Riemannian momentum variants transport accumulated directions between tangent spaces \citep{becigneul2019riemannian}, while natural-gradient and Kronecker-factored approximate-curvature (K-FAC) methods precondition gradients using the Fisher information metric or its approximation \citep{amari1998natural,martens2015kfac}. In these momentum variants, the retained history remains parameter-dimensional. Closest in spirit is SPRING \citep{goldshlager2024spring}, which refreshes a carried stochastic-reconfiguration solution by projecting it onto the equations sampled at the current step, so that history is re-expressed through the current Jacobian; its carried state is parameter-shaped and the refresh is a per-sample regularized least-squares projection. \bpm{} keeps the same ordering with none of that machinery: one $O(d_{\mathrm{out}})$ task-space EMA enters through the loss, and a single backward pass performs the reprojection (Section~\ref{sec:substitution-rule}).

\paragraph{Objective-side corrections at the output.}
Logit adjustment addresses class imbalance through prior-based logit offsets, applied either after training or within the training loss \citep{menon2021logitadjustment}. \bpm{} uses a running average of output residuals as the detached coefficient of a logit-linear surrogate.

\section{Backpropagated Output Momentum}
\label{sec:method}

\paragraph{Notation.}
At step $t$, example $i\in\{1,\ldots,B\}$ of the batch has logits $z_{t,i}\in\mathbb R^{d_{\mathrm{out}}}$ and label $y_{t,i}$. The logit-loss gradient is the output residual $r_{t,i}$, with batch mean $s_t=B^{-1}\sum_i r_{t,i}$. For parameters $\theta\in\mathbb R^P$, let $J_{t,i}=\partial z_{t,i}/\partial\theta$ and $\bar J_t=B^{-1}\sum_iJ_{t,i}$. A subscript $\ell$ restricts a Jacobian or gradient to trainable tensor $\theta_\ell\in\mathbb R^{P_\ell}$, $\ell=1,\ldots,L$. Residuals and their EMA lie in task space $\mathbb R^{d_{\mathrm{out}}}$; projected gradients lie in parameter space. For historical steps $\tau\le t$, the EMA weights are $w_{t,\tau}=(1-\beta_1)\beta_1^{t-\tau}$, where $\beta_1\in[0,1)$ is the first-order EMA decay: AdamW applies it to $m_t$, and \bpm{} applies the same value to its task-space EMA (Section~\ref{sec:formal-update}). Norms $\|\cdot\|$ and $\|\cdot\|_{\mathrm{op}}$ are Euclidean and induced operator norms.

\subsection{Intuition: keep the history, refresh its projection}
\label{sec:intuition}
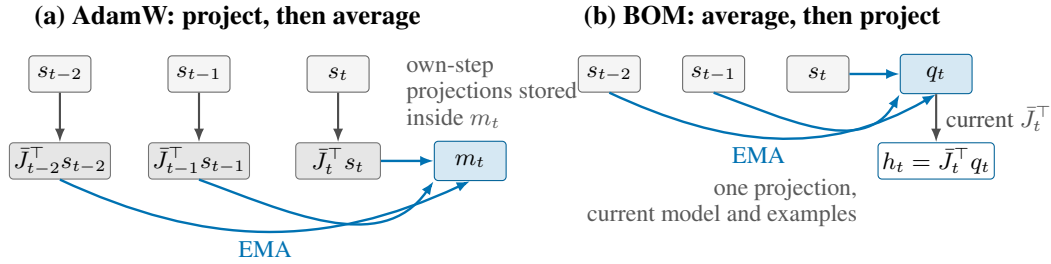
\begin{figure}[H]
\centering
\begin{tikzpicture}[font=\footnotesize, x=0.92cm, y=1cm,
    sig/.style={draw=black!60, rounded corners=1.5pt, fill=black!4, minimum height=0.48cm, minimum width=0.82cm, inner sep=1.5pt},
    stale/.style={draw=black!55, fill=black!10, rounded corners=1.5pt, minimum height=0.48cm, minimum width=1.12cm, inner sep=1.5pt},
    hist/.style={draw=bpmblue!85!black, fill=bpmblue!16, rounded corners=1.5pt, minimum height=0.52cm, minimum width=0.95cm, inner sep=1.5pt},
    proj/.style={-{Latex[length=1.8mm]}, line width=0.8pt, black!70},
    ema/.style={-{Latex[length=1.8mm]}, line width=0.9pt, bpmblue}]

\node[font=\bfseries, anchor=west] at (-0.5,2.75) {(a) AdamW: project, then average};
\node[sig] (sa1) at (0.0,2.0) {$s_{t-2}$};
\node[sig] (sa2) at (2.0,2.0) {$s_{t-1}$};
\node[sig] (sa3) at (4.0,2.0) {$s_{t}$};
\node[stale] (ga1) at (0.0,0.85) {$\bar J_{t-2}^{\top}s_{t-2}$};
\node[stale] (ga2) at (2.0,0.85) {$\bar J_{t-1}^{\top}s_{t-1}$};
\node[stale] (ga3) at (4.0,0.85) {$\bar J_{t}^{\top}s_{t}$};
\draw[proj] (sa1) -- (ga1);
\draw[proj] (sa2) -- (ga2);
\draw[proj] (sa3) -- (ga3);
\node[hist] (ma) at (5.9,0.85) {$m_t$};
\draw[ema] (ga1.south) to[out=-25,in=-155] node[pos=0.5, below, font=\footnotesize, text=bpmblue] {EMA} (ma.south);
\draw[ema] (ga2.south) to[out=-20,in=-125] (ma.south west);
\draw[ema] (ga3) -- (ma);
\node[anchor=west, text=black!65, font=\footnotesize, align=left] at (4.85,1.75) {own-step\\projections stored\\inside $m_t$};

\node[font=\bfseries, anchor=west] at (7.4,2.75) {(b) BOM: average, then project};
\node[sig] (sb1) at (7.9,2.0) {$s_{t-2}$};
\node[sig] (sb2) at (9.4,2.0) {$s_{t-1}$};
\node[sig] (sb3) at (10.9,2.0) {$s_{t}$};
\node[hist] (qb) at (12.6,2.0) {$q_t$};
\draw[ema] (sb1.south) to[out=-25,in=-150] node[pos=0.45, below, font=\footnotesize, text=bpmblue] {EMA} (qb.south);
\draw[ema] (sb2.south) to[out=-20,in=-120] (qb.south west);
\draw[ema] (sb3) -- (qb);
\node[stale, draw=bpmblue!85!black, fill=white] (hb) at (12.6,0.85) {$h_t=\bar J_t^{\top}q_t$};
\draw[proj] (qb) -- node[right, font=\footnotesize] {current $\bar J_t^{\top}$} (hb);
\node[anchor=east, text=black!65, font=\footnotesize, align=right] at (11.6,0.30) {one projection,\\current model and examples};
\end{tikzpicture}
\caption{Where first-order history lives. AdamW projects each $s_\tau$ through its own step's model and sampled examples before averaging, storing historical projections inside $m_t$; BOM averages the task signals first and applies one projection through the current model and current examples, refreshing the same weighted history at every step (covariance terms omitted; Section~\ref{sec:reprojection} is exact).}
\label{fig:history-transport}
\end{figure}

\bpm{} stores task errors and recomputes their parameter-space direction at each step. It maintains the residual EMA $q_t=\beta_1q_{t-1}+(1-\beta_1)s_t$ and applies the current projection $\bar J_t^\top q_t$. Figure~\ref{fig:history-transport} contrasts this order with AdamW; Section~\ref{sec:reprojection} gives their exact batch-level difference. The current vector--Jacobian product reaches every trainable tensor despite the compact stored history.

\paragraph{How small the stored history is.}
For binary softmax classification, every residual $r_{t,i}=\operatorname{softmax}(z_{t,i})-\operatorname{onehot}(y_{t,i})$ has zero coordinate sum, hence so do its batch mean $s_t$ and the EMA $q_t$. The stored history is therefore a single scalar: $q_t=(a_t,-a_t)$ with $a_t=q_{t,1}$. Writing $\bar J_{t,k}^\top\in\mathbb R^{P}$ for the gradient of the batch-mean $k$th logit (the $k$th row of $\bar J_t$), $\bar J_t^\top q_t=a_t(\bar J_{t,1}-\bar J_{t,2})^\top=a_t\nabla_\theta[B^{-1}\sum_i(z_{t,i,1}-z_{t,i,2})]$. Here $a_t$ is a running average of the batch-mean class-1 residual, i.e.\ of how far the mean predicted class-1 probability exceeded the class-1 label frequency in past batches; the coefficient thus carries past label-conditioned prediction errors, while the parameter-space direction is the gradient of the current batch-mean logit margin under the current model (bias correction rescales $a_t$ and leaves the direction unchanged). One scalar of first-order history reaches every trainable tensor, while the full current supervised gradient and the parameter-space second moment remain. This is a compact, batch-shared history, not per-example historical credit assignment or a fixed class-prior offset.

\paragraph{Assumption (stable output coordinates).}
The construction uses a fixed output dimension with stable coordinate meanings, as supplied by class labels in classification and vocabulary indices in language modeling. These shared coordinates allow residual history to accumulate across batches. Appendix~\ref{app:drift-measurement} measures how much residual magnitude survives batch averaging, and Section~\ref{sec:cross-domain} evaluates the vocabulary-sized construction in language pretraining.

\subsection{Formal update}
\label{sec:formal-update}
For cross-entropy (CE), the per-example residual is $r_{t,i}=\operatorname{softmax}(z_{t,i})-\operatorname{onehot}(y_{t,i})$, and $s_t=B^{-1}\sum_i r_{t,i}$ is its batch mean. \bpm{} stores its EMA, with the decay $\beta_1$ of the notation paragraph, and its bias-corrected form,
\begin{equation}
    q_t = \beta_1 q_{t-1} + (1-\beta_1)s_t,
    \qquad \hat q_t = \frac{q_t}{1-\beta_1^t}.
\end{equation}
To feed the task-space EMA through standard reverse-mode autodiff, \bpm{} forms a mixed scalar objective that blends the current cross-entropy loss $\mathcal L^{\mathrm{CE}}_t$ with a logit-linear surrogate whose coefficient is the detached EMA $\hat q_t$; the mixture weight $\lambda$ is fixed at $0.5$ in every reported \bpm{} run except the labelled sensitivity diagnostics:
\begin{equation}
\label{eq:mix-objective}
    \mathcal{L}^{\mathrm{mix}}_t
    =(1-\lambda)\mathcal{L}^{\mathrm{CE}}_t
    +\frac{\lambda}{B}\sum_{i=1}^{B}\langle z_{t,i},\hat q_t\rangle.
\end{equation}
Because $\hat q_t$ is treated as a constant with respect to the logits, one backward pass through $\mathcal L^{\mathrm{mix}}_t$ returns, for every trainable tensor, the per-step update numerator
\begin{equation}
\label{eq:mix-gradient}
    u_{\ell,t}=\nabla_{\theta_\ell}\mathcal{L}^{\mathrm{mix}}_t
    =(1-\lambda)g_{\ell,t}+\lambda\bar J_{\ell,t}^{\top}\hat q_t.
\end{equation}
The surrogate is a computational device: it yields $\bar J_{\ell,t}^\top\hat q_t$ through ordinary backpropagation, reprojecting the residual history through the current model and examples without materializing a Jacobian. The mixture and EMA coefficients of each experiment are listed in Appendix~\ref{app:reproducibility}.

\paragraph{What $u_{\ell,t}$ contains.}
Substituting the recursion for $q_t$ into Eq.~(\ref{eq:mix-gradient}) (with bias correction negligible) separates the current and historical components:
\begin{equation}
\label{eq:u-decomposition}
    u_{\ell,t}=\bigl[(1-\lambda)+\lambda(1-\beta_1)\bigr]\,\bar J_{\ell,t}^\top s_t
    +(1-\lambda)\,c_{\ell,t}
    +\lambda\beta_1\,\bar J_{\ell,t}^\top q_{t-1},
\end{equation}
whose three terms are the current batch-mean signal, the current covariance ($c_{\ell,t}$ is the $\ell$th block of $c_t$ in Eq.~(\ref{eq:batch-decomposition})), and the reprojected history. At the standard setting $\lambda=0.5$, $\beta_1=0.9$, the batch-mean component gives total weight $0.55$ to the current signal and $0.45$ to the past history $q_{t-1}$, whose tail decays at $0.9$, while the current covariance keeps weight $0.5$; AdamW instead weights the current full gradient $0.10$ and its accumulated history $0.90$. Section~\ref{sec:ablations} therefore runs two separate controls: an AdamW variant that matches only this $0.55/0.45$ mass split ($\beta_1=0.45$), and a parameter-space control that matches \bpm{}'s complete temporal kernel.

\subsection{Current-Jacobian reprojection and historical projection drift}
\label{sec:reprojection}
The contrast of Section~\ref{sec:intuition} can be stated exactly at batch level. Recall from Eq.~(\ref{eq:batch-decomposition}) that at any step $\tau$ the batch gradient decomposes as $g_\tau=\bar J_\tau^\top s_\tau+c_\tau$, with $s_\tau$, $\bar J_\tau$ and $c_\tau$ the batch-mean residual, batch-mean Jacobian and within-batch covariance of that step. For two examples with residuals $r$ and $-r$, the batch mean is zero, but $g=\tfrac12(J_1-J_2)^\top r$ can be nonzero. Averaging residuals therefore loses example-specific information even when the output coordinates are well defined. On a common parameter and batch trajectory with the same first-order decay, ignoring bias correction and recalling the EMA weights $w_{t,\tau}$, AdamW's first-moment construction and \bpm{}'s historical component are respectively
\begin{equation}
    m_t^{\mathrm{AdamW}}=\sum_{\tau=1}^{t}w_{t,\tau}\left(\bar J_\tau^\top s_\tau+c_\tau\right),
    \qquad
    h_t^{\mathrm{BOM}}=\bar J_t^\top\sum_{\tau=1}^{t}w_{t,\tau}s_\tau.
\end{equation}
Their difference isolates what \bpm{} declines to preserve in historical state,
\begin{equation}
\label{eq:jacobian-drift}
    m_t^{\mathrm{AdamW}}-h_t^{\mathrm{BOM}}
    =\sum_{\tau=1}^{t}w_{t,\tau}
    \left[(\bar J_\tau-\bar J_t)^\top s_\tau+c_\tau\right],
\end{equation}
whose norm is at most $\sum_{\tau=1}^{t}w_{t,\tau}\bigl(\|\bar J_\tau-\bar J_t\|_{\mathrm{op}}\|s_\tau\|+\|c_\tau\|\bigr)$.
Summing the first term inside the bracket of Eq.~(\ref{eq:jacobian-drift}) over $\tau$ gives $D_t:=\sum_{\tau=1}^{t}w_{t,\tau}(\bar J_\tau-\bar J_t)^\top s_\tau$, the \emph{historical-projection drift}: the part of AdamW's history that remains expressed through earlier Jacobians. Its norm is bounded by the operator-norm terms of that inequality, and $\bar J_\tau-\bar J_t$ changes for two reasons, parameter movement and sample replacement. To measure it, we maintain along one training trajectory both the bias-corrected EMA of own-step projections and \bpm{}'s current reprojection of the same history,
\begin{equation}
\label{eq:drift-telemetry}
    \hat M_t=\frac{1}{1-\beta_1^{t}}\sum_{\tau=1}^{t}w_{t,\tau}\,\bar J_\tau^\top s_\tau,
    \qquad
    h_t=\bar J_t^\top\hat q_t,
    \qquad
    \hat M_t-h_t=\frac{D_t}{1-\beta_1^{t}},
\end{equation}
and report the cosine $\cos(\hat M_t,h_t)$ and the relative drift $\lVert\hat M_t-h_t\rVert/\lVert\hat M_t\rVert$. Averaged over telemetry steps, the five evaluation seeds and the five RoBERTa tasks, the mean cosine is $0.81$ in the classification head but $0.52$ in the encoder, with relative drift $0.46$ and $0.79$ (Appendix~\ref{app:drift-measurement}); fixed probes separate the two sources (Table~\ref{tab:drift-decomposition}). These measurements establish a structural difference; Section~\ref{sec:ablations} compares the two projection rules by training quality. The covariance term $c_\tau$ is distinct from drift: \bpm{} omits it from task-space history but retains the full current gradient, including $c_t$, at weight $(1-\lambda)$. The decomposition thus identifies exactly which historical information the construction changes.

\subsection{Adaptive scaling and state complexity}
\label{sec:adaptive-scaling}
\bpm{} retains a parameter-space second moment of its actual numerator, where $\beta_2\in[0,1)$ is its decay,
\begin{equation}
    v_{\ell,t}=\beta_2 v_{\ell,t-1}+(1-\beta_2)u_{\ell,t}^{\odot2}.
\end{equation}
With the bias-corrected $\hat v_{\ell,t}=v_{\ell,t}/(1-\beta_2^{t})$, learning rate $\eta_t$, denominator constant $\epsilon$, and decoupled weight-decay coefficient $\omega_\ell$ (zero for bias and normalization tensors, Appendix~\ref{app:shared-protocol}), each trainable tensor follows
\begin{equation}
    \theta_{\ell,t+1}=\theta_{\ell,t}
    -\eta_t \frac{u_{\ell,t}}{\sqrt{\hat v_{\ell,t}}+\epsilon}
    -\eta_t\omega_\ell\theta_{\ell,t}.
\end{equation}
Eliminating the dense first moment $m_{\ell,t}$ halves the optimizer state: AdamW keeps two parameter-shaped buffers per tensor, whereas \bpm{} keeps one plus a single $d_{\mathrm{out}}$-vector, so the state ratio is approximately $\tfrac12$ whenever $d_{\mathrm{out}}\ll P$ and both buffers use the same storage precision. This ratio counts optimizer state only; peak allocated memory also includes parameters, gradients, activations, and any task-space workspace. Appendix~\ref{app:finetuning-resources} (Table~\ref{tab:memory-unified}) reports the measured optimizer state and peak allocated memory of the bf16 fine-tuning runs.

The asymmetry is structural: vector--Jacobian products are linear in the output signal, whereas elementwise squaring does not commute with that map, $(J^\top s)^{\odot2}\neq J^\top(s^{\odot2})$. Both optimizers use the same scaling form and coefficients, but AdamW accumulates $g_{\ell,t}^{\odot2}$ and \bpm{} accumulates $u_{\ell,t}^{\odot2}$, so their realized preconditioners can differ; the experiments evaluate the complete substitution under this matched rule. Algorithm~\ref{alg:bpm} in Appendix~\ref{app:per-dataset-results} gives the full step.

Appendix~\ref{app:convergence} gives a conditional descent and stationarity guarantee for this update, with explicit terms for history tracking, covariance, and other update perturbations.

\subsection{Substituting \bpm{} into a base optimizer}
\label{sec:substitution-rule}
The construction produces $u_{\ell,t}$ with the same shape as the ordinary gradient, allowing substitution into other first-moment update rules.

\paragraph{The substitution.} Let a base optimizer maintain a parameter-shaped first-moment accumulator and update
\begin{equation}
    m_{\ell,t}=\beta_1 m_{\ell,t-1}+(1-\beta_1)g_{\ell,t},
    \qquad
    \theta_{\ell,t+1}=\theta_{\ell,t}-\eta\,\Phi\!\left(\hat m_{\ell,t},\,\Sigma_{\ell,t}\right),
\end{equation}
where $\Sigma_{\ell,t}$ collects the base optimizer's remaining state and $\Phi(a,\Sigma_{\ell,t})$ is its update map: the function that takes the first-moment numerator $a$ together with that state and returns the parameter-shaped step. For AdamW, $\Sigma_{\ell,t}=\{v_{\ell,t}\}$ and $\Phi(a,\Sigma_{\ell,t})=a/(\sqrt{\hat v_{\ell,t}}+\epsilon)$; Adam-mini uses the same form with a block-shared $\hat v$; GaLore applies it in projected coordinates and reconstructs (Appendix~\ref{app:optimizer-implementation}). Here $\hat m_{\ell,t}$ includes the base optimizer's applicable first-moment bias correction; for AdamW it is $m_{\ell,t}/(1-\beta_1^t)$, and without such a correction it equals $m_{\ell,t}$. The substitution deletes $m_{\ell,t}$ and uses $\theta_{\ell,t+1}=\theta_{\ell,t}-\eta\,\Phi(u_{\ell,t},\Sigma_{\ell,t})$, with no additional local-first-moment correction applied to $u_{\ell,t}$. The task-space EMA and retained second moment keep their own bias corrections. Thus one backward pass through $\mathcal{L}^{\mathrm{mix}}_t$ supplies the numerator without a local first-moment accumulator.

\paragraph{Scope.} The formula above uses a dense representation. For a base with projected first moments, its original projection maps $u_{\ell,t}$ into the base's update coordinates, where it replaces the first-moment numerator; reconstruction is retained. Remaining adaptive state keeps its construction and bias correction but is updated from the new numerator, so its realized values can differ between arms. A base without a first-moment accumulator is outside this substitution's scope. Section~\ref{sec:composition} evaluates three compositions, while Appendix~\ref{app:out-rms} tests a separate output-only second-moment diagnostic.

\section{Experiments}
\label{sec:experiments}

\subsection{Setup and reporting}
We evaluate \bpm{} on RoBERTa-base \citep{liu2019roberta}, DeBERTa-v3-base \citep{he2023debertav3}, and Qwen3-1.7B \citep{qwen3technical} across five General Language Understanding Evaluation (GLUE) tasks \citep{wang2019glue}: Matthews correlation for CoLA, F1 for MRPC/QQP, and accuracy for RTE/SST-2. The macro weights the five task means equally; one point is $0.01$. Hyperparameter optimization (HPO) uses deterministic 90/10 partitions of each official training split and two HPO seeds, never the official validation split. Final runs start afresh, train five epochs on the full training split, and report the official-validation score at the fixed epoch-5 endpoint over five consecutive evaluation seeds disjoint from the HPO seeds. Unless stated otherwise, $\beta_1=0.9$, $\beta_2=0.999$, and \bpm{} uses AdamW-style scaling. Timing is excluded from selection and paired within task, seed, GPU model, and implementation stack (Appendix~\ref{app:reproducibility}).

\subsection{Natural-language processing (NLP) transfer across three backbones}
\label{sec:breadth}
The five-task macro improves by $1.42$ points on RoBERTa-base, $1.38$ on DeBERTa-v3-base and $1.35$ on Qwen3-1.7B; 13 of 15 task means improve (Table~\ref{tab:nlp_appendix_bpm}). All seeds are retained. One low DeBERTa AdamW SST-2 seed contributes $44.8\%$ of that backbone's margin, so the aggregate reflects both endpoint quality and stability (Table~\ref{tab:perseed-roberta}). Appendix~\ref{app:nlp-results} reports taskwise effects and uncertainty.

\begin{table}[H]
\centering
\caption{Primary epoch-5 NLP results. Scores average five tasks and five evaluation seeds; taskwise intervals are in Appendix~\ref{app:nlp-results}. $\Delta$Time is the mean paired bf16 step-time change of \bpm{} relative to AdamW; with FP32 master weights, RoBERTa-base changes by $-4.3\%$.}
\label{tab:primary-summary}
\begin{tabular}{lcccc}
\toprule
Backbone & AdamW & \bpm{} & $\Delta$Score & $\Delta$Time \\
\midrule
RoBERTa-base & 0.7962 & \textbf{0.8104} & +0.0142 & $-0.7\%$ \\
DeBERTa-v3-base & 0.8329 & \textbf{0.8467} & +0.0138 & $-4.8\%$ \\
Qwen3-1.7B & 0.8169 & \textbf{0.8304} & +0.0135 & $-6.4\%$ \\
\midrule
\textbf{Coverage} & \multicolumn{4}{c}{\textbf{3/3 backbones: higher mean score, lower mean step time}} \\
\bottomrule
\end{tabular}

\end{table}

On matched RoBERTa-base runs with FP32 master weights, optimizer state falls from $951.0$ to $475.5$ mebibytes (MiB) and peak allocated memory by $329$~MiB; in bf16, state falls from $475.5$ to $237.7$~MiB and peak memory by $232$~MiB (Table~\ref{tab:memory-unified}). In bf16, the paired mean step-time ratio is $0.993$ on RoBERTa-base, $0.952$ on DeBERTa-v3-base, and $0.936$ on Qwen3-1.7B, a mean reduction of $4.0\%$ (Table~\ref{tab:primary-summary}); with FP32 master weights, the RoBERTa-base ratio falls to $0.957$ ($-4.3\%$), as the removed buffer doubles in size.

\subsection{Fixed-backbone optimizer comparison}
\label{sec:balance}
Among six independently tuned fixed-backbone controls, \bpm{} has the highest five-task validation macro (Appendix~\ref{app:roberta-control-results}, Figure~\ref{fig:roberta-control-summary}). Its paired margin over AdamW is $+0.0142$ (descriptive 95\% confidence interval (CI) $[+0.0027,+0.0258]$, $p=0.027$); the other five margins range from $+0.0139$ to $+0.0513$ and all remain significant at $0.05$ after Holm correction. The appendix reports losses, state measurements, per-task results, and the retained degenerate SCALE seed; Figure~\ref{fig:roberta-control-curves} shows the training and validation trajectories.

\subsection{Transfer: substituting \bpm{} into other optimizers, and what it costs}
\label{sec:composition}
Table~\ref{tab:composition} substitutes \bpm{} into AdamW, block-adaptive Adam-mini, and low-rank GaLore. Their macros rise by $1.42$, $1.23$, and $1.14$ points, respectively, with 14 of 15 task means improving. Parameter-shaped state falls by $50.0\%$, $99.8\%$, and $49.7\%$, while paired mean step time falls by $0.7\%$ ($4.3\%$ with FP32 master weights), $2.0\%$, and $3.1\%$. Thus every composition improves all three mean estimates relative to its own base, including bases that already compress optimizer state.

\begin{table}[H]
\centering
\caption{Pairwise \bpm{} composition on RoBERTa-base: five-seed epoch-5 macros, parameter-shaped optimizer state (excluding the $O(d_{\mathrm{out}})$ vector), and step-time changes paired within each base's implementation stack. Each base keeps a parameter-local first-moment EMA, which \bpm{} replaces; Adam-mini shares $v$ within blocks, and GaLore keeps its moments in a low-rank subspace. Macro-margin 95\% CIs: AdamW $[+0.0027,+0.0258]$, Adam-mini $[-0.0203,+0.0449]$, GaLore $[-0.0078,+0.0307]$. All entries are measured in bf16; with FP32 master weights, the AdamW step-time change is $-4.3\%$. Details: Appendices~\ref{app:composition-results}, \ref{app:optimizer-implementation} and~\ref{app:measurement-protocol}.}
\label{tab:composition}
\setlength{\tabcolsep}{4pt}
\begin{tabular}{lcccccc}
\toprule
& \multicolumn{3}{c}{Five-task macro} & \multicolumn{2}{c}{Optimizer state (MiB)} & Step time \\
\cmidrule(lr){2-4}\cmidrule(lr){5-6}\cmidrule(lr){7-7}
Base optimizer & Base & With \bpm{} & $\Delta$ & Base & With \bpm{} & $\Delta$ \\
\midrule
AdamW & 0.7962 & \textbf{0.8104} & $+0.0142$ & 475.5 & 237.7 & $-0.7\%$ \\
Adam-mini & 0.7659 & \textbf{0.7782} & $+0.0123$ & 238.2 & 0.42 & $-2.0\%$ \\
GaLore & 0.7739 & \textbf{0.7854} & $+0.0114$ & 155.7 & 78.3 & $-3.1\%$ \\
\bottomrule
\end{tabular}
\end{table}

\subsection{Mechanism ablations}
\label{sec:ablations}
Table~\ref{tab:mechanism-main} examines buffer removal, objective mixing, temporal weighting, projection choice, and adaptive scaling; Table~\ref{tab:mechanism} and Appendices~\ref{app:roberta-ablations} and~\ref{app:diagnostic-protocols} give questions, constructions and reporting protocols.

\begin{table}[H]
\centering
\setlength{\tabcolsep}{4pt}
\caption{RoBERTa-base mechanism controls (epoch-5 macro, five seeds; $\Delta$ from \bpm{}'s $0.8104$). Details: Table~\ref{tab:mechanism}.}
\label{tab:mechanism-main}
\begin{tabular}{@{}lcc@{\hspace{12pt}}lcc@{}}
\toprule
Control & Macro & $\Delta$ & Control & Macro & $\Delta$ \\
\midrule
CE, no local $m$ & $0.7841$ & $-2.63$ & Own-step projection & $0.7807$ & $-2.97$ \\
Mixed + local $m$ & $0.7806$ & $-2.98$ & \bpm{} numerator, CE $v$ & $0.7892$ & $-2.12$ \\
AdamW, $\beta_1=0.45$ & $0.8008$ & $-0.96$ & Classifier-only history & $0.7910$ & $-1.94$ \\
Full-gradient kernel & $0.7977$ & $-1.28$ & $\beta_1=0$ (RTE acc.) & $0.5235$ & $4/5$ deg. \\
\apm{} anchor & $0.7934$ & $-1.70$ & Drift $\cos$ head/enc. & $0.81$/$0.52$ & -- \\
\bottomrule
\end{tabular}
\end{table}

\emph{Temporal and objective controls.} Under the matched HPO budget, removing the local first moment, mixing the objective, matching only the current/history mass, and matching the complete temporal kernel all finish below \bpm{} (kernel: 95\% CI $[+0.66,+1.90]$); none of these alone reproduces the margin.

\emph{Projection and scaling controls.} Drift is larger in the encoder than the head, and independently tuned own-step projection finishes $2.97$ points below current reprojection (95\% CI $[+0.14,+5.80]$). The independently tuned multiclass STL10 control agrees in direction on both backbones: current reprojection leads by $0.79$ points on ConvNeXt-Tiny (95\% CI $[-1.21,+2.79]$) and $0.63$ points on ViT-Tiny (95\% CI $[-0.09,+1.35]$; Appendix~\ref{app:cv-ownstep-results}). Removing history makes four of five RTE seeds degenerate (Appendix~\ref{app:coefficient-results}); numerator/second-moment cross-pairings and the output-only second-moment diagnostic further support retaining parameter-local adaptive scaling (Appendices~\ref{app:stale-control}, \ref{app:numerator-preconditioner}, and~\ref{app:out-rms}).

\emph{History scope and precision.} Full \bpm{} exceeds the classifier-only historical control on all five task means and by $1.94$ macro points under its protocol-selected configurations (95\% CI $[-0.85,+4.74]$), consistent with a contribution of full-network history in this setting (Appendix~\ref{app:classifier-history}). With independently tuned 32-bit floating-point parameter/state storage and bf16 compute, the positive RoBERTa macro margin persists at $+1.67$ points (95\% CI $[+0.73,+2.61]$; Appendix~\ref{app:fp32-replication}).

\subsection{Cross-domain validation: matched AdamW with and without \bpm{}}
\label{sec:cross-domain}
Table~\ref{tab:cross-domain} reports the matched AdamW--\bpm{} comparison on STL10 fine-tuning on ConvNeXt-Tiny and ViT-Tiny \citep{coates2011stl10,liu2022convnext,dosovitskiy2021vit}, ImageNet-1k pretraining on ResNet-50 \citep{deng2009imagenet,he2016resnet}, and Qwen3 language pretraining on Python code, FineWeb-Edu \citep{penedo2024fineweb}, and C4 \citep{raffel2020t5}. The independently tuned C4 comparisons span 55M, 110M, and 440M parameters with 3.0B, 5.0B, and 8.5B training tokens, and a 1.1B model is trained for 15.0B tokens with a preset peak learning rate (Figure~\ref{fig:c4-scale-losses}). Other pretraining settings use five consecutive-seed reporting runs. C4-440M uses three such runs and one HPO seed per screening evaluation, with the same coarse grid, local refinement rule, and 10-evaluation budget, internal-holdout loss criterion and selection rule as the other language-pretraining studies and a screening horizon of 1.5B tokens; the arms are screened separately and both select $4\times10^{-3}$. The 1.1B study also uses three runs and fixes the peak learning rate at $10^{-3}$ for both arms before reporting.

\begin{table}[H]
\centering
\caption{Matched AdamW--\bpm{} cross-domain results; each group states its metric and $\Delta$ is the paired \bpm{}$-$AdamW mean. Five consecutive-seed reporting runs except C4 440M and 1.1B ($n=3$ each); pretraining entries are mean $\pm$ sample sd. Configurations: Appendices~\ref{app:cv-params} and~\ref{app:pretraining-params}; paired intervals: Tables~\ref{tab:cv-stl10-check} and~\ref{tab:pretraining-results-appendix}.}
\label{tab:cross-domain}
\setlength{\tabcolsep}{4pt}
\begin{tabular}{lccc}
\toprule
Setting & AdamW & \bpm{} & $\Delta$ \\
\midrule
\multicolumn{4}{l}{\textit{Fine-tuning (vision), 40 epochs: validation accuracy $\uparrow$}} \\
STL10, ConvNeXt-Tiny & 0.9647 & $\mathbf{0.9769}$ & $+0.0122$ \\
STL10, ViT-Tiny & 0.9567 & $\mathbf{0.9669}$ & $+0.0102$ \\
\midrule
\multicolumn{4}{l}{\textit{Pretraining (vision), 90 epochs: epoch-90 validation top-1 $\uparrow$}} \\
ImageNet-1k, ResNet-50 & $0.7612\pm0.0005$ & $\mathbf{0.7646\pm0.0008}$ & $+0.0035$ \\
\midrule
\multicolumn{4}{l}{\textit{Pretraining (language): final evaluation loss $\downarrow$}} \\
Python code, Qwen3-55M, 3.0B tokens & $1.7353\pm0.0093$ & $\mathbf{1.7337\pm0.0106}$ & $-0.0016$ \\
FineWeb-Edu, Qwen3-55M, 3.0B tokens & $\mathbf{3.2931\pm0.0092}$ & $3.2948\pm0.0045$ & $+0.0017$ \\
C4, Qwen3-55M, 3.0B tokens & $3.5587\pm0.0105$ & $\mathbf{3.5506\pm0.0041}$ & $-0.0081$ \\
C4, Qwen3-110M, 5.0B tokens & $3.7859\pm0.0178$ & $\mathbf{3.7311\pm0.0034}$ & $-0.0547$ \\
C4, Qwen3-440M, 8.5B tokens & $3.1336\pm0.0004$ & $\mathbf{3.1292\pm0.0063}$ & $-0.0044$ \\
C4, Qwen3-1.1B, 15.0B tokens & $3.0106\pm0.0009$ & $\mathbf{3.0085\pm0.0008}$ & $-0.0021$ \\
\bottomrule
\end{tabular}
\end{table}

\bpm{} improves mean accuracy on both STL10 backbones and every ImageNet reporting run. On the three independently tuned C4 scales, it lowers mean final loss by $0.0081$ at 55M, $0.0547$ at 110M, and $0.0044$ at 440M. The 110M paired interval excludes zero and all five runs improve; at 55M and 440M, three of five and two of three pairs improve, respectively, with intervals that include zero. In the 1.1B study with a preset peak learning rate, all three pairs improve and the mean final-loss difference is $-0.0021$. The Python-code and FineWeb-Edu mean differences are small and have opposite signs. Appendix Figures~\ref{fig:pretraining-language-curves} and~\ref{fig:c4-scale-losses} report the loss trajectories.

\begin{figure}[H]
\centering
\begin{tikzpicture}
\begin{axis}[scale only axis,width=3.3cm,height=3.0cm,
xmode=log,log basis x=10,xmin=45,xmax=5000,
xtick={50,200,1000,4000},
xticklabels={50M,200M,1B,4B},
axis line style={black!60},tick style={black!60},
tick label style={font=\fontsize{8}{9}\selectfont},
label style={font=\fontsize{9}{10}\selectfont},
title style={font=\fontsize{9}{10}\selectfont\bfseries},
xlabel={Parameters (log scale)},
grid=major,grid style={gray!16},minor x tick num=0,
scaled ticks=false,clip=false,
legend style={font=\fontsize{8}{9}\selectfont,draw=none,fill=none,
inner sep=2pt,cells={anchor=west}},
name=timepanel,title={(a) Time ratio},
ymin=86,ymax=111,ytick={90,95,100,105,110},
ylabel={BOM / AdamW (\%)},legend pos=south west]
\addplot[black!45,densely dotted,forget plot] coordinates {(45,100) (5000,100)};
\addplot[color=bpmblue,very thick,mark=square*,mark size=2pt,
error bars/.cd,y dir=both,y explicit]
table[x=parameters_m,y expr=100*\thisrow{e2e_time_ratio},
y error expr=100*\thisrow{e2e_ratio_sd},col sep=comma]{data_c4_scale_resources.csv};
\addlegendentry{C4 end-to-end}
\addplot[color=lionorange,thick,mark=triangle,mark size=2.7pt,
mark options={solid,fill=white}]
table[x=parameters_m,y expr=100*\thisrow{step_ratio},col sep=comma]{data_h800_resource_ladder.csv};
\addlegendentry{H800 single step}
\addplot[color=black,thick,dashed]
table[x=parameters_m,y expr=100*\thisrow{fit_ratio},col sep=comma]{data_h800_cost_fit.csv};
\addlegendentry{Cost-model fit}
\node[font=\fontsize{8}{9}\selectfont,anchor=south west,text=black!65]
at (axis cs:48,100.3) {parity};
\end{axis}
\begin{axis}[scale only axis,width=3.3cm,height=3.0cm,
xmode=log,log basis x=10,xmin=45,xmax=5000,
xtick={50,200,1000,4000},
xticklabels={50M,200M,1B,4B},
axis line style={black!60},tick style={black!60},
tick label style={font=\fontsize{8}{9}\selectfont},
label style={font=\fontsize{9}{10}\selectfont},
title style={font=\fontsize{9}{10}\selectfont\bfseries},
xlabel={Parameters (log scale)},
grid=major,grid style={gray!16},minor x tick num=0,
scaled ticks=false,clip=false,
legend style={font=\fontsize{8}{9}\selectfont,draw=none,fill=none,
inner sep=2pt,cells={anchor=west}},
name=mempanel,at={(timepanel.east)},anchor=west,xshift=1.35cm,
title={(b) Peak memory},ymin=-2,ymax=17,ytick={-2,0,5,10,15},
ylabel={Memory saved (GiB)},legend pos=north west]
\addplot[black!45,densely dotted,forget plot] coordinates {(45,0) (5000,0)};
\addplot[color=bpmblue,very thick,mark=square*,mark size=2pt]
table[x=parameters_m,y expr=-\thisrow{memory_delta_gib},col sep=comma]{data_c4_scale_resources.csv};
\addlegendentry{C4 measured}
\addplot[color=lionorange,thick,mark=triangle,mark size=2.7pt,
mark options={solid,fill=white}]
table[x=parameters_m,y=peak_saved_gib,col sep=comma]{data_h800_resource_ladder.csv};
\addlegendentry{H800 measured}
\addplot[color=black!60,thick,dashed,domain=50:3954.407168,samples=150]
{4*(x*1000000-32768)/1073741824};
\addlegendentry{FP32 state ref.}
\node[font=\fontsize{8}{9}\selectfont,text=lionorange,anchor=south east,
yshift=3pt] at (axis cs:3954.407168,14.73119068) {14.73};
\node[font=\fontsize{8}{9}\selectfont,text=bpmblue,anchor=south east,
yshift=3pt] at (axis cs:1099.275264,5.34367609) {5.34};
\end{axis}
\begin{axis}[scale only axis,width=3.3cm,height=3.0cm,
xmode=log,log basis x=10,xmin=45,xmax=5000,
xtick={50,200,1000,4000},
xticklabels={50M,200M,1B,4B},
axis line style={black!60},tick style={black!60},
tick label style={font=\fontsize{8}{9}\selectfont},
label style={font=\fontsize{9}{10}\selectfont},
title style={font=\fontsize{9}{10}\selectfont\bfseries},
xlabel={Parameters (log scale)},
grid=major,grid style={gray!16},minor x tick num=0,
scaled ticks=false,clip=false,
legend style={font=\fontsize{8}{9}\selectfont,draw=none,fill=none,
inner sep=2pt,cells={anchor=west}},
at={(mempanel.east)},anchor=west,xshift=1.35cm,
title={(c) Checkpoint size},ymin=0,ymax=50,ytick={0,16,32,48},
ylabel={Checkpoint size (GiB)},legend pos=north west]
\addplot[color=adamgray,thick,dashed,domain=54.540288:3954.407168,samples=150]
{12*x*1000000/1073741824};
\addlegendentry{AdamW theory}
\addplot[color=bpmblue,thick,dashed,domain=54.540288:3954.407168,samples=150]
{8*x*1000000/1073741824};
\addlegendentry{BOM theory}
\addplot[only marks,color=adamgray,mark=o,mark size=2.7pt,mark options={fill=white,thick}]
table[x=parameters_m,y=adamw_checkpoint_gib,col sep=comma]{data_c4_checkpoint_sizes.csv};
\addlegendentry{AdamW data}
\addplot[only marks,color=bpmblue,mark=square*,mark size=2.3pt]
table[x=parameters_m,y=bom_checkpoint_gib,col sep=comma]{data_c4_checkpoint_sizes.csv};
\addlegendentry{BOM data}
\end{axis}
\end{tikzpicture}
\caption{\textbf{Pretraining resource efficiency across scales.} (a) C4 end-to-end time ratios (paired mean $\pm$ SD), H800 single-step ratios, and the calibrated cost model (Appendix~\ref{app:cost-model}). (b) Peak allocated-memory savings with the FP32 state reference $4(P-V)$ bytes. (c) Checkpoint sizes with FP32 predictions $12P$ (AdamW) and $8P$ (\bpm{}) bytes. Details: Table~\ref{tab:steptime-h800}, Figure~\ref{fig:c4-scale-resources}, Appendix~\ref{app:measurement-protocol}.}
\label{fig:pretraining-resources}
\end{figure}
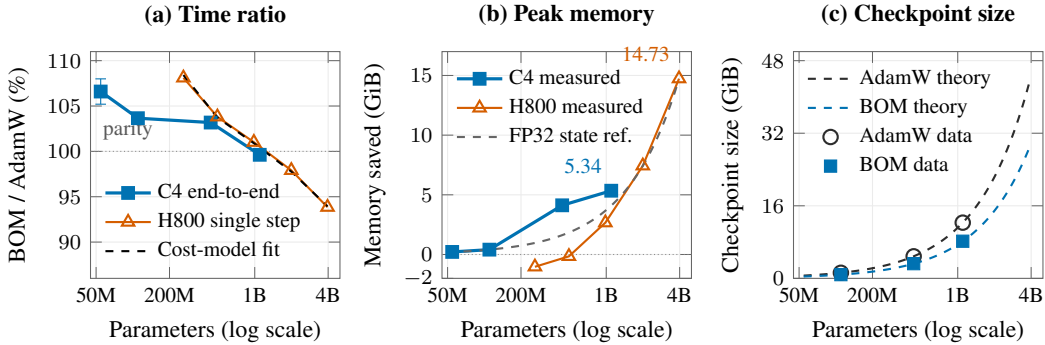

\paragraph{Pretraining resource scaling.}
\label{sec:steptime}
Figure~\ref{fig:pretraining-resources} compares the C4 training measurements with the H800 single-step benchmark on a common parameter axis. From 55M through 1.1B, the mean paired C4 duration ratios, measured over complete runs at 55M and 110M and over matched intervals at 440M and 1.1B, are $1.066$, $1.037$, $1.032$, and $0.996$. At 1.1B, \bpm{} saves $5.34$~GiB of peak allocation and reduces checkpoint size from $12.29$ to $8.19$~GiB. The measured checkpoint sizes follow the FP32 tensor-count predictions, $12P$ bytes for AdamW and $8P$ for \bpm{}. On H800, the step-time ratio crosses parity between 1B and 2B and reaches $0.938$ at 4B, where peak-memory savings are $14.73$~GiB. Calibrating the element-count model of Eq.~(\ref{eq:cost-model}) to the five H800 timing differences captures this shift from overhead to savings. Appendix~\ref{app:cost-model} gives the calibration; Appendix~\ref{app:measurement-protocol} specifies the distinct timing protocols and matched C4 intervals.

\section{Conclusion}
\bpm{} is a plug-in first-moment replacement for momentum-based optimizers: it relocates dense parameter-space history to an $O(d_{\mathrm{out}})$ task-space EMA and reprojects it through the current network. An exact batch-level decomposition separates historical-projection drift from example-specific covariance; probe measurements establish the drift, and matched mechanism controls support the contribution of current reprojection. In fine-tuning, every composition, including those on bases that already compress state, cuts parameter-shaped state by $49.7$--$99.8\%$ and lowers RoBERTa-base peak memory, while \bpm{} also improves the five-task macro on all three NLP backbones and compositions. For pretraining, a calibrated cost model, the H800 ladder and C4 measurements locate where step-time overhead turns into savings and quantify peak-memory and checkpoint reductions. Together, these results position task-space history as a plug-in that makes momentum-based optimizers leaner and faster at scale, with better fine-tuning quality as an additional benefit.

\section{Limitations}
\label{sec:limitations}
Language-pretraining quality reports held-out validation rather than an independent test set. The C4-440M and C4-1.1B comparisons each use three consecutive-seed reporting runs, and the 1.1B comparison fixes one shared, a priori learning rate rather than running method-specific HPO. The method assumes stable output coordinates; it retains within-batch covariance only through the current gradient, leaving changing output spaces and subgroup effects open. Systems timings are implementation-specific and use one unsharded device without gradient accumulation; distributed state, output, and communication costs require separate measurement (Appendix~\ref{app:distributed-accounting}).

\section*{AI Use Statement}
We used generative AI tools in four ways, matching the disclosures made at submission. (i)~Writing assistance: English-language editing and polishing of the text and LaTeX restructuring. (ii)~Retrieval and discovery: identifying related work and verifying bibliographic entries against their original sources. (iii)~Research execution: the method and its design were conceived by the authors without generative AI; generative AI tools were used to implement the proposed optimizer and the experimental pipeline in code from the authors' specification, and for numerical and statistical checks of reported values, figure preparation, feedback on experimental reporting, and assistance in interpreting results. (iv)~Drafting: drafting parts of the paper's sections, which the authors subsequently revised. We have reviewed all AI-assisted work and take responsibility for the final content of this work, including the text, analyses, citations, and scientific claims produced with the aid of generative AI.

\section*{Ethics Statement}
This work uses public benchmark datasets and involves no human subjects, personally identifying information, or new data collection. It changes optimizer state within standard training pipelines and introduces no identified application-specific risk. Dataset use and compute are documented in Appendix~\ref{app:reproducibility}.

\section*{Reproducibility Statement}
Algorithm~\ref{alg:bpm} specifies the optimizer. Code will be released at \url{https://github.com/lyclyq/Optimizor_arxiv}; it will provide \bpm{}, its AdamW, Adam-mini and GaLore integrations, and the language-modeling construction, with training entry points for RoBERTa-base fine-tuning and Qwen3 pretraining. Appendix~\ref{app:reproducibility} documents the reported training settings, HPO policy, and selected hyperparameters; Appendix~\ref{app:per-dataset-results} gives task-level, control, ablation, precision, and aggregate pretraining results. Pretraining uses five consecutive-seed reporting runs except for the C4-440M and C4-1.1B studies, which each use three.

\bibliographystyle{iclr2027_conference}
\bibliography{refs}

\clearpage
\appendix
\renewcommand{\topfraction}{0.95}
\renewcommand{\bottomfraction}{0.90}
\renewcommand{\textfraction}{0.05}
\renewcommand{\floatpagefraction}{0.80}
\setcounter{topnumber}{5}
\setcounter{bottomnumber}{3}
\setcounter{totalnumber}{8}
\makeatletter
\setlength{\@fptop}{0pt}
\setlength{\@fpsep}{8pt plus 2pt minus 2pt}
\setlength{\@fpbot}{0pt plus 1fil}
\makeatother
\section{Supplementary Results}
\label{app:per-dataset-results}
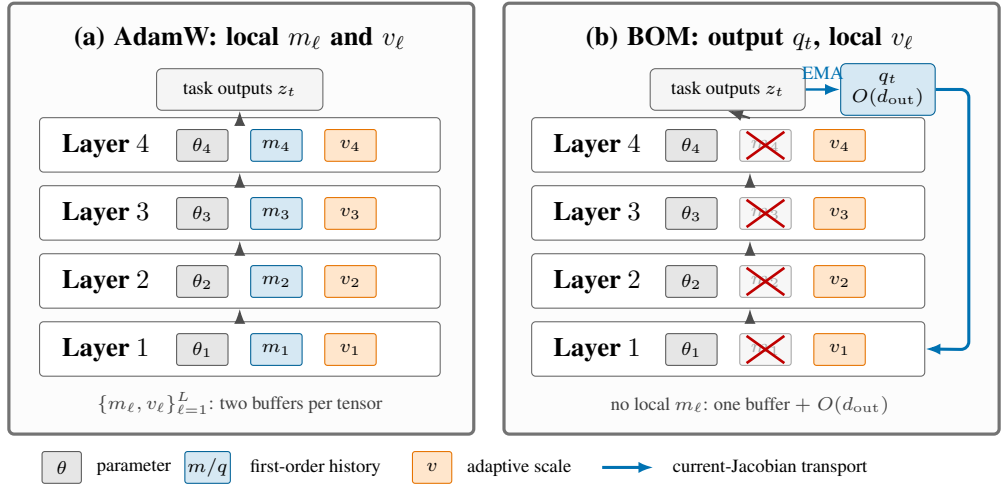
\begin{figure}[!htbp]
\centering
\begin{tikzpicture}[
    x=1cm,y=1cm,
    font=\scriptsize,
    panel/.style={draw=black!55, rounded corners=2pt, very thick, fill=black!1},
    layer/.style={draw=black!60, rounded corners=1.5pt, minimum height=0.72cm, fill=white},
    param/.style={draw=black!65, fill=black!10, rounded corners=1pt, minimum height=0.42cm, minimum width=0.68cm, inner sep=1pt},
    first/.style={draw=bpmblue!85!black, fill=bpmblue!16, rounded corners=1pt, minimum height=0.42cm, minimum width=0.68cm, inner sep=1pt},
    second/.style={draw=orange!90!black, fill=orange!20, rounded corners=1pt, minimum height=0.42cm, minimum width=0.68cm, inner sep=1pt},
    compact/.style={draw=bpmblue!85!black, fill=bpmblue!16, rounded corners=2pt, minimum height=0.58cm, minimum width=1.25cm, align=center},
    fwd/.style={-{Latex[length=2mm]}, line width=0.8pt, black!70},
    bwd/.style={-{Latex[length=2mm]}, line width=1.15pt, bpmblue},
    cross/.style={line width=1.0pt, red!75!black}
]
\definecolor{bpmblue}{RGB}{0,114,178}

\node[panel, minimum width=6.15cm, minimum height=5.70cm, anchor=south west] (adamPanel) at (0,0) {};
\node[font=\bfseries, align=center, text width=5.65cm, anchor=north] at (3.075,5.54) {(a) AdamW: local $m_\ell$ and $v_\ell$};

\foreach \y/\idx in {0.78/1,1.68/2,2.58/3,3.48/4}{
    \node[layer, minimum width=5.30cm, anchor=south west] (al\idx) at (0.42,\y) {};
    \node[anchor=west, font=\bfseries] at (0.60,\y+0.36) {Layer $\idx$};
    \node[param] at (2.58,\y+0.36) {$\theta_{\idx}$};
    \node[first] at (3.56,\y+0.36) {$m_{\idx}$};
    \node[second] at (4.54,\y+0.36) {$v_{\idx}$};
}
\draw[fwd] (3.075,1.50) -- (3.075,1.68);
\draw[fwd] (3.075,2.40) -- (3.075,2.58);
\draw[fwd] (3.075,3.30) -- (3.075,3.48);
\node[draw=black!60, fill=black!4, rounded corners=2pt, minimum width=2.2cm, minimum height=0.55cm] (aout) at (3.075,4.58) {task outputs $z_t$};
\draw[fwd] (3.075,4.20) -- (aout.south);
\node[anchor=south, align=center, text=black!75] at (3.075,0.16) {$\{m_\ell,v_\ell\}_{\ell=1}^{L}$: two buffers per tensor};

\node[panel, minimum width=6.55cm, minimum height=5.70cm, anchor=south west] (bpmPanel) at (6.55,0) {};
\node[font=\bfseries, align=center, text width=5.80cm, anchor=north] at (9.825,5.54) {(b) BOM: output $q_t$, local $v_\ell$};

\foreach \y/\idx in {0.78/1,1.68/2,2.58/3,3.48/4}{
    \node[layer, minimum width=5.20cm, anchor=south west] (bl\idx) at (6.93,\y) {};
    \node[anchor=west, font=\bfseries] at (7.10,\y+0.36) {Layer $\idx$};
    \node[param] at (9.05,\y+0.36) {$\theta_{\idx}$};
    \node[first, draw=black!35, fill=black!3, text=black!35] (bm\idx) at (10.03,\y+0.36) {$m_{\idx}$};
    \draw[cross] ($(bm\idx.center)+(-0.25,-0.18)$) -- ($(bm\idx.center)+(0.25,0.18)$);
    \draw[cross] ($(bm\idx.center)+(-0.25,0.18)$) -- ($(bm\idx.center)+(0.25,-0.18)$);
    \node[second] at (11.01,\y+0.36) {$v_{\idx}$};
}
\draw[fwd] (9.825,1.50) -- (9.825,1.68);
\draw[fwd] (9.825,2.40) -- (9.825,2.58);
\draw[fwd] (9.825,3.30) -- (9.825,3.48);
\node[draw=black!60, fill=black!4, rounded corners=2pt, minimum width=2.05cm, minimum height=0.55cm] (bout) at (9.53,4.58) {task outputs $z_t$};
\node[compact] (q) at (11.65,4.58) {$q_t$\\$O(d_{\mathrm{out}})$};
\draw[fwd] (9.825,4.20) -- (bout.south);
\draw[-{Latex[length=2mm]}, bpmblue, line width=0.9pt] (bout.east) -- node[above, text=bpmblue, font=\scriptsize] {EMA} (q.west);
\draw[bwd, rounded corners=3pt] (q.east) -- (12.72,4.58) -- (12.72,1.14) -- (bl1.east);
\node[anchor=south, align=center, text=black!75] at (9.825,0.16) {no local $m_\ell$: one buffer $+\;O(d_{\mathrm{out}})$};

\node[param, minimum width=0.52cm] at (0.72,-0.42) {$\theta$};
\node[anchor=west] at (1.06,-0.42) {parameter};
\node[first, minimum width=0.52cm] at (2.65,-0.42) {$m/q$};
\node[anchor=west] at (3.09,-0.42) {first-order history};
\node[second, minimum width=0.52cm] at (5.62,-0.42) {$v$};
\node[anchor=west] at (5.96,-0.42) {adaptive scale};
\draw[bwd] (7.87,-0.42) -- (8.55,-0.42);
\node[anchor=west] at (8.68,-0.42) {current-Jacobian transport};
\end{tikzpicture}
\caption{Optimizer-state placement in AdamW and BOM. AdamW retains dense first- and second-moment tensors for every trainable layer. BOM deletes every dense local first moment (crossed out), retains the local second moment used for adaptive scaling, and stores one compact task-space history $q_t$. The current backward graph re-expresses that history as updates for all trainable tensors, rather than preserving the parameter coordinates induced by earlier Jacobians. Parameter, gradient, and activation memory are omitted because the diagram isolates optimizer state.}
\label{fig:bpm-architecture}
\end{figure}

This appendix reports every task-level result behind the aggregate claims of Section~\ref{sec:experiments}, together with the per-seed statistics and the control and ablation tables. Figure~\ref{fig:bpm-architecture} first contrasts where the two optimizers hold their state, the structural picture summarized in Section~\ref{sec:intuition}, and Algorithm~\ref{alg:bpm} lists the complete update step with AdamW-style scaling.

\begin{algorithm}[H]
\caption{Backpropagated Output Momentum with AdamW-style scaling}
\label{alg:bpm}
\begin{algorithmic}[1]
\Require Parameters $\theta_\ell$, step sizes $\eta_t$, coefficients $\lambda,\beta_1,\beta_2,\epsilon,\omega_\ell$.
\State Initialize one shared $q_0=0\in\mathbb R^{d_{\mathrm{out}}}$ and $v_{\ell,0}=0$ for each trainable tensor.
\For{$t=1,2,\ldots$}
\State Let $\mathcal I_t$ index the supervised outputs of the batch, $N_t=|\mathcal I_t|>0$: examples in classification, non-ignored target positions in language modeling. Compute their logits $z_{t,i}$, $i\in\mathcal I_t$, and mean cross-entropy $\mathcal L_t^{\mathrm{CE}}$; all means below are over $\mathcal I_t$.
\State $s_t\gets N_t^{-1}\sum_{i\in\mathcal I_t}[\operatorname{softmax}(\operatorname{stopgrad}(z_{t,i}))-\operatorname{onehot}(y_{t,i})]$.
\State $q_t\gets\beta_1q_{t-1}+(1-\beta_1)s_t$; $\hat q_t\gets q_t/(1-\beta_1^t)$; keep this state detached.
\State $\mathcal L_t^{\mathrm{mix}}\gets(1-\lambda)\mathcal L_t^{\mathrm{CE}}+\lambda N_t^{-1}\sum_{i\in\mathcal I_t}\langle z_{t,i},\hat q_t\rangle$.
\State One backward pass gives $u_{\ell,t}\gets\nabla_{\theta_\ell}\mathcal L_t^{\mathrm{mix}}$ for all trainable tensors.
\State $v_{\ell,t}\gets\beta_2v_{\ell,t-1}+(1-\beta_2)u_{\ell,t}^{\odot2}$; $\hat v_{\ell,t}\gets v_{\ell,t}/(1-\beta_2^t)$.
\State $\theta_\ell\gets(1-\eta_t\omega_\ell)\theta_\ell-\eta_t u_{\ell,t}/(\sqrt{\hat v_{\ell,t}}+\epsilon)$.
\EndFor
\end{algorithmic}
\end{algorithm}

\subsection{NLP Per-Dataset Results}
\label{app:nlp-results}
Table~\ref{tab:nlp_appendix_bpm} reports the NLP per-dataset results for the five-task main suite. Table~\ref{tab:pertask-ci} separates the descriptive RoBERTa-base task-level margins in panel~(a) from the five-task macro estimate and descriptive interval in panel~(b). On CoLA, RoBERTa-base improves Matthews correlation despite higher validation cross-entropy; DeBERTa-v3-base worsens on both measures, while Qwen3-1.7B improves on both. Matthews correlation summarizes discrete classification decisions, whereas cross-entropy also depends on predicted probabilities. Table~\ref{tab:perseed-roberta} gives the per-seed validation scores behind the primary AdamW--\bpm{} rows on all three backbones, with the paired difference and its interval for each task, so every mean can be read against its five constituent runs.

\begin{table}[!htbp]
\centering
\setlength{\tabcolsep}{4.0pt}
\caption{Per-dataset results for AdamW and \bpm{} on the three pretrained
backbones, at the epoch-5 endpoint over the five evaluation seeds. Score is the metric named in
each row; loss is cross-entropy.}
\label{tab:nlp_appendix_bpm}
\begin{tabular}{llllcccc}
\toprule
Backbone & Dataset & Metric & Method & Train score & Val score & Train loss & Val loss \\
\midrule
RoBERTa-base     & CoLA   & MCC  & AdamW    & 0.8566 & 0.5743 & 0.1851 & 0.4777 \\
RoBERTa-base     & CoLA   & MCC  & \bpm{}   & 0.9641 & 0.5970 & 0.0495 & 0.5794 \\
RoBERTa-base     & MRPC   & F1   & AdamW    & 0.9609 & 0.9077 & 0.1324 & 0.3240 \\
RoBERTa-base     & MRPC   & F1   & \bpm{}   & 0.9862 & 0.9191 & 0.0616 & 0.3305 \\
RoBERTa-base     & QQP    & F1   & AdamW    & 0.8501 & 0.8463 & 0.2620 & 0.2683 \\
RoBERTa-base     & QQP    & F1   & \bpm{}   & 0.9004 & 0.8626 & 0.1960 & 0.2481 \\
RoBERTa-base     & RTE    & Acc  & AdamW    & 0.7844 & 0.7155 & 0.4386 & 0.5894 \\
RoBERTa-base     & RTE    & Acc  & \bpm{}   & 0.8906 & 0.7350 & 0.2936 & 0.5678 \\
RoBERTa-base     & SST-2  & Acc  & AdamW    & 0.9641 & 0.9372 & 0.0994 & 0.1975 \\
RoBERTa-base     & SST-2  & Acc  & \bpm{}   & 0.9820 & 0.9385 & 0.0611 & 0.1873 \\
\midrule
DeBERTa-v3-base  & CoLA   & MCC  & AdamW    & 0.8634 & 0.6628 & 0.1718 & 0.3906 \\
DeBERTa-v3-base  & CoLA   & MCC  & \bpm{}   & 0.9531 & 0.6547 & 0.0698 & 0.4884 \\
DeBERTa-v3-base  & MRPC   & F1   & AdamW    & 0.9593 & 0.9102 & 0.1553 & 0.3382 \\
DeBERTa-v3-base  & MRPC   & F1   & \bpm{}   & 0.9932 & 0.9246 & 0.0506 & 0.3223 \\
DeBERTa-v3-base  & QQP    & F1   & AdamW    & 0.8509 & 0.8597 & 0.2559 & 0.2518 \\
DeBERTa-v3-base  & QQP    & F1   & \bpm{}   & 0.9103 & 0.8700 & 0.1693 & 0.2426 \\
DeBERTa-v3-base  & RTE    & Acc  & AdamW    & 0.9563 & 0.8123 & 0.1395 & 0.4962 \\
DeBERTa-v3-base  & RTE    & Acc  & \bpm{}   & 0.9891 & 0.8318 & 0.0497 & 0.5100 \\
DeBERTa-v3-base  & SST-2  & Acc  & AdamW    & 0.9055 & 0.9195 & 0.2007 & 0.1812 \\
DeBERTa-v3-base  & SST-2  & Acc  & \bpm{}   & 0.9844 & 0.9525 & 0.0560 & 0.1406 \\
\midrule
Qwen3-1.7B       & CoLA   & MCC  & AdamW    & 0.9839 & 0.5596 & 0.0205 & 0.6892 \\
Qwen3-1.7B       & CoLA   & MCC  & \bpm{}   & 0.9839 & 0.5804 & 0.0296 & 0.4849 \\
Qwen3-1.7B       & MRPC   & F1   & AdamW    & 1.0000 & 0.8841 & 0.0020 & 0.7610 \\
Qwen3-1.7B       & MRPC   & F1   & \bpm{}   & 0.9994 & 0.9012 & 0.0080 & 0.4642 \\
Qwen3-1.7B       & QQP    & F1   & AdamW    & 0.9952 & 0.8730 & 0.0213 & 0.3986 \\
Qwen3-1.7B       & QQP    & F1   & \bpm{}   & 0.9936 & 0.8852 & 0.0215 & 0.2874 \\
Qwen3-1.7B       & RTE    & Acc  & AdamW    & 0.9977 & 0.8137 & 0.0164 & 0.6187 \\
Qwen3-1.7B       & RTE    & Acc  & \bpm{}   & 1.0000 & 0.8390 & 0.0116 & 0.4952 \\
Qwen3-1.7B       & SST-2  & Acc  & AdamW    & 1.0000 & 0.9539 & 0.0041 & 0.2184 \\
Qwen3-1.7B       & SST-2  & Acc  & \bpm{}   & 0.9953 & 0.9461 & 0.0147 & 0.1962 \\
\bottomrule
\end{tabular}
\end{table}

\begin{table}[!htbp]
\centering
\setlength{\tabcolsep}{3.5pt}
\caption{Paired \bpm{}$-$AdamW validation-score margins on RoBERTa-base at the epoch-5 endpoint over the five evaluation seeds. Panel~(a) reports the mean paired difference in each task's native metric with its descriptive paired 95\% interval; panel~(b) reports the five-task macro comparison of Section~\ref{sec:balance}. Intervals use $t_{0.975,4}=2.776$ on the five paired differences. Seeds are matched across the two arms, so the paired spread is the relevant one: on RTE each arm's own standard deviation ($0.021$ and $0.029$) exceeds the margin, while the paired standard deviation is $0.0098$. Per-seed values are in Table~\ref{tab:perseed-roberta}.}
\label{tab:pertask-ci}

\begin{tabular}{lccc}
\toprule
\multicolumn{4}{l}{\textit{(a) Task-level paired margins}} \\
Dataset & Metric & $\Delta$ & 95\% CI \\
\midrule
CoLA & MCC & $+0.0226$ & $[-0.0226,\,+0.0679]$ \\
MRPC & F1 & $+0.0114$ & $[-0.0079,\,+0.0306]$ \\
QQP & F1 & $+0.0163$ & $[+0.0059,\,+0.0267]$ \\
RTE & Acc & $+0.0195$ & $[+0.0074,\,+0.0316]$ \\
SST-2 & Acc & $+0.0014$ & $[-0.0116,\,+0.0144]$ \\
\addlinespace
\midrule
\multicolumn{4}{l}{\textit{(b) Aggregate comparison}} \\
Aggregation & & $\Delta$ & 95\% CI \\
\midrule
Five-task macro & & $+0.0142$ & $[+0.0027,\,+0.0258]$ \\
\bottomrule
\end{tabular}
\end{table}

\FloatBarrier

\begingroup
\setlength{\tabcolsep}{3pt}
\begin{longtable}{llrrrrrr}
\caption{Per-seed validation scores for the primary AdamW--\bpm{} comparison on all three backbones, at the epoch-5 endpoint. Score is the task-appropriate metric named with each block; the five evaluation seeds are disjoint from the two HPO seeds. Labels A--E identify the matched evaluation runs, and the Mean column is the value reported in Table~\ref{tab:nlp_appendix_bpm}. $\Delta$ is the mean paired difference with its descriptive 95\% interval ($t_{0.975,4}=2.776$ on the five paired differences); intervals use the variation of matched seed differences, which need not be smaller than either arm's marginal variation.}
\label{tab:perseed-roberta}\\
\toprule
Task & Method & A & B & C & D & E & Mean \\
\midrule

\endfirsthead
\caption[]{Per-seed validation scores (continued).}\\
\toprule
Task & Method & A & B & C & D & E & Mean \\
\midrule

\endhead
\bottomrule
\endfoot
\multicolumn{8}{l}{\textbf{RoBERTa-base}} \\
CoLA (MCC) & \adamw{} & 0.5435 & 0.5778 & 0.6131 & 0.5834 & 0.5539 & 0.5743 \\*
 & \bpm{} & 0.5885 & 0.5536 & 0.6107 & 0.6111 & 0.6209 & 0.5970 \\*
\multicolumn{8}{r}{$\Delta$ (95\% CI): $+0.0226$ $[-0.0226,\,+0.0679]$} \\
\addlinespace[2pt]
MRPC (F1) & \adamw{} & 0.9004 & 0.9127 & 0.9171 & 0.9084 & 0.8998 & 0.9077 \\*
 & \bpm{} & 0.9255 & 0.9308 & 0.9020 & 0.9206 & 0.9164 & 0.9191 \\*
\multicolumn{8}{r}{$\Delta$ (95\% CI): $+0.0114$ $[-0.0079,\,+0.0306]$} \\
\addlinespace[2pt]
QQP (F1) & \adamw{} & 0.8457 & 0.8486 & 0.8456 & 0.8448 & 0.8468 & 0.8463 \\*
 & \bpm{} & 0.8653 & 0.8537 & 0.8661 & 0.8706 & 0.8572 & 0.8626 \\*
\multicolumn{8}{r}{$\Delta$ (95\% CI): $+0.0163$ $[+0.0059,\,+0.0267]$} \\
\addlinespace[2pt]
RTE (Acc) & \adamw{} & 0.7148 & 0.7076 & 0.6895 & 0.7473 & 0.7184 & 0.7155 \\*
 & \bpm{} & 0.7292 & 0.7184 & 0.7076 & 0.7834 & 0.7365 & 0.7350 \\*
\multicolumn{8}{r}{$\Delta$ (95\% CI): $+0.0195$ $[+0.0074,\,+0.0316]$} \\
\addlinespace[2pt]
SST-2 (Acc) & \adamw{} & 0.9381 & 0.9346 & 0.9346 & 0.9392 & 0.9392 & 0.9372 \\*
 & \bpm{} & 0.9461 & 0.9358 & 0.9472 & 0.9243 & 0.9392 & 0.9385 \\*
\multicolumn{8}{r}{$\Delta$ (95\% CI): $+0.0014$ $[-0.0116,\,+0.0144]$} \\
\addlinespace[2pt]
\multicolumn{8}{l}{\textbf{DeBERTa-v3-base}} \\
CoLA (MCC) & \adamw{} & 0.6751 & 0.6676 & 0.6556 & 0.6505 & 0.6654 & 0.6628 \\*
 & \bpm{} & 0.7090 & 0.6178 & 0.6192 & 0.6946 & 0.6331 & 0.6547 \\*
\multicolumn{8}{r}{$\Delta$ (95\% CI): $-0.0081$ $[-0.0622,\,+0.0460]$} \\
\addlinespace[2pt]
MRPC (F1) & \adamw{} & 0.9091 & 0.9211 & 0.9074 & 0.9062 & 0.9069 & 0.9102 \\*
 & \bpm{} & 0.9319 & 0.9244 & 0.9170 & 0.9283 & 0.9215 & 0.9246 \\*
\multicolumn{8}{r}{$\Delta$ (95\% CI): $+0.0145$ $[+0.0041,\,+0.0248]$} \\
\addlinespace[2pt]
QQP (F1) & \adamw{} & 0.8581 & 0.8585 & 0.8603 & 0.8597 & 0.8619 & 0.8597 \\*
 & \bpm{} & 0.8749 & 0.8568 & 0.8756 & 0.8773 & 0.8652 & 0.8700 \\*
\multicolumn{8}{r}{$\Delta$ (95\% CI): $+0.0103$ $[-0.0007,\,+0.0212]$} \\
\addlinespace[2pt]
RTE (Acc) & \adamw{} & 0.7906 & 0.8159 & 0.8303 & 0.8159 & 0.8087 & 0.8123 \\*
 & \bpm{} & 0.8556 & 0.8484 & 0.8014 & 0.8267 & 0.8267 & 0.8318 \\*
\multicolumn{8}{r}{$\Delta$ (95\% CI): $+0.0195$ $[-0.0229,\,+0.0618]$} \\
\addlinespace[2pt]
SST-2 (Acc) & \adamw{} & 0.9530 & 0.9495 & 0.9484 & 0.7993 & 0.9472 & 0.9195 \\*
 & \bpm{} & 0.9530 & 0.9484 & 0.9553 & 0.9541 & 0.9518 & 0.9525 \\*
\multicolumn{8}{r}{$\Delta$ (95\% CI): $+0.0330$ $[-0.0516,\,+0.1176]$} \\
\addlinespace[2pt]
\multicolumn{8}{l}{\textbf{Qwen3-1.7B}} \\
CoLA (MCC) & \adamw{} & 0.5608 & 0.6041 & 0.5074 & 0.5470 & 0.5786 & 0.5596 \\*
 & \bpm{} & 0.6141 & 0.5765 & 0.5925 & 0.5674 & 0.5514 & 0.5804 \\*
\multicolumn{8}{r}{$\Delta$ (95\% CI): $+0.0208$ $[-0.0408,\,+0.0824]$} \\
\addlinespace[2pt]
MRPC (F1) & \adamw{} & 0.8744 & 0.8897 & 0.8789 & 0.8776 & 0.8998 & 0.8841 \\*
 & \bpm{} & 0.9129 & 0.8862 & 0.8982 & 0.9005 & 0.9082 & 0.9012 \\*
\multicolumn{8}{r}{$\Delta$ (95\% CI): $+0.0171$ $[-0.0024,\,+0.0367]$} \\
\addlinespace[2pt]
QQP (F1) & \adamw{} & 0.8753 & 0.8706 & 0.8711 & 0.8753 & 0.8725 & 0.8730 \\*
 & \bpm{} & 0.8731 & 0.8890 & 0.8888 & 0.8888 & 0.8864 & 0.8852 \\*
\multicolumn{8}{r}{$\Delta$ (95\% CI): $+0.0123$ $[+0.0018,\,+0.0227]$} \\
\addlinespace[2pt]
RTE (Acc) & \adamw{} & 0.8267 & 0.8159 & 0.7978 & 0.8159 & 0.8123 & 0.8137 \\*
 & \bpm{} & 0.8664 & 0.8484 & 0.8339 & 0.8123 & 0.8339 & 0.8390 \\*
\multicolumn{8}{r}{$\Delta$ (95\% CI): $+0.0253$ $[+0.0035,\,+0.0470]$} \\
\addlinespace[2pt]
SST-2 (Acc) & \adamw{} & 0.9530 & 0.9576 & 0.9599 & 0.9484 & 0.9507 & 0.9539 \\*
 & \bpm{} & 0.9507 & 0.9369 & 0.9495 & 0.9484 & 0.9450 & 0.9461 \\*
\multicolumn{8}{r}{$\Delta$ (95\% CI): $-0.0078$ $[-0.0179,\,+0.0023]$} \\
\addlinespace[2pt]
\end{longtable}
\endgroup

\FloatBarrier

\subsection{Fixed-Backbone RoBERTa Control Results}
\label{app:roberta-control-results}
\begin{figure}[!htbp]
\centering
\begin{minipage}{0.495\textwidth}
\centering
\begin{tikzpicture}
\begin{axis}[
  width=0.92\linewidth,height=5.2cm,
  title={(a) Validation task score},
  xlabel={Difference from AdamW (points)\\Higher is better},
  xmin=-6,xmax=3,xtick={-6,-3,0,3},
  ymin=0.4,ymax=6.6,ytick={1,2,3,4,5,6},
  yticklabels={SCALE,Adam-mini,GaLore,Muon,Lion,\bpm{}-AdamW},
  tick label style={font=\scriptsize},label style={font=\scriptsize},
  xlabel style={align=center},
  title style={font=\small\bfseries},
  axis x line*=bottom,axis y line*=left,
  axis line style={draw=gray!55},tick style={draw=gray!55},
  xmajorgrids=true,grid style={line width=.1pt,draw=gray!18},
  scaled x ticks=false,
  xticklabel style={/pgf/number format/fixed,/pgf/number format/precision=2},
]
\addplot[gray!65,densely dashed,no marks] coordinates {(0,0.4) (0,6.6)};
\addplot[only marks,mark=square*,mark size=2.3pt,color=bpmblue,solid,line width=0.85pt,
  mark options={draw=bpmblue,fill=bpmblue,solid},
  error bars/.cd,x dir=both,x explicit,error bar style={draw=bpmblue,solid,line width=0.85pt},
  error mark options={draw=bpmblue,solid,rotate=90,mark size=2.5pt,line width=0.85pt}]
  coordinates {(1.423865164,6) +- (1.153304046,0)};
\addplot[only marks,mark=*,mark size=1.8pt,color=adamgray,solid,line width=0.85pt,
  mark options={draw=adamgray,fill=adamgray,solid},
  error bars/.cd,x dir=both,x explicit,error bar style={draw=adamgray,solid,line width=0.85pt},
  error mark options={draw=adamgray,solid,rotate=90,mark size=2.5pt,line width=0.85pt}]
  coordinates {(0.034364592,5) +- (1.180061798,0)};
\addplot[only marks,mark=*,mark size=1.8pt,color=adamgray,solid,line width=0.85pt,
  mark options={draw=adamgray,fill=adamgray,solid},
  error bars/.cd,x dir=both,x explicit,error bar style={draw=adamgray,solid,line width=0.85pt},
  error mark options={draw=adamgray,solid,rotate=90,mark size=2.5pt,line width=0.85pt}]
  coordinates {(-1.767501180,4) +- (0.950006235,0)};
\addplot[only marks,mark=*,mark size=1.8pt,color=adamgray,solid,line width=0.85pt,
  mark options={draw=adamgray,fill=adamgray,solid},
  error bars/.cd,x dir=both,x explicit,error bar style={draw=adamgray,solid,line width=0.85pt},
  error mark options={draw=adamgray,solid,rotate=90,mark size=2.5pt,line width=0.85pt}]
  coordinates {(-2.225872004,3) +- (1.350260350,0)};
\addplot[only marks,mark=*,mark size=1.8pt,color=adamgray,solid,line width=0.85pt,
  mark options={draw=adamgray,fill=adamgray,solid},
  error bars/.cd,x dir=both,x explicit,error bar style={draw=adamgray,solid,line width=0.85pt},
  error mark options={draw=adamgray,solid,rotate=90,mark size=2.5pt,line width=0.85pt}]
  coordinates {(-3.029741428,2) +- (1.420742347,0)};
\addplot[only marks,mark=*,mark size=1.8pt,color=adamgray,solid,line width=0.85pt,
  mark options={draw=adamgray,fill=adamgray,solid},
  error bars/.cd,x dir=both,x explicit,error bar style={draw=adamgray,solid,line width=0.85pt},
  error mark options={draw=adamgray,solid,rotate=90,mark size=2.5pt,line width=0.85pt}]
  coordinates {(-3.709141109,1) +- (1.523678383,0)};
\end{axis}
\end{tikzpicture}
\end{minipage}
\hfill
\begin{minipage}{0.495\textwidth}
\centering
\begin{tikzpicture}
\begin{axis}[
  width=0.92\linewidth,height=5.2cm,
  title={(b) Validation cross-entropy},
  xlabel={Difference from AdamW\\Lower is better},
  xmin=-0.05,xmax=0.30,xtick={0,0.1,0.2,0.3},
  ymin=0.4,ymax=6.6,ytick={1,2,3,4,5,6},
  yticklabels={SCALE,Adam-mini,GaLore,Muon,Lion,\bpm{}-AdamW},
  tick label style={font=\scriptsize},label style={font=\scriptsize},
  xlabel style={align=center},
  title style={font=\small\bfseries},
  axis x line*=bottom,axis y line*=left,
  axis line style={draw=gray!55},tick style={draw=gray!55},
  xmajorgrids=true,grid style={line width=.1pt,draw=gray!18},
  scaled x ticks=false,
  xticklabel style={/pgf/number format/fixed,/pgf/number format/precision=2},
]
\addplot[gray!65,densely dashed,no marks] coordinates {(0,0.4) (0,6.6)};
\addplot[only marks,mark=square*,mark size=2.3pt,color=bpmblue,solid,line width=0.85pt,
  mark options={draw=bpmblue,fill=bpmblue,solid},
  error bars/.cd,x dir=both,x explicit,error bar style={draw=bpmblue,solid,line width=0.85pt},
  error mark options={draw=bpmblue,solid,rotate=90,mark size=2.5pt,line width=0.85pt}]
  coordinates {(0.011226508,6) +- (0.033758744,0)};
\addplot[only marks,mark=*,mark size=1.8pt,color=adamgray,solid,line width=0.85pt,
  mark options={draw=adamgray,fill=adamgray,solid},
  error bars/.cd,x dir=both,x explicit,error bar style={draw=adamgray,solid,line width=0.85pt},
  error mark options={draw=adamgray,solid,rotate=90,mark size=2.5pt,line width=0.85pt}]
  coordinates {(0.062569755,5) +- (0.023766084,0)};
\addplot[only marks,mark=*,mark size=1.8pt,color=adamgray,solid,line width=0.85pt,
  mark options={draw=adamgray,fill=adamgray,solid},
  error bars/.cd,x dir=both,x explicit,error bar style={draw=adamgray,solid,line width=0.85pt},
  error mark options={draw=adamgray,solid,rotate=90,mark size=2.5pt,line width=0.85pt}]
  coordinates {(0.191371382,4) +- (0.044526871,0)};
\addplot[only marks,mark=*,mark size=1.8pt,color=adamgray,solid,line width=0.85pt,
  mark options={draw=adamgray,fill=adamgray,solid},
  error bars/.cd,x dir=both,x explicit,error bar style={draw=adamgray,solid,line width=0.85pt},
  error mark options={draw=adamgray,solid,rotate=90,mark size=2.5pt,line width=0.85pt}]
  coordinates {(0.015938726,3) +- (0.016417842,0)};
\addplot[only marks,mark=*,mark size=1.8pt,color=adamgray,solid,line width=0.85pt,
  mark options={draw=adamgray,fill=adamgray,solid},
  error bars/.cd,x dir=both,x explicit,error bar style={draw=adamgray,solid,line width=0.85pt},
  error mark options={draw=adamgray,solid,rotate=90,mark size=2.5pt,line width=0.85pt}]
  coordinates {(0.011190386,2) +- (0.024600034,0)};
\addplot[only marks,mark=*,mark size=1.8pt,color=adamgray,solid,line width=0.85pt,
  mark options={draw=adamgray,fill=adamgray,solid},
  error bars/.cd,x dir=both,x explicit,error bar style={draw=adamgray,solid,line width=0.85pt},
  error mark options={draw=adamgray,solid,rotate=90,mark size=2.5pt,line width=0.85pt}]
  coordinates {(0.069063898,1) +- (0.021886831,0)};
\end{axis}
\end{tikzpicture}
\end{minipage}
\caption{RoBERTa-base control effects relative to AdamW at the epoch-5 endpoint (Table~\ref{tab:roberta-control-endpoints}). Each point is the mean method-minus-AdamW difference; bars are descriptive, unadjusted 95\% paired $t$ confidence intervals over five evaluation seeds ($df=4$). For each seed, we first average the task-appropriate score or cross-entropy across the five tasks, then pair the resulting macro with AdamW at the same seed. Score differences are multiplied by 100; positive values favor the method. Negative cross-entropy differences favor the method. The dashed line marks no difference. These intervals describe seed variation conditional on the selected configurations and this fixed task set.}
\label{fig:roberta-control-summary}
\end{figure}
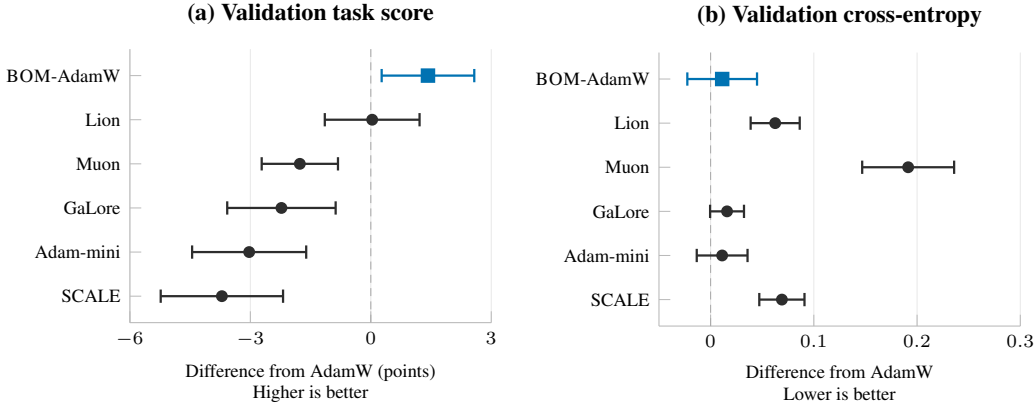
Table~\ref{tab:roberta_controls_appendix} gives the per-dataset RoBERTa-base results for the additional optimizer controls used in the fixed-backbone mechanism comparison; the matched AdamW and \bpm{} rows are included for reference.
The suite combines smaller datasets (CoLA, MRPC and RTE) with larger ones (SST-2 and QQP), while the vision and from-scratch studies extend coverage across domains and initialization regimes. 
The controls follow the shared selection policy in Appendix~\ref{app:reproducibility}; Muon additionally receives an update-scale search (Table~\ref{tab:repro-params-method}). Implementations are documented in Appendix~\ref{app:optimizer-implementation}.

\begin{table}[!htbp]
\centering
\setlength{\tabcolsep}{3.5pt}
\caption{RoBERTa-base five-task macro endpoints for the seven methods in the
fixed-backbone comparison, at the epoch-5 endpoint over the five evaluation seeds. Parameter-shaped optimizer state for the substitution pairs
is reported with the composition comparison (Table~\ref{tab:composition}).}
\label{tab:roberta-control-endpoints}

\begin{tabular}{lcccc}
\toprule
Method & Train score & Val score & Train loss & Val loss \\
\midrule
\bpm{}-AdamW           & 0.9447 & 0.8104 & 0.1324 & 0.3826 \\
AdamW                  & 0.8832 & 0.7962 & 0.2235 & 0.3714 \\
Lion                   & 0.9302 & 0.7965 & 0.1340 & 0.4340 \\
Muon                   & 0.8833 & 0.7785 & 0.2029 & 0.5628 \\
GaLore                 & 0.8134 & 0.7739 & 0.3128 & 0.3873 \\
Adam-mini              & 0.7967 & 0.7659 & 0.3406 & 0.3826 \\
SCALE                  & 0.8409 & 0.7591 & 0.2883 & 0.4405 \\
\bottomrule
\end{tabular}
\end{table}

\FloatBarrier

\begingroup
\setlength{\tabcolsep}{4pt}
\begin{longtable}{lllcccc}
\caption{RoBERTa-base per-dataset results for every optimizer control and every
\bpm{} composition, at the epoch-5 endpoint over the five evaluation seeds. Score is the
metric named in each row; loss is cross-entropy. Train quantities are measured
on a fixed probe subset of the training set.}
\label{tab:roberta_controls_appendix}
\\
\toprule
Dataset & Metric & Method & Train score & Val score & Train loss & Val loss \\
\midrule
\endfirsthead
\caption[]{RoBERTa-base optimizer results (continued).}\\
\toprule
Dataset & Metric & Method & Train score & Val score & Train loss & Val loss \\
\midrule
\endhead
\bottomrule
\endfoot

CoLA   & MCC  & AdamW                  & 0.8566 & 0.5743 & 0.1851 & 0.4777 \\
CoLA   & MCC  & \bpm{}-AdamW           & 0.9641 & 0.5970 & 0.0495 & 0.5794 \\
CoLA   & MCC  & Lion                   & 0.9173 & 0.5780 & 0.0945 & 0.5178 \\
CoLA   & MCC  & Adam-mini              & 0.6403 & 0.5348 & 0.3614 & 0.4639 \\
CoLA   & MCC  & \bpm{}-Adam-mini       & 0.7098 & 0.5478 & 0.3197 & 0.4568 \\
CoLA   & MCC  & GaLore                 & 0.6668 & 0.5495 & 0.3521 & 0.4386 \\
CoLA   & MCC  & \bpm{}-GaLore          & 0.6918 & 0.5561 & 0.3133 & 0.4607 \\
CoLA   & MCC  & Muon                   & 0.6628 & 0.5068 & 0.3568 & 0.4915 \\
CoLA   & MCC  & SCALE                  & 0.7209 & 0.5348 & 0.3058 & 0.5144 \\
\midrule
MRPC   & F1   & AdamW                  & 0.9609 & 0.9077 & 0.1324 & 0.3240 \\
MRPC   & F1   & \bpm{}-AdamW           & 0.9862 & 0.9191 & 0.0616 & 0.3305 \\
MRPC   & F1   & Lion                   & 0.9938 & 0.9099 & 0.0350 & 0.3936 \\
MRPC   & F1   & Adam-mini              & 0.9277 & 0.9073 & 0.2348 & 0.2956 \\
MRPC   & F1   & \bpm{}-Adam-mini       & 0.9285 & 0.9086 & 0.2453 & 0.3181 \\
MRPC   & F1   & GaLore                 & 0.9385 & 0.8948 & 0.1932 & 0.3572 \\
MRPC   & F1   & \bpm{}-GaLore          & 0.9480 & 0.9063 & 0.1787 & 0.3404 \\
MRPC   & F1   & Muon                   & 0.8700 & 0.8538 & 0.3947 & 0.4396 \\
MRPC   & F1   & SCALE                  & 0.9282 & 0.8763 & 0.2409 & 0.3853 \\
\midrule
QQP    & F1   & AdamW                  & 0.8501 & 0.8463 & 0.2620 & 0.2683 \\
QQP    & F1   & \bpm{}-AdamW           & 0.9004 & 0.8626 & 0.1960 & 0.2481 \\
QQP    & F1   & Lion                   & 0.8284 & 0.8257 & 0.2910 & 0.2965 \\
QQP    & F1   & Adam-mini              & 0.8185 & 0.8170 & 0.3256 & 0.3212 \\
QQP    & F1   & \bpm{}-Adam-mini       & 0.8264 & 0.8193 & 0.3275 & 0.3285 \\
QQP    & F1   & GaLore                 & 0.8448 & 0.8363 & 0.2808 & 0.2872 \\
QQP    & F1   & \bpm{}-GaLore          & 0.8526 & 0.8301 & 0.2767 & 0.2976 \\
QQP    & F1   & Muon                   & 0.9214 & 0.8737 & 0.1611 & 0.2469 \\
QQP    & F1   & SCALE                  & 0.8842 & 0.8349 & 0.2433 & 0.3288 \\
\midrule
RTE    & Acc  & AdamW                  & 0.7844 & 0.7155 & 0.4386 & 0.5894 \\
RTE    & Acc  & \bpm{}-AdamW           & 0.8906 & 0.7350 & 0.2936 & 0.5678 \\
RTE    & Acc  & Lion                   & 0.9516 & 0.7372 & 0.1307 & 0.7726 \\
RTE    & Acc  & Adam-mini              & 0.6531 & 0.6433 & 0.6188 & 0.6334 \\
RTE    & Acc  & \bpm{}-Adam-mini       & 0.6906 & 0.6787 & 0.5687 & 0.5989 \\
RTE    & Acc  & GaLore                 & 0.6648 & 0.6664 & 0.6186 & 0.6299 \\
RTE    & Acc  & \bpm{}-GaLore          & 0.7477 & 0.7025 & 0.5079 & 0.5751 \\
RTE    & Acc  & Muon                   & 0.9898 & 0.7155 & 0.0240 & 1.4303 \\
RTE    & Acc  & SCALE                  & 0.7039 & 0.6578 & 0.5258 & 0.6077 \\
\midrule
SST-2  & Acc  & AdamW                  & 0.9641 & 0.9372 & 0.0994 & 0.1975 \\
SST-2  & Acc  & \bpm{}-AdamW           & 0.9820 & 0.9385 & 0.0611 & 0.1873 \\
SST-2  & Acc  & Lion                   & 0.9602 & 0.9319 & 0.1191 & 0.1893 \\
SST-2  & Acc  & Adam-mini              & 0.9437 & 0.9271 & 0.1625 & 0.1988 \\
SST-2  & Acc  & \bpm{}-Adam-mini       & 0.9477 & 0.9367 & 0.1404 & 0.1933 \\
SST-2  & Acc  & GaLore                 & 0.9523 & 0.9227 & 0.1194 & 0.2238 \\
SST-2  & Acc  & \bpm{}-GaLore          & 0.9656 & 0.9319 & 0.0945 & 0.2010 \\
SST-2  & Acc  & Muon                   & 0.9727 & 0.9427 & 0.0779 & 0.2055 \\
SST-2  & Acc  & SCALE                  & 0.9672 & 0.8917 & 0.1256 & 0.3661 \\
\midrule
\textbf{Macro mean} & -- & AdamW                  & 0.8832 & 0.7962 & 0.2235 & 0.3714 \\
\textbf{Macro mean} & -- & \bpm{}-AdamW           & 0.9447 & 0.8104 & 0.1324 & 0.3826 \\
\textbf{Macro mean} & -- & Lion                   & 0.9302 & 0.7965 & 0.1340 & 0.4340 \\
\textbf{Macro mean} & -- & Adam-mini              & 0.7967 & 0.7659 & 0.3406 & 0.3826 \\
\textbf{Macro mean} & -- & \bpm{}-Adam-mini       & 0.8206 & 0.7782 & 0.3203 & 0.3791 \\
\textbf{Macro mean} & -- & GaLore                 & 0.8134 & 0.7739 & 0.3128 & 0.3873 \\
\textbf{Macro mean} & -- & \bpm{}-GaLore          & 0.8411 & 0.7854 & 0.2742 & 0.3750 \\
\textbf{Macro mean} & -- & Muon                   & 0.8833 & 0.7785 & 0.2029 & 0.5628 \\
\textbf{Macro mean} & -- & SCALE                  & 0.8409 & 0.7591 & 0.2883 & 0.4405 \\
\end{longtable}
\endgroup

The comparison in the main text rests on the paired per-seed margins. On the five-task validation macro the paired per-seed margins favour \bpm{} over every control. Against AdamW, the paired five-seed difference is $+0.0142$, with a descriptive 95\% CI of $[+0.0027,+0.0258]$ and $t=3.4$ ($p=0.027$, uncorrected). The five remaining margins are $+0.0513$ over SCALE (95\% CI $[+0.0363,+0.0663]$), $+0.0445$ over Adam-mini ($[+0.0204,+0.0687]$), $+0.0365$ over GaLore ($[+0.0180,+0.0549]$), $+0.0319$ over Muon ($[+0.0296,+0.0342]$), and $+0.0139$ over Lion ($[+0.0089,+0.0189]$); their uncorrected two-sided paired-$t$ $p$-values ($4$ degrees of freedom) range from $<10^{-4}$ (Muon) to $0.0069$ (Adam-mini); the GaLore comparison has $p=0.0054$, so all five remain significant at $0.05$ after Holm correction. The corresponding means are in Table~\ref{tab:roberta_controls_appendix} and Table~\ref{tab:primary-summary}.

SCALE uses the authors' official implementation \citep{glentis2025minimalist}. Its backbone-gradient normalization and parameter-group update rules follow that implementation. During the two-seed search, every rate at or above $1\times10^{-4}$ sent all five tasks toward majority-class solutions; CoLA and RTE selected the smallest rate of the coarse ladder (Table~\ref{tab:repro-params-roberta}). The selected rates were frozen before evaluation. Across the five evaluation seeds, one RTE seed remains degenerate and CoLA spans $14$ points. The all-seed five-task mean is $0.7591$, against $0.8104$ for \bpm{} and $0.7962$ for AdamW.

The cross-entropy control without a local first moment finishes $1.21$ points below AdamW (Table~\ref{tab:qqp-factorial-control}), a smaller drop than SCALE's. Because SCALE also changes gradient normalization, its gap does not isolate the effect of deleting first-moment history. These runs evaluate SCALE after learning-rate tuning for encoder fine-tuning; its original application was language-model pretraining from random initialization. The results compare the methods under the shared encoder fine-tuning protocol.

\clearpage

\subsection{BOM Composition Across Three Optimizers}
\label{app:composition-results}

\begin{figure}[H]
\centering
\def\AdamWTrainScore{(0.5,0.665380) (1.0,0.753944) (1.5,0.777306) (2.0,0.794365) (2.5,0.817640) (3.0,0.836595) (3.5,0.844694) (4.0,0.861778) (4.5,0.878317) (5.0,0.883208)}
\def\AdamWValScore{(0.5,0.660206) (1.0,0.727959) (1.5,0.741593) (2.0,0.752966) (2.5,0.762731) (3.0,0.775093) (3.5,0.778496) (4.0,0.788614) (4.5,0.796169) (5.0,0.796192)}
\def\AdamWTrainLoss{(0.5,0.451899) (1.0,0.404925) (1.5,0.368305) (2.0,0.346177) (2.5,0.324133) (3.0,0.301915) (3.5,0.287626) (4.0,0.261594) (4.5,0.233069) (5.0,0.223524)}
\def\AdamWValLoss{(0.5,0.462945) (1.0,0.424983) (1.5,0.407175) (2.0,0.396604) (2.5,0.392867) (3.0,0.379567) (3.5,0.389328) (4.0,0.373580) (4.5,0.364605) (5.0,0.371398)}
\def\BOMTrainScore{(0.5,0.718658) (1.0,0.772973) (1.5,0.813823) (2.0,0.853717) (2.5,0.869896) (3.0,0.885053) (3.5,0.902995) (4.0,0.909948) (4.5,0.925809) (5.0,0.944658)}
\def\BOMValScore{(0.5,0.696060) (1.0,0.731532) (1.5,0.751292) (2.0,0.781470) (2.5,0.782426) (3.0,0.783050) (3.5,0.789229) (4.0,0.791076) (4.5,0.802291) (5.0,0.810431)}
\def\BOMTrainLoss{(0.5,0.434695) (1.0,0.366599) (1.5,0.320900) (2.0,0.275885) (2.5,0.248989) (3.0,0.219566) (3.5,0.207894) (4.0,0.180658) (4.5,0.169789) (5.0,0.132358)}
\def\BOMValLoss{(0.5,0.448816) (1.0,0.407871) (1.5,0.392868) (2.0,0.364176) (2.5,0.379488) (3.0,0.375405) (3.5,0.401186) (4.0,0.395817) (4.5,0.402081) (5.0,0.382625)}
\def\LionTrainScore{(0.5,0.712364) (1.0,0.763282) (1.5,0.813441) (2.0,0.840602) (2.5,0.868598) (3.0,0.889049) (3.5,0.894137) (4.0,0.907595) (4.5,0.920346) (5.0,0.930248)}
\def\LionValScore{(0.5,0.702502) (1.0,0.737216) (1.5,0.767508) (2.0,0.780384) (2.5,0.789328) (3.0,0.787660) (3.5,0.789354) (4.0,0.798393) (4.5,0.800249) (5.0,0.796536)}
\def\LionTrainLoss{(0.5,0.444329) (1.0,0.379578) (1.5,0.334466) (2.0,0.290964) (2.5,0.253110) (3.0,0.221337) (3.5,0.204650) (4.0,0.171742) (4.5,0.154822) (5.0,0.134043)}
\def\LionValLoss{(0.5,0.458027) (1.0,0.410924) (1.5,0.390639) (2.0,0.372393) (2.5,0.375655) (3.0,0.379757) (3.5,0.406646) (4.0,0.407521) (4.5,0.427160) (5.0,0.433968)}
\def\MuonTrainScore{(0.5,0.609819) (1.0,0.649337) (1.5,0.741599) (2.0,0.785570) (2.5,0.818366) (3.0,0.838196) (3.5,0.853573) (4.0,0.863991) (4.5,0.869152) (5.0,0.883341)}
\def\MuonValScore{(0.5,0.606718) (1.0,0.631309) (1.5,0.708561) (2.0,0.741889) (2.5,0.753093) (3.0,0.763474) (3.5,0.768043) (4.0,0.770605) (4.5,0.772290) (5.0,0.778517)}
\def\MuonTrainLoss{(0.5,0.491724) (1.0,0.462140) (1.5,0.399931) (2.0,0.364641) (2.5,0.318608) (3.0,0.278287) (3.5,0.253161) (4.0,0.231999) (4.5,0.221940) (5.0,0.202894)}
\def\MuonValLoss{(0.5,0.493610) (1.0,0.473064) (1.5,0.446929) (2.0,0.434629) (2.5,0.462177) (3.0,0.444072) (3.5,0.510207) (4.0,0.519603) (4.5,0.584830) (5.0,0.562769)}
\def\ScaleTrainScore{(0.5,0.662805) (1.0,0.697646) (1.5,0.750111) (2.0,0.776070) (2.5,0.789038) (3.0,0.805326) (3.5,0.810352) (4.0,0.820011) (4.5,0.841010) (5.0,0.840889)}
\def\ScaleValScore{(0.5,0.652243) (1.0,0.686732) (1.5,0.720378) (2.0,0.738067) (2.5,0.737144) (3.0,0.750530) (3.5,0.753940) (4.0,0.759064) (4.5,0.763932) (5.0,0.759101)}
\def\ScaleTrainLoss{(0.5,0.507398) (1.0,0.467139) (1.5,0.430276) (2.0,0.389609) (2.5,0.377621) (3.0,0.346718) (3.5,0.336296) (4.0,0.324851) (4.5,0.305279) (5.0,0.288271)}
\def\ScaleValLoss{(0.5,0.512490) (1.0,0.491580) (1.5,0.469550) (2.0,0.449179) (2.5,0.447885) (3.0,0.436776) (3.5,0.447667) (4.0,0.443661) (4.5,0.445597) (5.0,0.440462)}
\def\AdamMiniTrainScore{(0.5,0.627514) (1.0,0.700830) (1.5,0.726423) (2.0,0.744035) (2.5,0.765824) (3.0,0.778671) (3.5,0.770941) (4.0,0.782210) (4.5,0.806336) (5.0,0.796692)}
\def\AdamMiniValScore{(0.5,0.624312) (1.0,0.695712) (1.5,0.719431) (2.0,0.733390) (2.5,0.747128) (3.0,0.753367) (3.5,0.748018) (4.0,0.760606) (4.5,0.768156) (5.0,0.765895)}
\def\AdamMiniTrainLoss{(0.5,0.473535) (1.0,0.446369) (1.5,0.413418) (2.0,0.404136) (2.5,0.379293) (3.0,0.366286) (3.5,0.371436) (4.0,0.352347) (4.5,0.337816) (5.0,0.340618)}
\def\AdamMiniValLoss{(0.5,0.471024) (1.0,0.444776) (1.5,0.420706) (2.0,0.415118) (2.5,0.400026) (3.0,0.393034) (3.5,0.399687) (4.0,0.385526) (4.5,0.382203) (5.0,0.382588)}
\def\GaLoreTrainScore{(0.5,0.608819) (1.0,0.671132) (1.5,0.730044) (2.0,0.746433) (2.5,0.763396) (3.0,0.781648) (3.5,0.778844) (4.0,0.787598) (4.5,0.809033) (5.0,0.813449)}
\def\GaLoreValScore{(0.5,0.612889) (1.0,0.676312) (1.5,0.718020) (2.0,0.732539) (2.5,0.733813) (3.0,0.752194) (3.5,0.749270) (4.0,0.757121) (4.5,0.767509) (5.0,0.773934)}
\def\GaLoreTrainLoss{(0.5,0.474803) (1.0,0.432595) (1.5,0.427459) (2.0,0.395574) (2.5,0.383023) (3.0,0.360726) (3.5,0.357589) (4.0,0.342229) (4.5,0.325346) (5.0,0.312812)}
\def\GaLoreValLoss{(0.5,0.472634) (1.0,0.439572) (1.5,0.448289) (2.0,0.424349) (2.5,0.411758) (3.0,0.404465) (3.5,0.403389) (4.0,0.399991) (4.5,0.394112) (5.0,0.387337)}

\tikzset{
  sAdamW/.style={adamgray,mark=o},
  sBOM/.style={bpmblue,mark=square},
  sLion/.style={lionorange,mark=triangle},
  sMuon/.style={muonpurple,mark=pentagon},
  sScale/.style={scaleolive,mark=asterisk},
  sAdamMini/.style={adamminisky,mark=diamond},
  sGaLore/.style={galoregreen,mark=otimes},
}
\begin{tikzpicture}[baseline=-0.55ex]
\draw[adamgray,line width=0.9pt] (0,0)--(0.38,0); \filldraw[fill=white,draw=adamgray] (0.19,0) circle (1.2pt); \node[anchor=west,font=\scriptsize] at (0.44,0) {AdamW};
\draw[lionorange,line width=0.9pt] (1.75,0)--(2.13,0); \filldraw[fill=white,draw=lionorange] (1.86,-0.06)--(2.02,-0.06)--(1.94,0.09)--cycle; \node[anchor=west,font=\scriptsize] at (2.19,0) {Lion};
\draw[muonpurple,line width=0.9pt] (3.15,0)--(3.53,0); \filldraw[fill=white,draw=muonpurple] (3.34,0) circle (1.4pt); \node[anchor=west,font=\scriptsize] at (3.59,0) {Muon};
\draw[scaleolive,line width=0.9pt] (4.80,0)--(5.18,0); \draw[scaleolive,line width=0.7pt] (4.93,-0.08)--(5.05,0.08) (4.93,0.08)--(5.05,-0.08); \node[anchor=west,font=\scriptsize] at (5.24,0) {SCALE};
\draw[adamminisky,line width=0.9pt] (6.55,0)--(6.93,0); \filldraw[fill=white,draw=adamminisky] (6.66,0)--(6.74,0.09)--(6.82,0)--(6.74,-0.09)--cycle; \node[anchor=west,font=\scriptsize] at (6.99,0) {Adam-mini};
\draw[galoregreen,line width=0.9pt] (8.80,0)--(9.18,0); \filldraw[fill=white,draw=galoregreen] (8.99,0) circle (1.4pt); \draw[galoregreen,line width=0.6pt] (8.92,-0.07)--(9.06,0.07) (8.92,0.07)--(9.06,-0.07); \node[anchor=west,font=\scriptsize] at (9.24,0) {GaLore};
\draw[bpmblue,line width=0.9pt] (10.65,0)--(11.03,0); \filldraw[fill=white,draw=bpmblue] (10.76,-0.08) rectangle (10.92,0.08); \node[anchor=west,font=\scriptsize] at (11.09,0) {\bpm{}-AdamW};
\end{tikzpicture}

\vspace{-3pt}
\begin{minipage}{\textwidth}
\begin{minipage}{0.485\linewidth}
\centering
\begin{tikzpicture}
\begin{axis}[
    width=\linewidth,height=3.2cm,title={(a) Train task score},
    xmin=0.5,xmax=5.0,ymin=0.58,ymax=0.97,
    xtick={1,2,3,4,5},ylabel={Mean score},
    tick label style={font=\scriptsize},label style={font=\scriptsize},title style={font=\small\bfseries},
    grid=major,grid style={line width=.1pt,draw=gray!22},axis line style={draw=gray!55},tick style={draw=gray!55},
    every axis plot/.append style={line width=0.9pt,mark=o,mark size=1.45pt},
]
\addplot[sAdamW,mark options={fill=white}] coordinates {\AdamWTrainScore};
\addplot[sBOM,mark options={fill=white}] coordinates {\BOMTrainScore};
\addplot[sLion,mark options={fill=white}] coordinates {\LionTrainScore};
\addplot[sMuon,mark options={fill=white}] coordinates {\MuonTrainScore};
\addplot[sScale] coordinates {\ScaleTrainScore};
\addplot[sAdamMini,mark options={fill=white}] coordinates {\AdamMiniTrainScore};
\addplot[sGaLore,mark options={fill=white}] coordinates {\GaLoreTrainScore};
\end{axis}
\end{tikzpicture}
\end{minipage}\hfill
\begin{minipage}{0.485\linewidth}
\centering
\begin{tikzpicture}
\begin{axis}[
    width=\linewidth,height=3.2cm,title={(b) Validation task score},
    xmin=0.5,xmax=5.0,ymin=0.58,ymax=0.83,
    xtick={1,2,3,4,5},ylabel={Mean score},
    tick label style={font=\scriptsize},label style={font=\scriptsize},title style={font=\small\bfseries},
    grid=major,grid style={line width=.1pt,draw=gray!22},axis line style={draw=gray!55},tick style={draw=gray!55},
    every axis plot/.append style={line width=0.9pt,mark=o,mark size=1.45pt},
]
\addplot[sAdamW,mark options={fill=white}] coordinates {\AdamWValScore};
\addplot[sBOM,mark options={fill=white}] coordinates {\BOMValScore};
\addplot[sLion,mark options={fill=white}] coordinates {\LionValScore};
\addplot[sMuon,mark options={fill=white}] coordinates {\MuonValScore};
\addplot[sScale] coordinates {\ScaleValScore};
\addplot[sAdamMini,mark options={fill=white}] coordinates {\AdamMiniValScore};
\addplot[sGaLore,mark options={fill=white}] coordinates {\GaLoreValScore};
\end{axis}
\end{tikzpicture}
\end{minipage}

\vspace{1pt}
\begin{minipage}{0.485\linewidth}
\centering
\begin{tikzpicture}
\begin{axis}[
    width=\linewidth,height=3.2cm,title={(c) Train loss},
    xmin=0.5,xmax=5.0,ymin=0.12,ymax=0.53,
    xtick={1,2,3,4,5},xlabel={Epoch},ylabel={Cross-entropy},
    tick label style={font=\scriptsize},label style={font=\scriptsize},title style={font=\small\bfseries},
    grid=major,grid style={line width=.1pt,draw=gray!22},axis line style={draw=gray!55},tick style={draw=gray!55},
    every axis plot/.append style={line width=0.9pt,mark=o,mark size=1.45pt},
]
\addplot[sAdamW,mark options={fill=white}] coordinates {\AdamWTrainLoss};
\addplot[sBOM,mark options={fill=white}] coordinates {\BOMTrainLoss};
\addplot[sLion,mark options={fill=white}] coordinates {\LionTrainLoss};
\addplot[sMuon,mark options={fill=white}] coordinates {\MuonTrainLoss};
\addplot[sScale] coordinates {\ScaleTrainLoss};
\addplot[sAdamMini,mark options={fill=white}] coordinates {\AdamMiniTrainLoss};
\addplot[sGaLore,mark options={fill=white}] coordinates {\GaLoreTrainLoss};
\end{axis}
\end{tikzpicture}
\end{minipage}\hfill
\begin{minipage}{0.485\linewidth}
\centering
\begin{tikzpicture}
\begin{axis}[
    width=\linewidth,height=3.2cm,title={(d) Validation loss},
    xmin=0.5,xmax=5.0,ymin=0.34,ymax=0.60,
    xtick={1,2,3,4,5},xlabel={Epoch},ylabel={Cross-entropy},
    tick label style={font=\scriptsize},label style={font=\scriptsize},title style={font=\small\bfseries},
    grid=major,grid style={line width=.1pt,draw=gray!22},axis line style={draw=gray!55},tick style={draw=gray!55},
    every axis plot/.append style={line width=0.9pt,mark=o,mark size=1.45pt},
]
\addplot[sAdamW,mark options={fill=white}] coordinates {\AdamWValLoss};
\addplot[sBOM,mark options={fill=white}] coordinates {\BOMValLoss};
\addplot[sLion,mark options={fill=white}] coordinates {\LionValLoss};
\addplot[sMuon,mark options={fill=white}] coordinates {\MuonValLoss};
\addplot[sScale] coordinates {\ScaleValLoss};
\addplot[sAdamMini,mark options={fill=white}] coordinates {\AdamMiniValLoss};
\addplot[sGaLore,mark options={fill=white}] coordinates {\GaLoreValLoss};
\end{axis}
\end{tikzpicture}
\end{minipage}
\end{minipage}
\footnotesize
\caption{RoBERTa-base training and validation trajectories for the six baseline optimizers and \bpm{}-AdamW. Each curve averages the five evaluation seeds within each task and then gives the five tasks equal weight. Training scores and losses use the training probe; validation scores use the task-appropriate metric and losses are cross-entropy. The logged intermediate evaluations describe training progress; the reported comparison uses the fixed epoch-5 endpoints in Table~\ref{tab:roberta-control-endpoints}.}
\label{fig:roberta-control-curves}
\end{figure}
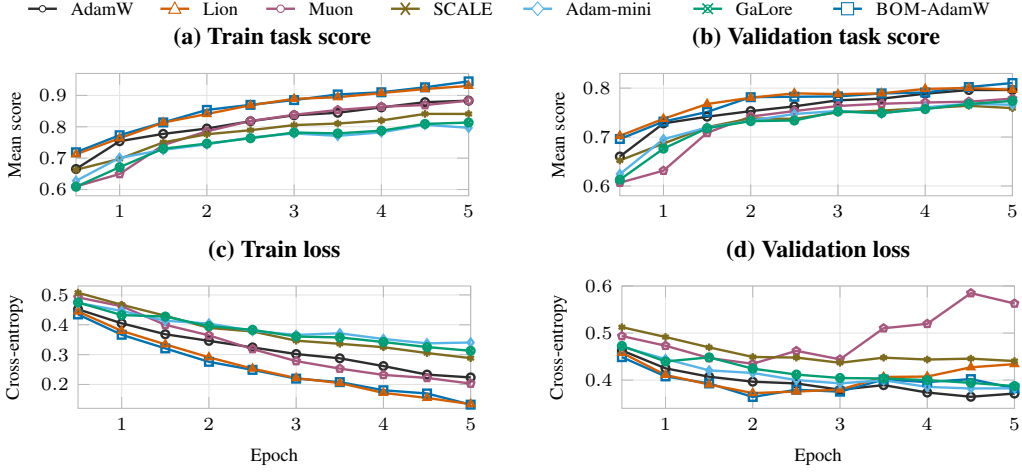

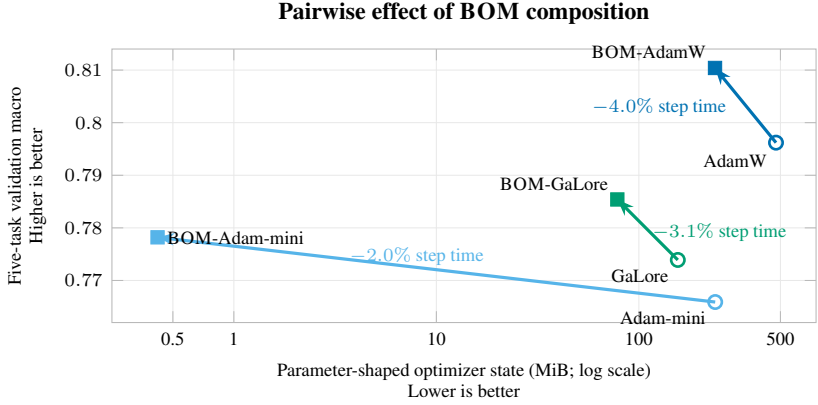
\begin{figure}[H]
\centering
\begin{tikzpicture}
\begin{axis}[
  width=0.78\textwidth,height=5.2cm,
  title={Pairwise effect of \bpm{} composition},
  xlabel={Parameter-shaped optimizer state (MiB; log scale)\\Lower is better},
  ylabel={Five-task validation macro\\Higher is better},
  xmode=log,log basis x=10,
  xmin=0.25,xmax=750,
  ymin=0.762,ymax=0.814,
  xtick={0.5,1,10,100,500},
  xticklabels={0.5,1,10,100,500},
  ytick={0.77,0.78,0.79,0.80,0.81},
  tick label style={font=\scriptsize},label style={font=\scriptsize},
  xlabel style={align=center},ylabel style={align=center},
  title style={font=\small\bfseries},
  axis line style={draw=gray!55},tick style={draw=gray!55},
  xmajorgrids=true,ymajorgrids=true,
  grid style={line width=.1pt,draw=gray!18},
  scaled y ticks=false,
  yticklabel style={/pgf/number format/fixed,/pgf/number format/precision=2},
  clip=false,
]
\addplot[color=bpmblue,very thick,-{Stealth[length=2.2mm]},mark=none]
  coordinates {(475.5,0.7962) (237.7,0.8104)};
\addplot[color=adamminisky,very thick,-{Stealth[length=2.2mm]},mark=none]
  coordinates {(238.2,0.7659) (0.42,0.7782)};
\addplot[color=galoregreen,very thick,-{Stealth[length=2.2mm]},mark=none]
  coordinates {(155.7,0.7739) (78.3,0.7854)};

\addplot[only marks,mark=o,mark size=2.5pt,color=bpmblue,
  mark options={solid,draw=bpmblue,fill=white,line width=1pt}]
  coordinates {(475.5,0.7962)};
\addplot[only marks,mark=o,mark size=2.5pt,color=adamminisky,
  mark options={solid,draw=adamminisky,fill=white,line width=1pt}]
  coordinates {(238.2,0.7659)};
\addplot[only marks,mark=o,mark size=2.5pt,color=galoregreen,
  mark options={solid,draw=galoregreen,fill=white,line width=1pt}]
  coordinates {(155.7,0.7739)};

\addplot[only marks,mark=square*,mark size=2.5pt,color=bpmblue,
  mark options={solid,draw=bpmblue,fill=bpmblue}]
  coordinates {(237.7,0.8104)};
\addplot[only marks,mark=square*,mark size=2.5pt,color=adamminisky,
  mark options={solid,draw=adamminisky,fill=adamminisky}]
  coordinates {(0.42,0.7782)};
\addplot[only marks,mark=square*,mark size=2.5pt,color=galoregreen,
  mark options={solid,draw=galoregreen,fill=galoregreen}]
  coordinates {(78.3,0.7854)};

\node[font=\scriptsize,anchor=north east,yshift=-1pt] at (axis cs:475.5,0.7962) {AdamW};
\node[font=\scriptsize,anchor=south east] at (axis cs:237.7,0.8104) {\bpm{}-AdamW};
\node[font=\scriptsize,anchor=east,text=bpmblue] at (axis cs:300,0.8030) {$-4.0\%$ step time};

\node[font=\scriptsize,anchor=north east] at (axis cs:238.2,0.7659) {Adam-mini};
\node[font=\scriptsize,anchor=west] at (axis cs:0.42,0.7782) {\bpm{}-Adam-mini};
\node[font=\scriptsize,anchor=south,text=adamminisky] at (axis cs:8,0.7712) {$-2.0\%$ step time};

\node[font=\scriptsize,anchor=north east] at (axis cs:155.7,0.7739) {GaLore};
\node[font=\scriptsize,anchor=south east] at (axis cs:78.3,0.7854) {\bpm{}-GaLore};
\node[font=\scriptsize,anchor=west,text=galoregreen] at (axis cs:105,0.7792) {$-3.1\%$ step time};
\end{axis}
\end{tikzpicture}
\footnotesize
\caption{Pairwise effects of substituting \bpm{} for the first-moment component of AdamW, Adam-mini and GaLore on RoBERTa-base. Each arrow starts at the base optimizer (open circle) and ends at its \bpm{} composition (filled square); movement up and left denotes a higher five-task validation macro and less parameter-shaped optimizer state. The annotations give each method's mean paired bf16 step-time change over its main comparison: the three NLP backbones for AdamW, and RoBERTa-base for Adam-mini and GaLore. All three compositions move up and left and have lower mean step time, so each composition Pareto-dominates its own base on the three reported mean estimates. Averaged over the three methods' bf16 main comparisons, adding \bpm{} raises the macro by $1.25$ points and lowers paired step time by $3.0\%$. State bytes use the common accounting convention. Scores are epoch-5 means over five evaluation seeds. Full estimates and paired intervals are in Tables~\ref{tab:composition} and~\ref{tab:memory-unified}.}
\label{fig:roberta-pareto}
\end{figure}

The composition study uses each base optimizer as its own reference, testing incremental savings across distinct update and state structures, including bases that already compress optimizer state.

Per-dataset results for the three pairs are in Table~\ref{tab:roberta_controls_appendix}.

Figure~\ref{fig:roberta-control-curves} gives the seven-method training and validation trajectories alongside Figure~\ref{fig:roberta-pareto}, which summarizes the paired quality, optimizer-state and step-time changes for the three compositions. Detailed fine-tuning memory and timing measurements are grouped with the pretraining resource results in Appendix~\ref{app:system-results}.

\paragraph{GaLore composition and projection.}
The composition improves the mean macro score by $1.14$ points, with a paired five-seed interval that includes zero (Table~\ref{tab:composition}). Its state reduction removes both the projected first moment and the dense first moment in the unprojected branch. The surrogate numerator passes through the base optimizer's unchanged projection, so the retained direction still depends on its current basis. Implementation details are in Appendix~\ref{app:optimizer-implementation}.

Table~\ref{tab:aggregate-effective-convergence} compares the epochs needed to reach the epoch-5 AdamW training-probe macro of $0.8832$. The interpolated \bpm{} crossing is epoch $2.94$. This diagnostic measures progress in the training-probe score; the final validation results and measured step times are reported separately.

\begin{table}[!htbp]
\centering
\setlength{\tabcolsep}{3.5pt}
\caption{Training-probe score progress on RoBERTa-base, macro-averaged over five tasks and five evaluation seeds. The reference is AdamW's epoch-5 training-probe macro of $0.8832$. The first \bpm{} crossing is linearly interpolated between adjacent half-epoch measurements. This diagnostic uses only training-probe scores; the validation evaluations logged every half epoch are shown descriptively in Figure~\ref{fig:roberta-control-curves} and do not enter it, and the reported validation comparison uses the fixed epoch-5 endpoint.}
\label{tab:aggregate-effective-convergence}

\begin{tabular}{lc}
\toprule
Method & Epoch at reference training score \\
\midrule
AdamW & 5.00 \\
\bpm{} & \textbf{2.94} \\
\bottomrule
\end{tabular}
\end{table}

\FloatBarrier

\subsection{Vision Transfer Results}
\label{app:cv-results}
Table~\ref{tab:cv-stl10-check} reports the STL10 transfer endpoints behind the cross-domain comparison of Section~\ref{sec:cross-domain}. 
\begin{table}[!htbp]
\centering
\setlength{\tabcolsep}{3.5pt}
\caption{STL10 transfer results for AdamW and \bpm{} on two pretrained vision
backbones, at the 40-epoch endpoint over the five evaluation seeds. Both arms
share the same 40-epoch budget (Table~\ref{tab:repro-params-vision}) and are
read under the same endpoint convention within each backbone pair. Score is
accuracy; loss is cross-entropy. The paired \bpm{}$-$AdamW mean accuracy differences and descriptive 95\% $t$ intervals are $+0.0122\;[-0.0052,+0.0297]$ for ConvNeXt-Tiny and $+0.0102\;[-0.0093,+0.0298]$ for ViT-Tiny. Intervals use the sample standard deviation of five matched seed differences and $t_{0.975,4}=2.776$, without multiplicity correction.}
\label{tab:cv-stl10-check}

\begin{tabular}{llccccc}
\toprule
Backbone & Method & Metric & Train score & Val score & Train loss & Val loss \\
\midrule
ConvNeXt-Tiny  & AdamW    & Acc & 0.9914 & 0.9647 & 0.0298 & 0.1487 \\
ConvNeXt-Tiny  & \bpm{}   & Acc & 1.0000 & 0.9769 & 0.0007 & 0.0855 \\
ViT-Tiny       & AdamW    & Acc & 0.9930 & 0.9567 & 0.0255 & 0.1711 \\
ViT-Tiny       & \bpm{}   & Acc & 0.9992 & 0.9669 & 0.0015 & 0.1262 \\
\bottomrule
\end{tabular}
\end{table}

\FloatBarrier

\subsection{From-Scratch Pretraining Results}
\label{app:pretraining-results}
\begin{figure}[!htbp]
\centering
\begin{tikzpicture}
\begin{groupplot}[
  group style={group size=4 by 1,horizontal sep=0.60cm},
  scale only axis,
  width=0.19\textwidth,
  height=0.23\textwidth,
  xlabel={Tokens (B)},
  tick label style={font=\scriptsize},
  label style={font=\scriptsize},
  title style={font=\small},
  grid=major,
  grid style={gray!18},
  axis line style={black!65},
  legend style={font=\scriptsize,draw=none,fill=none,at={(0.98,0.98)},anchor=north east},
  legend cell align={left},
  scaled ticks=false,
]
\nextgroupplot[
  title={55M / 3.0B},
  ylabel={Validation loss},
  xmin=0,xmax=3.05,
  ymin=3.45,ymax=4.50,
  xtick={0,1,2,3},
]
\addplot[draw=none,forget plot,name path=adam55upper] table[x=tokens_b,y=upper_loss,col sep=comma]{data_c4_55m_adamw_curve.csv};
\addplot[draw=none,forget plot,name path=adam55lower] table[x=tokens_b,y=lower_loss,col sep=comma]{data_c4_55m_adamw_curve.csv};
\addplot[adamgray!12,forget plot] fill between[of=adam55upper and adam55lower];
\addplot[adamgray,very thick,solid,mark=o,mark repeat=5,mark size=1.2pt] table[x=tokens_b,y=mean_loss,col sep=comma]{data_c4_55m_adamw_curve.csv};
\addlegendentry{AdamW}
\addplot[draw=none,forget plot,name path=bom55upper] table[x=tokens_b,y=upper_loss,col sep=comma]{data_c4_55m_bom_curve.csv};
\addplot[draw=none,forget plot,name path=bom55lower] table[x=tokens_b,y=lower_loss,col sep=comma]{data_c4_55m_bom_curve.csv};
\addplot[bpmblue!14,forget plot] fill between[of=bom55upper and bom55lower];
\addplot[bpmblue,very thick,dashed,mark=square*,mark repeat=5,mark size=1.2pt] table[x=tokens_b,y=mean_loss,col sep=comma]{data_c4_55m_bom_curve.csv};
\addlegendentry{\bpm{}}

\nextgroupplot[
  title={110M / 5.0B},
  xmin=0,xmax=5.08,
  ymin=3.60,ymax=5.08,
  xtick={0,1,2,3,4,5},
]
\addplot[draw=none,forget plot,name path=adam110upper] table[x=tokens_b,y=upper_loss,col sep=comma]{data_c4_110m_adamw_curve.csv};
\addplot[draw=none,forget plot,name path=adam110lower] table[x=tokens_b,y=lower_loss,col sep=comma]{data_c4_110m_adamw_curve.csv};
\addplot[adamgray!12,forget plot] fill between[of=adam110upper and adam110lower];
\addplot[adamgray,very thick,solid,mark=o,mark repeat=20,mark size=1.2pt] table[x=tokens_b,y=mean_loss,col sep=comma]{data_c4_110m_adamw_curve.csv};
\addplot[draw=none,forget plot,name path=bom110upper] table[x=tokens_b,y=upper_loss,col sep=comma]{data_c4_110m_bom_curve.csv};
\addplot[draw=none,forget plot,name path=bom110lower] table[x=tokens_b,y=lower_loss,col sep=comma]{data_c4_110m_bom_curve.csv};
\addplot[bpmblue!14,forget plot] fill between[of=bom110upper and bom110lower];
\addplot[bpmblue,very thick,dashed,mark=square*,mark repeat=20,mark size=1.2pt] table[x=tokens_b,y=mean_loss,col sep=comma]{data_c4_110m_bom_curve.csv};

\nextgroupplot[
  title={440M / 8.5B},
  xmin=0,xmax=8.62,
  ymin=3.00,ymax=5.20,
  xtick={0,2,4,6,8},
]
\addplot[draw=none,forget plot,name path=adam440upper] table[x=tokens_b,y=upper_loss,col sep=comma]{data_c4_440m_adamw_curve.csv};
\addplot[draw=none,forget plot,name path=adam440lower] table[x=tokens_b,y=lower_loss,col sep=comma]{data_c4_440m_adamw_curve.csv};
\addplot[adamgray!12,forget plot] fill between[of=adam440upper and adam440lower];
\addplot[adamgray,very thick,solid,mark=o,mark repeat=34,mark size=1.2pt] table[x=tokens_b,y=mean_loss,col sep=comma]{data_c4_440m_adamw_curve.csv};
\addplot[draw=none,forget plot,name path=bom440upper] table[x=tokens_b,y=upper_loss,col sep=comma]{data_c4_440m_bom_curve.csv};
\addplot[draw=none,forget plot,name path=bom440lower] table[x=tokens_b,y=lower_loss,col sep=comma]{data_c4_440m_bom_curve.csv};
\addplot[bpmblue!14,forget plot] fill between[of=bom440upper and bom440lower];
\addplot[bpmblue,very thick,dashed,mark=square*,mark repeat=34,mark size=1.2pt] table[x=tokens_b,y=mean_loss,col sep=comma]{data_c4_440m_bom_curve.csv};

\nextgroupplot[
  title={1.1B / 15.0B},
  xmin=0,xmax=15.15,
  ymin=2.95,ymax=5.65,
  xtick={0,5,10,15},
]
\addplot[draw=none,forget plot,name path=adam1p1upper] table[x=tokens_b,y=upper_loss,col sep=comma]{data_c4_1p1b_adamw_curve.csv};
\addplot[draw=none,forget plot,name path=adam1p1lower] table[x=tokens_b,y=lower_loss,col sep=comma]{data_c4_1p1b_adamw_curve.csv};
\addplot[adamgray!12,forget plot] fill between[of=adam1p1upper and adam1p1lower];
\addplot[adamgray,very thick,solid,mark=o,mark repeat=60,mark size=1.2pt] table[x=tokens_b,y=mean_loss,col sep=comma]{data_c4_1p1b_adamw_curve.csv};
\addplot[draw=none,forget plot,name path=bom1p1upper] table[x=tokens_b,y=upper_loss,col sep=comma]{data_c4_1p1b_bom_curve.csv};
\addplot[draw=none,forget plot,name path=bom1p1lower] table[x=tokens_b,y=lower_loss,col sep=comma]{data_c4_1p1b_bom_curve.csv};
\addplot[bpmblue!14,forget plot] fill between[of=bom1p1upper and bom1p1lower];
\addplot[bpmblue,very thick,dashed,mark=square*,mark repeat=60,mark size=1.2pt] table[x=tokens_b,y=mean_loss,col sep=comma]{data_c4_1p1b_bom_curve.csv};
\end{groupplot}
\end{tikzpicture}
\caption{C4 validation-loss trajectories at four model and token-budget scales. Curves are means and bands are $\pm1$ sample standard deviation over five consecutive-seed reporting runs at 55M/110M and three at 440M/1.1B. The 1.1B setting uses $10^{-3}$ in both arms. Configurations are in Table~\ref{tab:repro-params-pretraining}; summary losses and perplexities are in Table~\ref{tab:pretraining-results-appendix}.}
\label{fig:c4-scale-losses}
\end{figure}
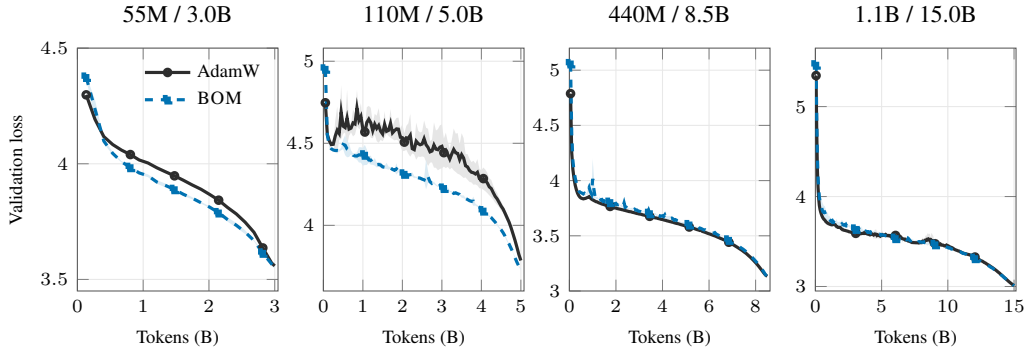

Table~\ref{tab:pretraining-results-appendix} reports aggregate quality readings, and Figures~\ref{fig:pretraining-language-curves} and~\ref{fig:c4-scale-losses} show validation trajectories. Configuration selection, budgets and data construction are specified in Appendix~\ref{app:pretraining-params}. ImageNet reports epoch-90 validation accuracy; all language studies use final evaluation loss. C4-440M and C4-1.1B each use three consecutive-seed reporting runs; every other pretraining study uses five.

\begin{table}[!htbp]
\centering
\setlength{\tabcolsep}{3.5pt}
\caption{Aggregate from-scratch pretraining endpoints, reported as mean $\pm$
sample sd over five consecutive-seed reporting runs except the C4 440M and 1.1B studies ($n=3$ each). $\Delta$ is the
paired mean difference (\bpm{} $-$ AdamW). ImageNet-1k reports epoch-90 top-1 and validation loss. For the language corpora, perplexity is computed as $\exp(\mathrm{loss})$
for each run before averaging. All language rows use the final evaluation. Descriptive paired 95\% intervals for the C4 loss differences are
$[-0.0226,+0.0065]$ at 55M, $[-0.0766,-0.0329]$ at 110M, and $[-0.0199,+0.0111]$ at 440M.
Budgets and schedules are in Table~\ref{tab:repro-params-pretraining}.}
\label{tab:pretraining-results-appendix}
\begin{tabular}{p{80pt}p{80pt}ccc}
\toprule
Setting & Metric & AdamW & \bpm{} & $\Delta$ \\
\midrule
ImageNet-1k, ResNet-50 & Epoch-90 val top-1 $\uparrow$ & $0.7612\pm0.0005$ & $\mathbf{0.7646\pm0.0008}$ & $+0.0035$ \\
                       & Val loss at that epoch $\downarrow$ & $0.9795\pm0.0027$ & $\mathbf{0.9530\pm0.0011}$ & $-0.0265$ \\
\addlinespace
Python code, Qwen3-55M & Final eval loss $\downarrow$ & $1.7353\pm0.0093$ & $\mathbf{1.7337\pm0.0106}$ & $-0.0016$ \\
                       & Perplexity $\downarrow$ & $5.6706\pm0.0529$ & $\mathbf{5.6619\pm0.0603}$ & $-0.0087$ \\
\addlinespace
FineWeb-Edu, Qwen3-55M & Final eval loss $\downarrow$ & $\mathbf{3.2931\pm0.0092}$ & $3.2948\pm0.0045$ & $+0.0017$ \\
                       & Perplexity $\downarrow$ & $\mathbf{26.9262\pm0.2501}$ & $26.9723\pm0.1221$ & $+0.0460$ \\
\addlinespace
C4, Qwen3-55M & Final eval loss $\downarrow$ & $3.5587\pm0.0105$ & $\mathbf{3.5506\pm0.0041}$ & $-0.0081$ \\
                       & Perplexity $\downarrow$ & $35.1192\pm0.3716$ & $\mathbf{34.8346\pm0.1437}$ & $-0.2846$ \\
\addlinespace
C4, Qwen3-110M & Final eval loss $\downarrow$ & $3.7859\pm0.0178$ & $\mathbf{3.7311\pm0.0034}$ & $-0.0547$ \\
                       & Perplexity $\downarrow$ & $44.0805\pm0.7812$ & $\mathbf{41.7270\pm0.1416}$ & $-2.3536$ \\
\addlinespace
C4, Qwen3-440M & Final eval loss $\downarrow$ & $3.1336\pm0.0004$ & $\mathbf{3.1292\pm0.0063}$ & $-0.0044$ \\
                       & Perplexity $\downarrow$ & $22.9555\pm0.0091$ & $\mathbf{22.8549\pm0.1432}$ & $-0.1006$ \\
\addlinespace
C4, Qwen3-1.1B & Final eval loss $\downarrow$ & $3.0106\pm0.0009$ & $\mathbf{3.0085\pm0.0008}$ & $-0.0021$ \\
 & Perplexity $\downarrow$ & $20.3000\pm0.0180$ & $\mathbf{20.2575\pm0.0166}$ & $-0.0425$ 
\\
\bottomrule
\end{tabular}
\end{table}

\begin{figure}[!p]
\centering
\begin{tikzpicture}
\begin{groupplot}[
  group style={group size=2 by 1,horizontal sep=1.0cm},
  width=0.44\textwidth,
  height=0.30\textwidth,
  axis lines=left,
  axis line style={black!55},
  tick align=outside,
  grid=major,
  grid style={black!9},
  xlabel={Processed tokens (billions)},
  xmin=0,
  xmax=3.04,
  xtick={0,0.5,1,1.5,2,2.5,3},
  scaled ticks=false,
  tick label style={font=\small},
  label style={font=\small},
  title style={font=\small},
  yticklabel style={/pgf/number format/fixed,/pgf/number format/precision=2},
  xticklabel style={/pgf/number format/fixed,/pgf/number format/precision=1},
  legend style={font=\small,draw=none,fill=white,fill opacity=0.9,text opacity=1,cells={anchor=west}},
  legend pos=north east
]
\nextgroupplot[title={Python code},ylabel={Validation cross-entropy}]
\addplot[draw=none,forget plot,name path=pythonadamupper]
  table[x=tokens_b,y=adamw_upper,col sep=comma]{data_python_validation_curves.csv};
\addplot[draw=none,forget plot,name path=pythonadamlower]
  table[x=tokens_b,y=adamw_lower,col sep=comma]{data_python_validation_curves.csv};
\addplot[adamgray!12,forget plot]
  fill between[of=pythonadamupper and pythonadamlower];
\addplot[draw=none,forget plot,name path=pythonbomupper]
  table[x=tokens_b,y=bom_upper,col sep=comma]{data_python_validation_curves.csv};
\addplot[draw=none,forget plot,name path=pythonbomlower]
  table[x=tokens_b,y=bom_lower,col sep=comma]{data_python_validation_curves.csv};
\addplot[bpmblue!14,forget plot]
  fill between[of=pythonbomupper and pythonbomlower];
\addplot[adamgray,very thick,solid,mark=o,mark repeat=5,mark size=1.2pt]
  table[x=tokens_b,y=adamw_mean,col sep=comma]{data_python_validation_curves.csv};
\addlegendentry{AdamW}
\addplot[bpmblue,very thick,dashed,mark=square*,mark repeat=5,mark size=1.2pt]
  table[x=tokens_b,y=bom_mean,col sep=comma]{data_python_validation_curves.csv};
\addlegendentry{\bpm{}}

\nextgroupplot[title={FineWeb-Edu}]
\addplot[draw=none,forget plot,name path=finewebadamupper]
  table[x=tokens_b,y=adamw_upper,col sep=comma]{data_fineweb_validation_curves.csv};
\addplot[draw=none,forget plot,name path=finewebadamlower]
  table[x=tokens_b,y=adamw_lower,col sep=comma]{data_fineweb_validation_curves.csv};
\addplot[adamgray!12,forget plot]
  fill between[of=finewebadamupper and finewebadamlower];
\addplot[draw=none,forget plot,name path=finewebbomupper]
  table[x=tokens_b,y=bom_upper,col sep=comma]{data_fineweb_validation_curves.csv};
\addplot[draw=none,forget plot,name path=finewebbomlower]
  table[x=tokens_b,y=bom_lower,col sep=comma]{data_fineweb_validation_curves.csv};
\addplot[bpmblue!14,forget plot]
  fill between[of=finewebbomupper and finewebbomlower];
\addplot[adamgray,very thick,solid,mark=o,mark repeat=5,mark size=1.2pt,forget plot]
  table[x=tokens_b,y=adamw_mean,col sep=comma]{data_fineweb_validation_curves.csv};
\addplot[bpmblue,very thick,dashed,mark=square*,mark repeat=5,mark size=1.2pt,forget plot]
  table[x=tokens_b,y=bom_mean,col sep=comma]{data_fineweb_validation_curves.csv};
\end{groupplot}
\end{tikzpicture}
\caption{Qwen3-55M validation-loss trajectories on Python code and FineWeb-Edu. Each curve is the mean over five runs and each band is $\pm1$ sample standard deviation; validation cross-entropy is evaluated on held-out batches. C4 validation-loss trajectories across four scales are reported separately in Figure~\ref{fig:c4-scale-losses}.}
\label{fig:pretraining-language-curves}
\end{figure}
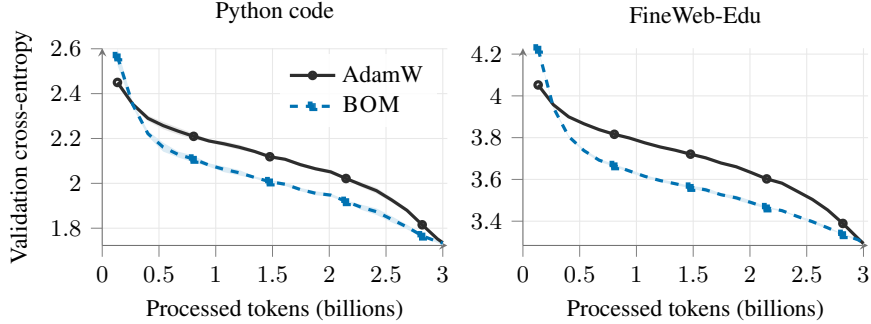

\FloatBarrier

\subsection{RoBERTa Ablations and Diagnostics}
\label{app:roberta-ablations}

Table~\ref{tab:mechanism} summarizes the mechanism controls of Section~\ref{sec:ablations}; the subsections below give their constructions and per-task results.

\begin{table}[H]
\centering
\setlength{\tabcolsep}{3pt}
\renewcommand{\arraystretch}{1.0}
\caption{RoBERTa-base mechanism controls. Most rows report the five-task epoch-5 macro over five seeds and their difference from \bpm{} ($0.8104$); rows 4 and 7 report labeled diagnostics. The classifier-only row uses its protocol-selected configurations; rescue searches are reported in Appendix~\ref{app:classifier-history}. \apm{} is the anchored per-tensor variant.}
\label{tab:mechanism}
\begin{tabularx}{\linewidth}{@{}>{\raggedright\arraybackslash}p{0.22\linewidth}>{\raggedright\arraybackslash}p{0.25\linewidth}>{\centering\arraybackslash}p{0.11\linewidth}>{\centering\arraybackslash}p{0.14\linewidth}>{\raggedright\arraybackslash}X@{}}
\toprule
Question & Control & Reported quantity & Comparison / diagnostic & Reading \\
\midrule
Is it buffer removal alone? & CE + no local $m$ & $0.7841$ & $-2.63$ & no degenerate seeds \\
Is it the objective alone? & Mixed + local $m$ & $0.7806$ & $-2.98$ & below AdamW \\
Is it just the current/history ratio? & AdamW $\beta_1=0.45$ ($0.55/0.45$) & $0.8008$ & $-0.96$ & vs.\ $\beta_1=0.9$: CI includes zero \\
Is historical projection drift real, and where? & $\cos(\hat M_t,h_t)$, head / encoder & $0.81$ / $0.52$ & -- & rel.\ drift $0.46$ / $0.79$ \\
Does a local scalar anchor help? & \apm{}: per-tensor anchor, $+O(L)$ scalars & $0.7934$ & $-1.70$ & lower mean score \\
What changes under current reprojection? & same history, own-step projection & $0.7807$ & $-2.97$ & matched scaling rule; observed drop \\
Is the history itself load-bearing? & $\beta_1=0$ on RTE & $0.5235$ & $4/5$ degenerate & $0/5$ at $\beta_1=0.9$; mixed objective retained \\
Is it the numerator alone? & \bpm{} numerator, CE second moment & $0.7892$ & $-2.12$ & reverse pairing mostly degenerate \\
Is it the full temporal kernel alone? & full-gradient history, $\beta_1=0.9$ & $0.7977$ & $-1.28$ & same kernel; lower mean score \\
Does the past-history term need the backbone? & past history on classifier tensors only & $0.7910$ & $-1.94$ & protocol-selected configurations; RTE seed-sensitive \\
\bottomrule
\end{tabularx}
\end{table}

\subsubsection{Objective--Momentum Factorial Control}
\label{app:qqp-factorial-control}
\begin{table}[!htbp]
\centering
\setlength{\tabcolsep}{3pt}
\caption{Update constructions behind the mechanism comparisons. $g_t$ is the full CE gradient, $u_t=(1-\lambda)g_t+\lambda\bar J_t^\top\hat q_t$ is the \bpm{} numerator, and $\widehat{\operatorname{EMA}}$ includes bias correction. The final column is the vector whose elementwise square feeds the second moment. The own-step buffer averages $\bar J_\tau^\top s_\tau$ and omits historical covariance; the full-kernel buffer averages $g_\tau$ and retains it. Each independently trained arm develops its own trajectory and adaptive statistics.}
\label{tab:history-control-design}
\begin{tabularx}{\textwidth}{l>{\raggedright\arraybackslash}X>{\raggedright\arraybackslash}Xl}
\toprule
Construction & First-order history & Update numerator & Second-moment source \\
\midrule
AdamW & Parameter-space full gradients & $\widehat{\operatorname{EMA}}(g)_t$ & $g_t$ \\
CE + no local $m$ & None & $g_t$ & $g_t$ \\
Mixed + local $m$ & Task residuals and parameter-space mixed updates & $\widehat{\operatorname{EMA}}(u)_t$ & $u_t$ \\
\bpm{} & Task-space residual EMA $\hat q_t$ & $u_t$ & $u_t$ \\
Own-step & Parameter-space mean-residual projections $\hat M_t$ & $(1-\lambda)g_t+\lambda\hat M_t$ & Own numerator \\
Full temporal kernel & Parameter-space full gradients $\hat m_t$ & $(1-\lambda)g_t+\lambda\hat m_t$ & Own numerator \\
Cell B & Task-space residual EMA & $u_t$ & $g_t$ \\
Cell C & Parameter-space gradient EMA; task residual EMA for scale & $\widehat{\operatorname{EMA}}(g)_t$ & $u_t$ \\
\bottomrule
\end{tabularx}
\end{table}

Table~\ref{tab:history-control-design} summarizes the update constructions behind the mechanism comparisons. Table~\ref{tab:qqp-factorial-control} isolates the training objective and parameter-local first-moment storage. Both controls are independently tuned under the common protocol in Appendix~\ref{app:reproducibility}; every condition retains the parameter-local second moment.

\begin{table}[!htbp]
\centering
\setlength{\tabcolsep}{3.5pt}
\caption{RoBERTa-base objective--momentum control. Entries are mean task-appropriate validation scores at epoch 5 over all five evaluation seeds. Both controls receive the same HPO budget as the primary pair; selected rates are in Table~\ref{tab:repro-params-mechanism}. Every condition retains parameter-local adaptive scaling and decoupled weight decay.}
\label{tab:qqp-factorial-control}

\begin{tabular}{llcccc}
\toprule
Dataset & Metric & AdamW & Mixed + local $m$ & CE + no local $m$ & \bpm{} \\
\midrule
CoLA & MCC & 0.5743 & 0.5864 & 0.5635 & \textbf{0.5970} \\
MRPC & F1 & 0.9077 & 0.9016 & 0.9011 & \textbf{0.9191} \\
QQP & F1 & 0.8463 & 0.8452 & 0.8547 & \textbf{0.8626} \\
RTE & Acc & 0.7155 & 0.6318 & 0.7069 & \textbf{0.7350} \\
SST-2 & Acc & 0.9372 & 0.9378 & 0.8945 & \textbf{0.9385} \\
\midrule
Macro mean & -- & 0.7962 & 0.7806 & 0.7841 & \textbf{0.8104} \\
\bottomrule
\end{tabular}
\par\vspace{4pt}
\begin{minipage}{\linewidth}
\textit{Interpretation.} The mixed objective with the parameter-local buffer (\emph{Mixed + local $m$}) reaches $0.7806$ against AdamW's $0.7962$, with the largest drop on RTE. In this arm, the local buffer averages numerators that already contain task-space history, adding parameter-space smoothing to the mixed update. Plain cross-entropy without the buffer (\emph{CE + no local $m$}) removes first-order history and reaches $0.7841$, $1.21$ points below AdamW and $2.63$ below \bpm{}. It remains below \bpm{} on all five tasks; its SST-2 mean is $0.8945$, with no degenerate seed. Under this matched HPO protocol, neither control reproduces \bpm{}'s macro endpoint. SST-2 contributes about $0.88$ of the $2.63$ macro points, approximately $33.5\%$. The all-seed comparison captures training stability together with endpoint quality and evaluates the complete objective--momentum construction.
\end{minipage}
\end{table}

\FloatBarrier

\subsubsection{Fixed-Coefficient Sensitivity and Temporal-Mass Control}
\label{app:coefficient-results}
\paragraph{Mixing coefficient.}
Table~\ref{tab:bpm-coefficient-sensitivity} tests local sensitivity to the fixed mixture. The post-selection protocol is given in Appendix~\ref{app:coefficient-protocol}.

\begin{table}[!htbp]
\centering
\setlength{\tabcolsep}{4.0pt}
\caption{Post-selection local sensitivity of \bpm{}'s fixed mixing coefficient $\lambda$ on RoBERTa-base. Entries are epoch-5 validation mean $\pm$ population standard deviation over the five evaluation seeds; $\Delta$ is relative to the paper configuration $\lambda=0.5,\ \beta_1=0.9$, whose scores are $0.9191$ on MRPC and $0.9385$ on SST-2. Only non-default settings are shown.}
\label{tab:bpm-coefficient-sensitivity}
\begin{tabular}{lcccccc}
\toprule
\shortstack{Varied\\coefficient} & Value & \shortstack{Fixed\\coefficient} & MRPC F1 & $\Delta$ & SST-2 Acc. & $\Delta$ \\
\midrule
$\lambda$ & 0.25 & $\beta_1=0.9$ & $0.9187\pm0.0075$ & $-0.0004$ & $0.9353\pm0.0019$ & $-0.0032$ \\
$\lambda$ & 0.75 & $\beta_1=0.9$ & $0.9156\pm0.0076$ & $-0.0035$ & $0.9385\pm0.0039$ & $0.0000$ \\
\bottomrule
\end{tabular}%

\par\vspace{4pt}\begin{minipage}{\linewidth}
The largest observed mean shift is $0.35$ points within the tested $\lambda$ neighborhood at $\beta_1=0.9$. These fixed-configuration results support local robustness of the mixture on the two tested tasks.
\end{minipage}
\end{table}

\paragraph{Task-space EMA decay: the full sweep.}
Table~\ref{tab:beta-sweep} varies $\beta_1$ alone over $\{0,\,0.8,\,0.9,\,0.99\}$ on all five RoBERTa-base tasks, holding the $0.5/0.5$ mixture, \bpm{}'s selected learning rates and every other coefficient fixed, so that differences isolate the effect of $\beta_1$. At $\beta_1=0$ the bias-corrected $\hat q_t$ reduces to the current batch-mean residual $s_t$, so the mixed objective is retained but no history is kept. The $\beta_1=0.9$ row is the anchor for this diagnostic sweep.

\begin{table}[!htbp]
\centering
\setlength{\tabcolsep}{4.0pt}
\caption{Task-space EMA decay $\beta_1$ across the five RoBERTa-base tasks, with the $0.5/0.5$ current/history mixture, the selected learning rates and all remaining coefficients held fixed. Entries are epoch-5 validation means over the five evaluation seeds; \emph{deg.} counts seeds whose endpoint sits at the majority-class solution ($0.5271$ on RTE, $0.5092$ on SST-2) with validation cross-entropy at chance, and no seed of any other task is degenerate at any setting. The $\beta_1=0.9$ row is the sweep anchor; the best entry in each column is bold.}
\label{tab:beta-sweep}
\begin{tabular}{lcccccccc}
\toprule
& \multicolumn{2}{c}{RTE (Acc.)} & MRPC (F1) & \multicolumn{2}{c}{SST-2 (Acc.)} & CoLA (MCC) & QQP (F1) \\
\cmidrule(lr){2-3}\cmidrule(lr){4-4}\cmidrule(lr){5-6}\cmidrule(lr){7-7}\cmidrule(lr){8-8}
$\beta_1$ & score & deg. & score & score & deg. & score & score \\
\midrule
$0.00$ & $0.5235$ & 4/5 & $0.9057$ & $0.8505$ & 1/5 & $0.5848$ & $0.8590$ \\
$0.80$ & $0.6383$ & 2/5 & $\mathbf{0.9210}$ & $0.9378$ & 0/5 & $0.5890$ & $0.8581$ \\
$0.90$ & $0.7350$ & 0/5 & $0.9191$ & $\mathbf{0.9385}$ & 0/5 & $\mathbf{0.5970}$ & $\mathbf{0.8626}$ \\
$0.99$ & $\mathbf{0.7437}$ & 0/5 & $0.9149$ & $0.9358$ & 0/5 & $0.5897$ & $0.8529$ \\
\bottomrule
\end{tabular}
\par\vspace{4pt}\begin{minipage}{\linewidth}
CoLA, MRPC and QQP stay within $1.5$ points of the sweep anchor at every setting. SST-2 has one degenerate seed at $\beta_1=0$; its four remaining seeds average $0.9358$, while the table retains all five seeds in its reported mean. RTE shows the strongest dependence: its endpoint rises monotonically with $\beta_1$, and the number of seeds reaching the majority-class solution falls from four of five at $\beta_1=0$ to none at $0.9$. Degeneracy therefore occurs on both RTE and SST-2, with the larger effect on RTE. Every cell uses \bpm{}'s selected task-specific learning rates; $\beta_1=0.9$ and $0.99$ differ by at most one point on every task.
\end{minipage}
\end{table}

\paragraph{AdamW-only current/history ratio.}
The mass-matched control tests whether changing AdamW's current/history total weights explains the gain. Settings are specified in Appendix~\ref{app:coefficient-protocol}.

\begin{table}[!htbp]
\centering
\setlength{\tabcolsep}{4.0pt}
\caption{AdamW-only current/history-ratio control on RoBERTa-base. The main-experiment AdamW configuration is unchanged except for $\beta_1$, which moves from $0.9$ to $0.45$ and changes the asymptotic current/past total weights from $0.10/0.90$ to $0.55/0.45$. This matches the mass split of \bpm{}'s batch-mean component, while the past tail decays at $0.45$ rather than $0.9$. Per-dataset entries are epoch-5 means over the five evaluation seeds; the macro row averages the five task means. Score is the task-appropriate validation metric, loss is validation cross-entropy, and $\Delta$ is $\beta_1=0.45$ minus $\beta_1=0.9$.}
\label{tab:adamw-momentum-ratio-control}
\begin{tabular}{llrrr}
\toprule
\multicolumn{5}{l}{\textit{Validation score}} \\
Dataset & Metric & $\beta_1=0.9$ & $\beta_1=0.45$ & $\Delta$ \\
\midrule
CoLA & MCC & $0.5743$ & $0.5827$ & $+0.0083$ \\
MRPC & F1 & $0.9077$ & $0.9039$ & $-0.0038$ \\
QQP & F1 & $0.8463$ & $0.8577$ & $+0.0115$ \\
RTE & Acc. & $0.7155$ & $0.7191$ & $+0.0036$ \\
SST-2 & Acc. & $0.9372$ & $0.9408$ & $+0.0037$ \\
\midrule
Macro & -- & $0.7962$ & $0.8008$ & $+0.0047$ \\
\bottomrule
\end{tabular}
\par\medskip
\begin{tabular}{llrrr}
\toprule
\multicolumn{5}{l}{\textit{Validation loss}} \\
Dataset & Metric & $\beta_1=0.9$ & $\beta_1=0.45$ & $\Delta$ \\
\midrule
CoLA & MCC & $0.4777$ & $0.4781$ & $+0.0004$ \\
MRPC & F1 & $0.3240$ & $0.3632$ & $+0.0392$ \\
QQP & F1 & $0.2683$ & $0.2558$ & $-0.0126$ \\
RTE & Acc. & $0.5894$ & $0.6297$ & $+0.0402$ \\
SST-2 & Acc. & $0.1975$ & $0.1912$ & $-0.0063$ \\
\midrule
Macro & -- & $0.3714$ & $0.3836$ & $+0.0122$ \\
\bottomrule
\end{tabular}%

\par\vspace{4pt}
\begin{minipage}{\linewidth}
\textit{Interpretation.} Matching the current/past mass split raises AdamW's validation-score macro from $0.7962$ to $0.8008$ (paired 95\% CI for the change, $[-0.11,+1.04]$ points), while macro validation loss rises from $0.3714$ to $0.3836$. It does not reproduce \bpm{}'s $0.8104$ endpoint. The past tail still decays at $0.45$ rather than $0.9$, so this control matches total mass rather than the complete temporal kernel; Table~\ref{tab:full-kernel} tests the latter. Settings are in Appendix~\ref{app:coefficient-protocol}.
\end{minipage}
\end{table}

\FloatBarrier

\subsubsection{Full Temporal-Kernel Parameter-History Control}
\label{app:full-kernel}
The mass-ratio control changes the past-tail decay as well as the current/history split. To match the complete temporal kernel, we instead retain the full cross-entropy gradient history:
\begin{align}
 m_t &= \beta_1 m_{t-1}+(1-\beta_1)g_t,\qquad \hat m_t=m_t/(1-\beta_1^t),\quad m_0=0,\\
 u_t^{\mathrm{FK}} &= (1-\lambda)g_t+\lambda\hat m_t,\qquad
 v_t=\beta_2v_{t-1}+(1-\beta_2)(u_t^{\mathrm{FK}})^{\odot2}.
\end{align}
The update uses the bias-corrected second moment and decoupled weight decay, without another first-moment EMA on $u_t^{\mathrm{FK}}$. With $\lambda=0.5$ and $\beta_1=0.9$, the current coefficient is $0.5+0.05/(1-0.9^t)$ and the coefficient of lag $k\geq1$ is $0.05\,0.9^k/(1-0.9^t)$. Thus it matches the temporal kernel of \bpm{}'s batch-mean component, including startup correction, while retaining full-gradient parameter-space history. Unlike the own-step control, this history also retains historical example-specific covariance. Each independently trained arm accumulates its second moment from its own numerator.

The control is independently tuned under the shared HPO policy; its final settings and history-buffer precision are recorded in Appendix~\ref{app:diagnostic-protocols}.

\begin{table}[!htbp]
\centering
\setlength{\tabcolsep}{3pt}
\caption{Full temporal-kernel control on RoBERTa-base under the unified HPO policy (Appendix~\ref{app:diagnostic-protocols}). Scores are five-seed means from the one-shot final test after epoch 5. $\Delta$ is \bpm{} minus the full-kernel control; intervals are descriptive paired 95\% $t$ intervals ($df=4$), without multiplicity correction. The macro interval averages tasks within each seed before forming paired differences.}
\label{tab:full-kernel}
\begin{tabular}{llrrrrr}
\toprule
Task & Metric & AdamW & Full kernel & \bpm{} & $\Delta$ & 95\% CI \\
\midrule
CoLA & MCC & 0.5743 & 0.5625 & 0.5970 & $+0.0345$ & $[+0.0053,+0.0636]$ \\
MRPC & F1 & 0.9077 & 0.9209 & 0.9191 & $-0.0018$ & $[-0.0159,+0.0122]$ \\
QQP & F1 & 0.8463 & 0.8584 & 0.8626 & $+0.0042$ & $[-0.0030,+0.0114]$ \\
RTE & Acc. & 0.7155 & 0.7235 & 0.7350 & $+0.0116$ & $[-0.0376,+0.0607]$ \\
SST-2 & Acc. & 0.9372 & 0.9232 & 0.9385 & $+0.0154$ & $[-0.0115,+0.0423]$ \\
Macro & -- & 0.7962 & 0.7977 & 0.8104 & $+0.0128$ & $[+0.0066,+0.0190]$ \\
\bottomrule
\end{tabular}
\par\vspace{4pt}
\begin{minipage}{\linewidth}
The full-kernel macro is $0.7977$, compared with AdamW's $0.7962$, the own-step control's $0.7807$, and \bpm{}'s $0.8104$. \bpm{} has a higher mean on four tasks; MRPC favors the full-kernel control. All 25 full-kernel endpoints have nondegenerate predictions. These results disfavor temporal weighting alone as an explanation for the observed macro gain under this protocol; the descriptive paired macro interval excludes zero.
\end{minipage}
\end{table}

\FloatBarrier

\subsubsection{Historical-Projection Control}
\label{app:stale-control}
The control of Section~\ref{sec:ablations} keeps \bpm{}'s task-space accumulation but projects each signal through the model and batch-mean Jacobian of its own step, maintaining a parameter-shaped $M_t=\beta_1 M_{t-1}+(1-\beta_1)\bar J_{\ell,t}^\top s_t$ and forming $u_{\ell,t}=(1-\lambda)g_{\ell,t}+\lambda\hat M_t$. Both arms share $\lambda$, $\beta_1$, $\beta_2$, $\epsilon$, weight decay, and the AdamW-style adaptive-scaling construction; each updates its scale statistic from its own numerator. Evaluated on a common trajectory, the two numerator constructions differ by $\lambda(\hat M_t-\bar J_{\ell,t}^\top\hat q_t)$, the bias-corrected historical-projection drift, and neither accumulates covariance in its history. On the classifier bias, $\bar J_{b,t}=I$ implies $\hat M_t=\hat q_t$ within that trajectory, so both constructions reduce to $(1-\lambda)s_t+\lambda\hat q_t$. This is equality of update formulas: independently trained arms can develop different residual histories and scale statistics even when their selected learning rates agree. The own-step control uses the shared independent HPO protocol (Appendix~\ref{app:reproducibility}). It is a diagnostic requiring an extra backward pass and a dense buffer, so its step time and state are not compared with the compositions. For a linear model with a fixed batch, the Jacobian is constant and the two rules coincide when initialized identically and run at the same learning rate.

\begin{table}[!htbp]
\centering
\setlength{\tabcolsep}{4pt}
\caption{Historical-projection control on RoBERTa-base, evaluated once after
five epochs on each of the five evaluation seeds under the final-test convention
of Appendix~\ref{app:reproducibility}. The control keeps \bpm{}'s task-space history but transports it through
each step's own model and examples rather than the current ones. On a common trajectory, its numerator differs
from \bpm{}'s by $\lambda$ times the bias-corrected historical-projection drift term of
Eq.~(\ref{eq:jacobian-drift}). Both arms share the coefficients, temporal
weighting, covariance treatment, AdamW-style adaptive-scaling rule and weight
decay; each updates its scale statistic from its own numerator.
Its independently selected per-task learning rates are listed in Table~\ref{tab:repro-params-roberta}.
$\Delta$ is the mean paired \bpm{}-minus-control score difference. Intervals are descriptive,
unadjusted 95\% paired-$t$ intervals over five matched evaluation runs ($df=4$), using the
sample standard deviation of the paired differences. Macro differences first average the five
tasks equally within each matched run before estimating uncertainty.}
\label{tab:stale-control}
\begin{tabular}{llccccc}
\toprule
Dataset & Metric & AdamW & Own-step control & \bpm{} & $\Delta$ & Paired 95\% CI \\
\midrule
CoLA  & MCC & 0.5743 & 0.5632 & \textbf{0.5970} & $+0.0338$ & $[-0.1003,+0.1678]$ \\
MRPC  & F1  & 0.9077 & 0.9162 & \textbf{0.9191} & $+0.0029$ & $[-0.0113,+0.0171]$ \\
QQP   & F1  & 0.8463 & 0.8561 & \textbf{0.8626} & $+0.0065$ & $[-0.0044,+0.0173]$ \\
RTE   & Acc & 0.7155 & 0.6440 & \textbf{0.7350} & $+0.0910$ & $[+0.0161,+0.1658]$ \\
SST-2 & Acc & 0.9372 & 0.9241 & \textbf{0.9385} & $+0.0144$ & $[+0.0037,+0.0252]$ \\
\midrule
Macro mean & -- & 0.7962 & 0.7807 & \textbf{0.8104} & $+0.0297$ & $[+0.0014,+0.0580]$ \\
\bottomrule
\end{tabular}
\par\vspace{4pt}
\begin{minipage}{\linewidth}
\textit{Interpretation.} The independently tuned current-reprojection endpoint is $2.97$ macro points above the own-step control ($\Delta=+0.0297$, descriptive paired 95\% CI $[+0.0014,+0.0580]$). RTE contributes $0.0910/5=0.0182$ to the macro difference, approximately $61.2\%$ of the total. The macro interval and the RTE and SST-2 task intervals exclude zero; the other three task intervals include zero. These intervals are unadjusted for multiple comparisons. The control also lands $1.55$ points below AdamW on the macro, whereas \bpm{} finishes above it. The matched-HPO own-step--\bpm{} pair compares historical and current projection under a common update family.
\end{minipage}
\end{table}

\FloatBarrier

\subsubsection{Measuring historical projection drift}
\label{app:drift-measurement}

The control of Appendix~\ref{app:stale-control} maintains both projection accumulations on one trajectory. At each telemetry step, $\hat M_t$ averages the task-space signals after their own-step projections, while $h_t=\bar J_t^\top\hat q_t$ reprojects the same residual history through the current model and examples. The latter is \bpm{}'s historical component, which enters its numerator as $(1-\lambda)g_t+\lambda h_t$. Their difference is the bias-corrected historical-projection drift vector on the same parameter and batch sequence, with no matching across runs required.

Table~\ref{tab:drift} reports the comparison split by depth over the five evaluation seeds, using telemetry steps from step 100 to the epoch-5 endpoint. Each task is weighted equally rather than each record, since the number of telemetry steps scales with dataset size.

The own-step accumulation stays closer to current reprojection in the classifier head than in the encoder: mean cosine $0.81$ against $0.52$, and relative drift $\lVert\hat M_t-h_t\rVert/\lVert\hat M_t\rVert$ of $0.46$ against $0.79$. These averages over the retained telemetry steps show a substantial encoder departure. The compact history remains $d_{\mathrm{out}}$-dimensional while the current model and examples determine its parameter-space projection at every step.

The last column measures within-batch cancellation. The ratio $\lVert s_t\rVert / B^{-1}\sum_i\lVert r_{t,i}\rVert$ is the fraction of the average per-example residual magnitude retained by batch averaging; complete cancellation would make it zero. Its task means exceed $0.13$ and average $0.25$. This quantifies the magnitude that survives within each batch; the projection-drift statistics above measure how the resulting historical direction changes across steps.

\begin{table}[!htbp]
\centering
\caption{Historical-projection drift measured inside the own-step control on RoBERTa-base, over the five evaluation seeds and all telemetry steps from step 100 to the epoch-5 endpoint. $\hat M_t$ accumulates the task-space signals through each step's own model and examples; $h_t=\bar J_t^\top \hat q_t$ reprojects the same history through the current model and examples. \emph{head} is the classification head, \emph{encoder} every other trainable tensor. Cancellation is $\lVert s_t\rVert / B^{-1}\sum_i \lVert r_{t,i}\rVert$. Tasks are weighted equally in the mean.}
\label{tab:drift}
\begin{tabular}{lccccc}
\toprule
& \multicolumn{2}{c}{$\cos(\hat M_t, h_t)$} & \multicolumn{2}{c}{$\lVert \hat M_t-h_t\rVert/\lVert \hat M_t\rVert$} & Cancellation \\
\cmidrule(lr){2-3}\cmidrule(lr){4-5}\cmidrule(lr){6-6}
Task & head & encoder & head & encoder & ratio \\
\midrule
CoLA  & $0.84$ & $0.46$ & $0.45$ & $0.86$ & $0.27$ \\
MRPC  & $0.63$ & $0.26$ & $0.71$ & $0.96$ & $0.28$ \\
QQP   & $0.81$ & $0.58$ & $0.47$ & $0.75$ & $0.25$ \\
RTE   & $0.98$ & $0.84$ & $0.20$ & $0.50$ & $0.14$ \\
SST-2 & $0.79$ & $0.44$ & $0.49$ & $0.88$ & $0.33$ \\
\midrule
Mean  & $0.81$ & $0.52$ & $0.46$ & $0.79$ & $0.25$ \\
\bottomrule
\end{tabular}
\end{table}

\paragraph{What the drift is made of.}
The difference $\bar J_\tau-\bar J_t$ in Eq.~(\ref{eq:jacobian-drift}) changes for two reasons at once: the parameters have moved, and the samples defining the mean Jacobian have been replaced. Table~\ref{tab:drift-decomposition} separates them on \bpm{} trajectories for all five RoBERTa tasks with two fixed probe batches and a fixed task-space direction, so that each single-factor comparison varies exactly one source. At lag 20, the mixed encoder cosine lies between $0.456$ and $0.577$ on every task. RTE and MRPC show substantial departure from both parameter movement (cosine $0.702$--$0.740$) and sample replacement ($0.655$--$0.686$). On CoLA, SST-2 and QQP, sample replacement is larger over this horizon ($0.475$--$0.619$) than parameter movement ($0.830$--$0.850$); at longer lags the parameter-only cosine decreases consistently. The measured term is therefore historical-projection drift jointly induced by the model that expressed a signal and the examples on which it was expressed. Current reprojection removes both components together, while the fixed probes quantify their respective contributions to historical-projection drift.

\begin{table}[!htbp]
\centering
\setlength{\tabcolsep}{3.5pt}
\caption{Decomposing historical-projection drift on \bpm{} trajectories across the five RoBERTa-base tasks (the five evaluation seeds, probes every 20 steps from step 100 to the epoch-5 endpoint). Fixed probe batches $A$ and $B$ and a fixed unit task-space direction $s^{\ast}$ define $v_X(\theta)=\bar J_X(\theta)^{\top}s^{\ast}$. \emph{Batch-only} compares $v_A$ and $v_B$ at the same parameters; \emph{parameter-only} compares $v_A$ at parameters $\Delta$ steps apart; \emph{mixed} changes both. Entries are encoder cosines averaged over seeds and probe steps; lower values indicate larger departure. The EMA at $\beta_1=0.9$ has a mean lag of nine steps.}
\label{tab:drift-decomposition}
\begin{tabular}{lcccccc}
\toprule
Task & Batch-only & \shortstack{Param.\\$\Delta=20$} & \shortstack{Param.\\$\Delta=100$} & \shortstack{Param.\\$\Delta=300$} & \shortstack{Mixed\\$\Delta=20$} & \shortstack{Mixed\\$\Delta=300$} \\
\midrule
RTE   & $0.686$ & $0.740$ & $0.588$ & $0.349$ & $0.577$ & $0.294$ \\
MRPC  & $0.655$ & $0.702$ & $0.579$ & $0.493$ & $0.545$ & $0.432$ \\
CoLA  & $0.619$ & $0.830$ & $0.704$ & $0.622$ & $0.559$ & $0.480$ \\
SST-2 & $0.548$ & $0.850$ & $0.736$ & $0.656$ & $0.507$ & $0.432$ \\
QQP   & $0.475$ & $0.848$ & $0.750$ & $0.672$ & $0.456$ & $0.402$ \\
\bottomrule
\end{tabular}

\end{table}

\FloatBarrier

\subsubsection{Classifier-Only Historical Contribution}
\label{app:classifier-history}
This control tests whether the mechanism can be reduced to an output-level correction. Let $\Gamma_{\mathrm{cls}}$ retain the four weight and bias tensors under the RoBERTa classifier and set every other parameter coordinate to zero. The control restricts only the past-history contribution to those tensors:
\begin{equation}
u_t^{\mathrm{cls}}
=(1-\lambda)g_t
+\lambda\frac{1-\beta_1}{1-\beta_1^t}\bar J_t^\top s_t
+\lambda\frac{\beta_1}{1-\beta_1^t}\Gamma_{\mathrm{cls}}\bar J_t^\top q_{t-1}.
\label{eq:classifier-history-control}
\end{equation}
The backbone remains trainable, and both the full-network supervised gradient and the current-step task-space term are unchanged. Each arm accumulates its parameter-local second moment from its own numerator. Training settings and the primary and rescue selection procedures are specified in Appendix~\ref{app:diagnostic-protocols}.

\begin{table}[!htbp]
\centering
\setlength{\tabcolsep}{3.0pt}
\caption{Classifier-only past-history control on RoBERTa-base. Each row retains all five evaluation runs (A--E) at epoch 5. Score is Matthews correlation for CoLA, F1 for MRPC/QQP and accuracy for RTE/SST-2; loss is cross-entropy. \emph{Primary} marks the protocol-selected configurations used in the primary macro and \emph{Rescue} the configurations selected by the three-HPO-seed rescue searches. Selection rules are in Appendix~\ref{app:diagnostic-protocols}.}
\label{tab:classifier-history-results}
\begin{tabular}{llrrrrrrrl}
\toprule
Task & LR & A & B & C & D & E & Mean score & Mean loss & Status \\
\midrule
CoLA  & $2.5\times10^{-5}$ & 0.6407 & 0.4624 & 0.6547 & 0.5417 & 0.6007 & 0.5800 & 0.4775 & Primary \\
MRPC  & $2.0\times10^{-5}$ & 0.9242 & 0.9078 & 0.9158 & 0.8538 & 0.9154 & 0.9034 & 0.3237 & Primary \\
MRPC  & $3.0\times10^{-5}$ & 0.7722 & 0.9154 & 0.9007 & 0.9061 & 0.9144 & 0.8818 & 0.4517 & Rescue \\
QQP   & $3.0\times10^{-5}$ & 0.8633 & 0.8561 & 0.8631 & 0.8627 & 0.8292 & 0.8549 & 0.2624 & Primary \\
QQP   & $2.0\times10^{-5}$ & 0.8537 & 0.8440 & 0.8588 & 0.8396 & 0.8556 & 0.8503 & 0.2694 & Rescue \\
RTE   & $2.0\times10^{-5}$ & 0.5632 & 0.7365 & 0.7401 & 0.7978 & 0.5740 & 0.6823 & 0.5960 & Primary \\
RTE   & $1.5\times10^{-5}$ & 0.7545 & 0.7076 & 0.7148 & 0.7112 & 0.6354 & 0.7047 & 0.6008 & Rescue \\
RTE   & $1.0\times10^{-5}$ & 0.6751 & 0.5560 & 0.6751 & 0.6715 & 0.5993 & 0.6354 & 0.6413 & Rescue \\
SST-2 & $3.0\times10^{-5}$ & 0.9346 & 0.9381 & 0.9392 & 0.9415 & 0.9186 & 0.9344 & 0.1939 & Primary \\
\bottomrule
\end{tabular}
\par\vspace{4pt}
\begin{minipage}{\linewidth}
\textit{Selection and interpretation.} The primary score uses the protocol-selected configurations. Rescue searches on MRPC (one), QQP (one) and RTE (three, yielding two distinct alternative rates) use a three-HPO-seed mean; none replaces a primary configuration (Appendix~\ref{app:diagnostic-protocols}). Full \bpm{} exceeds the protocol-selected control on all five task means and by $1.94$ macro points (descriptive paired 95\% CI $[-0.85,+4.74]$). The rescue configurations do not recover full \bpm{} performance. No selected endpoint predicts only one class: QQP has one weaker run, and two RTE runs have near-majority predictions while still using both classes. Across the tested tasks, these results are consistent with a contribution from full-network history.
\end{minipage}
\end{table}

\FloatBarrier

\subsubsection{Numerator and Second Moment}
\label{app:numerator-preconditioner}
\bpm{} replaces two things at once relative to \adamw{}: the numerator handed to the update
map, and the tensor whose elementwise square accumulates into $v_{\ell,t}$.
Table~\ref{tab:numerator-preconditioner} reports the two off-diagonal cells of that
factorial. Cell~B keeps \bpm{}'s numerator $u_{\ell,t}$ but builds $v_{\ell,t}$ from the
plain cross-entropy gradient; cell~C hands \adamw{}'s first-moment numerator a second moment
built from $u_{\ell,t}$. Both need $\nabla_{\theta}\mathcal{L}^{\mathrm{CE}}_t$ and
$\nabla_{\theta}\mathcal{L}^{\mathrm{mix}}_t$ in the same step, so they take one forward pass
and two backward passes; like the control of Appendix~\ref{app:stale-control} they are
diagnostics rather than proposed optimizers.

These cross-pairings include baseline-rate checks and targeted learning-rate searches, whose candidates and selection rule are specified in Appendix~\ref{app:diagnostic-protocols}. Changing the source of $v_{\ell,t}$ changes per-coordinate effective step sizes, so a shared nominal learning rate does not hold update magnitudes fixed.

\begin{table}[!htbp]
\centering
\setlength{\tabcolsep}{3.5pt}
\caption{Numerator/second-moment cross-pairings on RoBERTa-base. Entries are epoch-5 means over all five evaluation seeds, including degenerate runs. Cell~B combines \bpm{}'s numerator with the CE-gradient second moment; cell~C combines AdamW's numerator with the \bpm{}-numerator second moment. Search rules and final rates are in Appendix~\ref{app:diagnostic-protocols} and Table~\ref{tab:repro-params-mechanism}.}
\label{tab:numerator-preconditioner}
\begin{tabular}{lccccc}
\toprule
& & & \multicolumn{2}{c}{\shortstack{Cell~B: \bpm{} numerator\\CE second moment}} & Cell~C \\
\cmidrule(lr){4-5}\cmidrule(lr){6-6}
Dataset & \adamw{} & \bpm{} & at \bpm{}'s rate & at $3\times10^{-5}$ & \shortstack{\adamw{} numerator\\\bpm{} $v$} \\
\midrule
CoLA  & 0.5743 & \textbf{0.5970} & 0.5501 & 0.5565 & 0.1146 \\
MRPC  & 0.9077 & \textbf{0.9191} & 0.9134 & 0.9134 & 0.8551 \\
QQP   & 0.8463 & \textbf{0.8626} & 0.8427 & 0.8427 & 0.5175 \\
RTE   & 0.7155 & \textbf{0.7350} & 0.6578 & 0.6968 & 0.5747 \\
SST-2 & 0.9372 & \textbf{0.9385} & 0.9365 & 0.9365 & 0.6716 \\
\midrule
Macro mean & 0.7962 & \textbf{0.8104} & 0.7801 & 0.7892 & 0.5467 \\
$\Delta$ vs.\ \bpm{} & $-0.0142$ & -- & $-0.0304$ & $-0.0212$ & $-0.2637$ \\
\bottomrule
\end{tabular}
\par\vspace{4pt}
\begin{minipage}{\linewidth}
\textit{Targeted searches.} Cell~B's targeted-search column changes only CoLA and RTE and reuses MRPC, QQP and SST-2; cell~C includes an additional CoLA stability screen. These targeted searches are separate from the common optimizer-comparison budget.

\textit{Interpretation.} All 25 cell~B runs
finish without degenerate endpoints. At $3\times10^{-5}$, cell~B ends $2.12$ macro
points below \bpm{} and $0.70$ below AdamW. Cell~C has 15 majority-class endpoints
among its 25 runs (CoLA 4/5, MRPC 3/5, QQP 2/5, RTE 3/5, SST-2 3/5), leaving its
all-seed macro $26.4$ points below \bpm{}. On CoLA, the additional two-seed screening
returns Matthews correlation $0.0000$ at all eight rates from $1\times10^{-6}$ to
$5\times10^{-5}$. The complete \bpm{} construction outperforms these tested cross-pairings, supporting the joint use of its numerator and adaptive scale statistic.
\end{minipage}
\end{table}

\FloatBarrier

\subsubsection{Strict Output-Only Second-Moment Diagnostic}
\label{app:out-rms}
The strict output root-mean-square (OutRMS) variant replaces parameter-space $v_{\ell,t-1}$ with a task-space statistic. For per-example residuals $r_{t,i}$ it stores $R_t=\beta_2R_{t-1}+(1-\beta_2)B^{-1}\sum_i r_{t,i}^{\odot2}$ and uses $\hat R_t=R_t/(1-\beta_2^t)$. Its detached per-example output signal is
\begin{equation}
    \tilde r_{t,i}=\frac{(1-\lambda)r_{t,i}+\lambda\hat q_t}{\sqrt{\hat R_t}+\epsilon},
\end{equation}
whose batch-mean logit-linear surrogate produces one vector--Jacobian product through the current graph. Thus the implementation averages squared per-example residuals rather than squaring their batch mean, and it keeps no parameter-shaped optimizer state. This two-dataset, two-seed diagnostic produces degenerate endpoints under the tested configuration (Table~\ref{tab:out-rms-ablation}); it supports retaining parameter-local scale information in the reported construction but does not rule out other compressed second moments.

\begin{table}[!htbp]
\centering
\setlength{\tabcolsep}{3.5pt}
\caption{Strict output-only second-moment diagnostic on RoBERTa-base at the fixed epoch-5 endpoint. Means use two paired seeds under the primary fine-tuning schedule. BOM-FullV retains parameter-space $v_{\ell,t-1}$; BOM-OutRMS replaces it with a task-space second-moment EMA and backpropagates the resulting RMS-normalized task signal. The output-only variant removes all parameter-shaped optimizer state while retaining task-space EMA state, but stays near a one-class/chance solution.}
\label{tab:out-rms-ablation}
\begin{tabular}{lcccccc}
\toprule
Task & FullV Train & OutRMS Train & FullV Val & OutRMS Val & $\Delta$Val & \shortstack{OutRMS parameter-\\shaped state (MiB)} \\
\midrule
RTE & 95.46 & 49.44 & 75.45 Acc & 49.28 Acc & -26.17 pt & 0 \\
MRPC & 98.28 & 50.00 & 92.15 F1 & 40.61 F1 & -51.54 pt & 0 \\
\bottomrule
\end{tabular}%

\end{table}

\FloatBarrier

\subsubsection{Anchored-Variant Ablation Results}
\label{app:apm-ablation-results}
The anchored per-tensor momentum variant (\apm{}) tests whether a minimal local reference improves the transported output-momentum direction. For tensor $\theta_\ell$, it fixes one anchor coordinate $i_\ell$ and maintains
\begin{equation}
    a_{\ell,t}=\beta_1a_{\ell,t-1}+(1-\beta_1)u_{\ell,t}[i_\ell].
\end{equation}
With $a_{\ell,0}=0$, it uses the bias-corrected anchor $\hat a_{\ell,t}=a_{\ell,t}/(1-\beta_1^t)$. The reported $K=1$ first-coordinate variant forms the exact clipped least-squares scalar and applies it to the \bpm{} numerator,
\begin{equation}
    r_{\ell,t}=\mathrm{clip}\left(
    \frac{\hat a_{\ell,t}u_{\ell,t}[i_\ell]}{\max\{u_{\ell,t}[i_\ell]^2,10^{-12}\}},
    0.1,3.0\right),
    \qquad
    u^{\mathrm{APM}}_{\ell,t}=r_{\ell,t}u_{\ell,t}.
\end{equation}
The ablation retains \bpm{}'s current reprojection and adds $O(L)$ scalar state for $L$ trainable tensors; it does not store an own-step parameter-space direction. Table~\ref{tab:apm_ablation_appendix} expands its results by dataset; the \bpm{}--APM aggregate appears in Table~\ref{tab:mechanism}.

\begin{table}[!htbp]
\centering
\setlength{\tabcolsep}{4.5pt}
\caption{RoBERTa-base anchored \apm{} variant by dataset. The corresponding \bpm{} aggregate is reported in Table~\ref{tab:mechanism}. Entries are epoch-5 endpoints over the five evaluation seeds. Score denotes Matthews correlation for CoLA, F1 for MRPC and QQP, and accuracy for RTE and SST-2; loss is validation cross-entropy.}
\label{tab:apm_ablation_appendix}
\begin{tabular}{llcc}
\toprule
Dataset & Metric & Val score & Val loss \\
\midrule
CoLA & MCC & 0.5508 & 0.5313 \\
MRPC & F1 & 0.9140 & 0.3737 \\
QQP & F1 & 0.8547 & 0.2607 \\
RTE & Acc & 0.7105 & 0.6167 \\
SST-2 & Acc & 0.9369 & 0.1982 \\
\midrule
Macro mean & -- & 0.7934 & 0.3961 \\
\bottomrule
\end{tabular}

\end{table}

\FloatBarrier

\subsubsection{FP32-State Replication}
\label{app:fp32-replication}
This replication tests whether the primary comparison persists with FP32 parameter and optimizer-state storage and bf16 compute. Both arms are independently tuned under the shared policy; selected rates and settings are in Appendix~\ref{app:diagnostic-protocols}.

Table~\ref{tab:fp32-replication} reports the resulting endpoints. \bpm{} has the higher task-appropriate mean score on four of five tasks, increasing the all-seed macro from $0.8055$ to $0.8222$. QQP is the exception, with means of $0.8809$ for AdamW and $0.8777$ for \bpm{}, whereas the other four score margins range from $+0.0094$ to $+0.0361$. Validation cross-entropy decreases on four tasks and on the macro, from $0.4558$ to $0.3956$; AdamW is lower on QQP, $0.2505$ against $0.2587$. This replication therefore shows that the positive macro comparison is not specific to bf16 parameter and optimizer-state storage.

\begin{table}[!htbp]
\centering
\setlength{\tabcolsep}{3.4pt}
\caption{FP32-state replication of the RoBERTa-base AdamW--\bpm{} comparison. Both arms follow the shared fine-tuning protocol and are evaluated once after epoch 5 on the official validation split used as the final test (Appendix~\ref{app:reproducibility}) over the five evaluation seeds. Scores use the task metric shown; losses are validation cross-entropy. Entries are mean $\pm$ sample standard deviation over five seeds, and $\Delta$ is the paired \bpm{}$-$AdamW mean. The macro score margin has paired 95\% CI $[+0.0073,+0.0261]$ ($p=0.0078$). Task-level paired intervals in Table~\ref{tab:fp32-paired-ci} use the same endpoint. All 50 endpoints are nondegenerate. \bpm{} has the higher mean score on four of five tasks and a five-task macro of $0.8222$ versus AdamW's $0.8055$; its validation cross-entropy is lower on four tasks and on the macro, while AdamW is lower on QQP.}
\label{tab:fp32-replication}
\begin{tabular}{llrrr}
\toprule
\multicolumn{5}{l}{\textit{Validation score}} \\
Dataset & Metric & AdamW & \bpm{} & $\Delta$ \\
\midrule
CoLA & MCC & $0.5789\pm0.0337$ & $\mathbf{0.6150}\pm0.0176$ & $+0.0361$ \\
MRPC & F1 & $0.9048\pm0.0058$ & $\mathbf{0.9186}\pm0.0075$ & $+0.0138$ \\
QQP & F1 & $\mathbf{0.8809}\pm0.0074$ & $0.8777\pm0.0070$ & $-0.0032$ \\
RTE & Acc & $0.7350\pm0.0161$ & $\mathbf{0.7625}\pm0.0065$ & $+0.0274$ \\
SST-2 & Acc & $0.9278\pm0.0047$ & $\mathbf{0.9372}\pm0.0073$ & $+0.0094$ \\
\midrule
\textbf{Macro mean} & -- & $0.8055\pm0.0047$ & $\mathbf{0.8222}\pm0.0053$ & $+0.0167$ \\
\bottomrule
\end{tabular}
\par\medskip
\begin{tabular}{llrrr}
\toprule
\multicolumn{5}{l}{\textit{Validation loss}} \\
Dataset & Metric & AdamW & \bpm{} & $\Delta$ \\
\midrule
CoLA & MCC & $0.6300\pm0.0812$ & $\mathbf{0.4857}\pm0.0425$ & $-0.1443$ \\
MRPC & F1 & $0.4287\pm0.0589$ & $\mathbf{0.3898}\pm0.0309$ & $-0.0389$ \\
QQP & F1 & $\mathbf{0.2505}\pm0.0061$ & $0.2587\pm0.0176$ & $+0.0081$ \\
RTE & Acc & $0.7355\pm0.0779$ & $\mathbf{0.6239}\pm0.0441$ & $-0.1116$ \\
SST-2 & Acc & $0.2341\pm0.0232$ & $\mathbf{0.2198}\pm0.0317$ & $-0.0143$ \\
\midrule
\textbf{Macro mean} & -- & $0.4558\pm0.0345$ & $\mathbf{0.3956}\pm0.0185$ & $-0.0602$ \\
\bottomrule
\end{tabular}%

\end{table}

The five-seed macro margin has a descriptive paired 95\% interval of $[+0.0073,+0.0261]$ ($t=4.935$, $p=0.0078$). Table~\ref{tab:fp32-paired-ci} reports the task-level and macro paired intervals at this same endpoint.

\begin{table}[!htbp]
\centering
\setlength{\tabcolsep}{3.5pt}
\caption{FP32-state replication: paired validation-score margins from the one-shot final test after epoch 5, matching Table~\ref{tab:fp32-replication} and the protocol in Appendix~\ref{app:reproducibility}. Intervals are descriptive, unadjusted 95\% paired-$t$ intervals over five matched evaluation seeds ($df=4$). Macro uncertainty is computed after averaging the five tasks within each seed. The macro margin excludes zero; QQP's interval includes zero.}
\label{tab:fp32-paired-ci}
\begin{tabular}{lccc}
\toprule
Dataset & $\Delta$ score & Paired 95\% CI & Two-sided $p$ \\
\midrule
CoLA & $+0.0361$ & $[-0.0176,+0.0898]$ & 0.1354 \\
MRPC & $+0.0138$ & $[+0.0011,+0.0264]$ & 0.0392 \\
QQP & $-0.0032$ & $[-0.0197,+0.0134]$ & 0.6250 \\
RTE & $+0.0274$ & $[+0.0010,+0.0539]$ & 0.0450 \\
SST-2 & $+0.0094$ & $[+0.00032,+0.01849]$ & 0.0453 \\
\midrule
Five-task macro & $+0.0167$ & $[+0.0073,+0.0261]$ & 0.0078 \\
\bottomrule
\end{tabular}
\end{table}

The cross-entropy ordering differs between the bf16 and FP32-state comparisons. The two comparisons are tuned independently; on QQP, for example, the selected \bpm{} rate is $3\times10^{-5}$ in bf16 and $2\times10^{-5}$ with FP32 state. Cross-entropy measures predicted probabilities as well as classification decisions, so a score gain need not imply a loss reduction. The replication confirms the positive macro result with FP32 parameters and optimizer state.

\FloatBarrier

\subsection{Multiclass Own-Step Projection Control}
\label{app:cv-ownstep-results}

Table~\ref{tab:cv-ownstep} adds the independently tuned multiclass own-step control under the same endpoint convention as Appendix~\ref{app:cv-results}.

\begin{table}[!htbp]
\centering
\setlength{\tabcolsep}{4.0pt}
\caption{Independently tuned multiclass own-step projection control on STL10. Entries are accuracy at the 40-epoch endpoint; A--E identify the five matched evaluation runs. Current-reprojection rows reuse Table~\ref{tab:cv-stl10-check}. Both arms use the same selection budget; training and HPO settings are in Appendix~\ref{app:cv-ownstep-protocol}.}
\label{tab:cv-ownstep}
\begin{tabular}{llrrrrrr}
\toprule
Backbone & Projection & A & B & C & D & E & Mean \\
\midrule
ConvNeXt-Tiny & current  & 0.9784 & 0.9688 & 0.9789 & 0.9771 & 0.9814 & 0.9769 \\
ConvNeXt-Tiny & own-step & 0.9541 & 0.9773 & 0.9774 & 0.9808 & 0.9555 & 0.9690 \\
ViT-Tiny      & current  & 0.9675 & 0.9703 & 0.9626 & 0.9634 & 0.9709 & 0.9669 \\
ViT-Tiny      & own-step & 0.9559 & 0.9655 & 0.9628 & 0.9611 & 0.9578 & 0.9606 \\
\bottomrule
\end{tabular}
\par\vspace{4pt}
\begin{minipage}{\linewidth}
Under the main CV reporting convention, the current-minus-own-step mean accuracy differences are $+0.79$ points for ConvNeXt-Tiny (descriptive paired 95\% CI $[-1.21,+2.79]$ points) and $+0.63$ points for ViT-Tiny (95\% CI $[-0.09,+1.35]$ points). The intervals include zero, and both backbones reproduce the direction of the binary-task control.
\end{minipage}
\end{table}

\FloatBarrier

\subsection{Systems Resource and Timing Results}
\label{app:system-results}

This section groups fine-tuning memory and step-time measurements, C4 pretraining end-to-end resource measurements, and the H800 single-step benchmark. Each study retains its stated precision, hardware and timing scope; Appendix~\ref{app:measurement-protocol} specifies the measurement protocols.

\subsubsection{Fine-Tuning Memory and Step Time}
\label{app:finetuning-resources}

Memory and training-step time are measured in paired runs over the five tasks and five evaluation seeds. On RoBERTa-base, they cover the three base optimizers and their \bpm{} compositions in bf16, and the AdamW pair is also measured with FP32 master weights and optimizer state; on DeBERTa-v3-base and Qwen3-1.7B, the AdamW pair is timed in bf16 (Appendix~\ref{app:measurement-protocol}). Table~\ref{tab:memory-unified} reports optimizer state, peak allocated memory, and step time per method, and Figure~\ref{fig:roberta-pareto} shows the paired quality, optimizer-state and step-time changes after \bpm{} is composed with each base. In bf16, peak allocated memory falls by $232$--$249$~MiB in the two compositions whose base keeps a full dense first moment, close to the size of the removed bf16 buffer, and by $78$~MiB on GaLore, which removes both the first moment in projected coordinates and the dense first moment of the unprojected branch. With FP32 master weights, the AdamW composition removes a $475.5$~MiB buffer and lowers peak allocated memory by $329$~MiB.

\begin{table}[!htbp]
\centering
\setlength{\tabcolsep}{3pt}
\caption{RoBERTa-base memory and step time for three base optimizers and their \bpm{} compositions over five tasks and five evaluation seeds. FP32 rows keep FP32 master weights and optimizer state with bf16 compute; the other rows use bf16 storage. State counts parameter-shaped optimizer buffers in MiB, excluding the task-space vector; peak denotes allocated GPU memory. Ratios and differences are paired within task and seed. Measurement settings and aggregation are specified in Appendix~\ref{app:measurement-protocol}.}
\label{tab:memory-unified}
\begin{tabular}{lcccccc}
\toprule
& Optimizer state & \multicolumn{3}{c}{Peak allocated (MiB)} & \multicolumn{2}{c}{Step time} \\
\cmidrule(lr){3-5}\cmidrule(lr){6-7}
Method & (MiB) & Mean & Paired ratio & Paired $\Delta$ & Mean (ms) & Paired ratio \\
\midrule
AdamW (FP32)        & 951.0 & 3194.0 & --    & --       & 49.35 & -- \\
\bpm{}-AdamW (FP32) & 475.5 & 2865.1 & 0.905 & $-328.9$ & 47.15 & 0.957 \\
\midrule
AdamW               & 475.5 & 2276.9 & --    & --       & 39.84 & -- \\
\bpm{}-AdamW        & 237.7 & 2045.2 & 0.894 & $-231.7$ & 39.50 & 0.993 \\
\midrule
Adam-mini        & 238.2 & 1724.5 & --    & --       & 35.08 & -- \\
\bpm{}-Adam-mini & 0.42  & 1475.3 & 0.845 & $-249.2$ & 34.35 & 0.980 \\
\midrule
GaLore        & 155.7 & 1631.7 & --    & --       & 45.58 & -- \\
\bpm{}-GaLore & 78.3  & 1553.1 & 0.948 & $-78.5$  & 44.07 & 0.969 \\
\bottomrule
\end{tabular}%

\par\vspace{4pt}
\begin{minipage}{\linewidth}
Adam-mini and GaLore use the authors' released implementations, with adaptation details in Appendix~\ref{app:optimizer-implementation}; absolute timings and peak memory are compared within each pair. All three compositions reduce mean state, peak memory and paired step time; state accounting is common across rows.
\end{minipage}
\end{table}

GaLore's official projection scope leaves embeddings, LayerNorms, biases and the classification head in its dense branch. Accordingly, GaLore's state is $155.7$~MiB and the composition removes $77.4$~MiB. Adam-mini's block partition is preserved by the RoBERTa name adaptation; its composition removes the dense first moment. The paired step-time ratios are $0.957$ for AdamW with FP32 master weights, $0.993$ for AdamW in bf16, $0.980$ for Adam-mini, and $0.969$ for GaLore (Table~\ref{tab:memory-unified}).

For the AdamW pair, the five-task mean paired saving is $2.19$~ms per step with FP32 master weights and $0.34$~ms in bf16 (Table~\ref{tab:memory-unified}); \bpm{} is faster in 21 of the 25 FP32 task--seed pairs. The larger FP32 saving is consistent with the removed first-moment buffer occupying twice as many bytes. On DeBERTa-v3-base and Qwen3-1.7B, the bf16 AdamW pairs average $73.5$ versus $69.9$~ms and $383.0$ versus $364.0$~ms per step, paired ratios of $0.952$ and $0.936$. The Adam-mini and GaLore pairs show reductions of $2.0\%$ and $3.1\%$. Adam-mini removes a dense first moment; GaLore removes its projected first moment together with the dense first moment of its unprojected branch, $77.4$~MiB in total. Their state structures, precision and update implementations require separate calibration before Eq.~(\ref{eq:cost-model}) can predict their step times quantitatively.

\FloatBarrier

\subsubsection{C4 Resource Measurements}

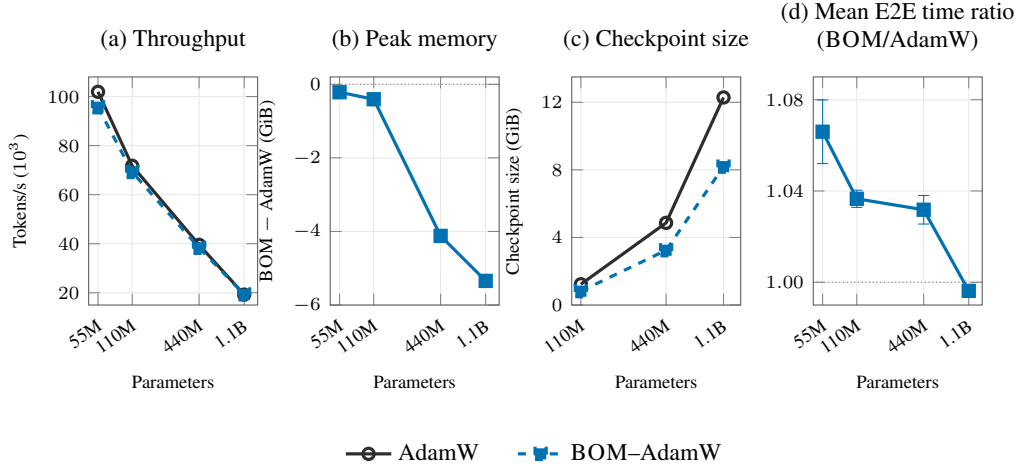
\begin{figure}[!htbp]
\centering
\begin{tikzpicture}
\begin{groupplot}[
  group style={group size=4 by 1,horizontal sep=0.96cm},
  scale only axis,
  width=0.16\textwidth,
  height=0.216\textwidth,
  xmode=log,
  log basis x={10},
  xmin=45,xmax=1450,
  xtick={54.540288,110.121216,435.21664,1099.275264},
  xticklabels={55M,110M,440M,1.1B},
  xticklabel style={rotate=35,anchor=north east},
  xlabel={Parameters},
  tick label style={font=\scriptsize},
  label style={font=\scriptsize},
  title style={font=\small},
  grid=major,
  grid style={gray!18},
  axis line style={black!65},
  scaled ticks=false,
]
\nextgroupplot[
  title={(a) Throughput},
  ylabel={Tokens/s ($10^3$)},
  ymin=15,ymax=108,
  ytick={20,40,60,80,100},
  legend to name=c4-resource-legend,
  legend columns=2,
  legend style={font=\footnotesize,draw=none,fill=none,/tikz/every even column/.append style={column sep=12pt}},
]
\addplot[adamgray,very thick,mark=o,mark size=2.1pt] table[x=parameters_m,y expr=\thisrow{adamw_e2e_tokens_s}/1000,col sep=comma]{data_c4_scale_resources.csv};
\addplot[bpmblue,very thick,dashed,mark=square*,mark size=2.1pt] table[x=parameters_m,y expr=\thisrow{bom_e2e_tokens_s}/1000,col sep=comma]{data_c4_scale_resources.csv};
\legend{AdamW,\bpm{}--AdamW}

\nextgroupplot[
  title={(b) Peak memory},
  ylabel={\bpm{} $-$ AdamW (GiB)},
  ymin=-6.0,ymax=0.2,
  ytick={-6,-4,-2,0},
]
\addplot[black!55,densely dotted] coordinates {(45,0) (1450,0)};
\addplot[bpmblue,very thick,mark=square*,mark size=2.1pt] table[x=parameters_m,y=memory_delta_gib,col sep=comma]{data_c4_scale_resources.csv};

\nextgroupplot[
  title={(c) Checkpoint size},
  ylabel={Checkpoint size (GiB)},
  xmin=95,xmax=1450,
  xtick={110.121216,435.21664,1099.275264},
  xticklabels={110M,440M,1.1B},
  ymin=0.0,ymax=13.5,
  ytick={0,4,8,12},
]
\addplot[adamgray,very thick,mark=o,mark size=2.1pt] table[x=parameters_m,y=adamw_checkpoint_gib,col sep=comma]{data_c4_checkpoint_sizes.csv};
\addplot[bpmblue,very thick,dashed,mark=square*,mark size=2.1pt] table[x=parameters_m,y=bom_checkpoint_gib,col sep=comma]{data_c4_checkpoint_sizes.csv};

\nextgroupplot[
  title={\shortstack{(d) Mean E2E time ratio\\(\bpm{}/AdamW)}},
  ymin=0.990,ymax=1.09,
  ytick={1.00,1.04,1.08},
  yticklabel style={/pgf/number format/fixed,/pgf/number format/precision=2,/pgf/number format/fixed zerofill},
]
\addplot[black!55,densely dotted] coordinates {(45,1) (1450,1)};
\addplot[bpmblue,very thick,mark=square*,mark size=2.1pt,
  error bars/.cd,y dir=both,y explicit]
  table[x=parameters_m,y=e2e_time_ratio,y error=e2e_ratio_sd,col sep=comma]{data_c4_scale_resources.csv};
\end{groupplot}
\path (group c1r1.south west) -- node[pos=0.5,below=48pt,inner sep=0pt]
  {\pgfplotslegendfromname{c4-resource-legend}} (group c4r1.south east);
\end{tikzpicture}
\caption{C4 resource comparison at four scales. (a) End-to-end throughput, $\sum_i N_i/\sum_i T_i$, includes in-window data I/O, evaluation, checkpoint writes and other process-wall overhead. (b) Peak allocated-memory difference. (c) Checkpoint sizes; 55M used no intermediate checkpoint. (d) Mean paired duration ratio with sample-standard-deviation bars over $n_t=5,5,3,3$ pairs; the dotted line marks parity. The 440M pairs use the common interval, steps 96,436--207,519. At 1.1B, one pair uses steps 388,184--732,421 and two pairs use complete runs; each pair has identical token endpoints. See Table~\ref{tab:c4-scale-resources} and Appendix~\ref{app:measurement-protocol} for timing intervals and aggregation.}
\label{fig:c4-scale-resources}
\end{figure}

Table~\ref{tab:c4-scale-resources} and Figure~\ref{fig:c4-scale-resources} report the C4 resource comparison at 55M, 110M, 440M and 1.1B. Panel~(a) pools processed tokens and process-wall seconds across the selected intervals, retaining in-window data I/O, evaluation, checkpoint writes and other overhead. Panel~(b) reports the \bpm{} peak-memory difference; panel~(c) shows checkpoint sizes. The shorter 55M runs used no intermediate checkpoint. Panel~(d) reports mean paired duration ratios with sample-standard-deviation bars. The 55M/110M timings cover complete runs, while all three 440M pairs use the common interval, steps 96,436--207,519 (4.55B tokens). At 1.1B, one pair uses steps 388,184--732,421 and the other two pairs use their full budgets. AdamW uses identical token endpoints within each pair; memory follows the definition in Appendix~\ref{app:measurement-protocol}. The 1.1B readings are approximately 19.252k versus 19.320k tokens/s, with a mean paired duration ratio of 0.996 and a 5.344~GiB reduction in peak allocation. Timing intervals and aggregation are defined in Appendix~\ref{app:measurement-protocol}.

\begin{table}[!htbp]
\centering
\setlength{\tabcolsep}{2pt}
\caption{C4 resource measurements used in Figure~\ref{fig:c4-scale-resources}. The study uses five consecutive-seed reporting runs at 55M/110M and three at 440M/1.1B. Times are arithmetic means of the selected end-to-end intervals, including data I/O, evaluation, checkpoint writes and other process-wall overhead. Peak denotes allocated GPU memory, measured as defined in Appendix~\ref{app:measurement-protocol}. Timing intervals and aggregation are defined in Appendix~\ref{app:measurement-protocol}.}
\label{tab:c4-scale-resources}
\begin{tabular}{lcccccccc}
\toprule
 & & & \multicolumn{3}{c}{Peak allocated (GiB)} & \multicolumn{2}{c}{E2E time (h)} & Mean paired \\
\cmidrule(lr){4-6}\cmidrule(lr){7-8}
Model / tokens & $B\!\times\!T$ & $n$ & AdamW & \bpm{} & $\Delta$ (\%) & AdamW & \bpm{} & time ratio \\
\midrule
55M / 3.0B & $128\times256$ & 5 & 27.116 & 26.901 & $-0.215$ ($-0.8\%$) & 8.175 & 8.714 & 1.066 \\
110M / 5.0B & $24\times256$  & 5 & 7.557  & 7.151  & $-0.406$ ($-5.4\%$) & 19.355 & 20.063 & 1.037 \\
440M / 8.5B & $160\times256$ & 3 & 87.239 & 83.118 & $-4.121$ ($-4.7\%$) & 31.959 & 32.975 & 1.032 \\
1.1B / 15.0B & $80\times256$ & 3 & 83.503 & 78.159 & $-5.344$ ($-6.4\%$) & 178.193 & 177.561 & 0.996 
\\
\bottomrule
\end{tabular}
\par\vspace{4pt}
\begin{minipage}{\linewidth}
The final column averages the paired ratios $T_{\bpm,i}/T_{\mathrm{AdamW},i}$; Figure~\ref{fig:c4-scale-resources}(d) shows their sample standard deviation. The 55M/110M times cover complete runs. At 440M, all three pairs use steps 96,436--207,519, a common 4,549,959,680-token interval. At 1.1B, one pair uses steps 388,184--732,421 (7,049,973,760 tokens); the other two pairs each cover all 14,999,982,080 tokens. Each AdamW interval matches its \bpm{} interval. The 1.1B mean durations average these unequal horizons and are not full-budget completion times; throughput pools tokens and seconds across intervals. Windowed timings exclude earlier initialization and resume loading.
\end{minipage}
\end{table}

\FloatBarrier

\subsubsection{H800 Single-Step Benchmark}

Table~\ref{tab:system-architectures} gives the H800 benchmark architectures.

\begin{figure}[!htbp]
\centering
\begin{tikzpicture}
\begin{axis}[
    width=0.78\textwidth, height=0.44\textwidth,
    xmode=log,
    xlabel={Parameters $P$},
    ylabel={\bpm{}/AdamW step-time ratio},
    ymin=0.90, ymax=1.12,
    xmin=2e8, xmax=5.5e9,
    xtick={2.6e8,4.94e8,9.84e8,2.0e9,3.95e9},
    xticklabels={250M,500M,1B,2B,4B},
    minor xtick={},
    grid=major, grid style={line width=.1pt, draw=gray!25},
    tick label style={font=\scriptsize},
    label style={font=\small},
]
\addplot[color=adamgray, dashed, line width=0.8pt] coordinates {(2e8,1.0) (5.5e9,1.0)};
\addplot[color=bpmblue, mark=*, mark size=2.2pt, line width=1.1pt] coordinates {
(2.6e8,1.081) (4.94e8,1.038) (9.84e8,1.010) (2.0e9,0.979) (3.95e9,0.938)};
\node[anchor=south west, font=\scriptsize, color=adamgray] at (axis cs:2.2e8,1.001) {parity};
\node[anchor=north east, font=\scriptsize, color=bpmblue] at (axis cs:3.95e9,0.975) {\bpm{} faster};
\end{axis}
\end{tikzpicture}
\caption{Paired \bpm{}/AdamW step-time ratio against parameter count on the 250M--4B ladder (single H800, 32k vocabulary, matched FP32 protocol, median of 200 timed steps). The ratio falls monotonically, crosses parity between the 1B and 2B rungs, and reaches $0.938$ at 4B. Per-device tokens per step halve from the 1B rung onward (Table~\ref{tab:steptime-h800}), so $r$ grows faster than $P$ along the ladder. The observed crossing describes these implementations and workloads.}
\label{fig:steptime}
\end{figure}
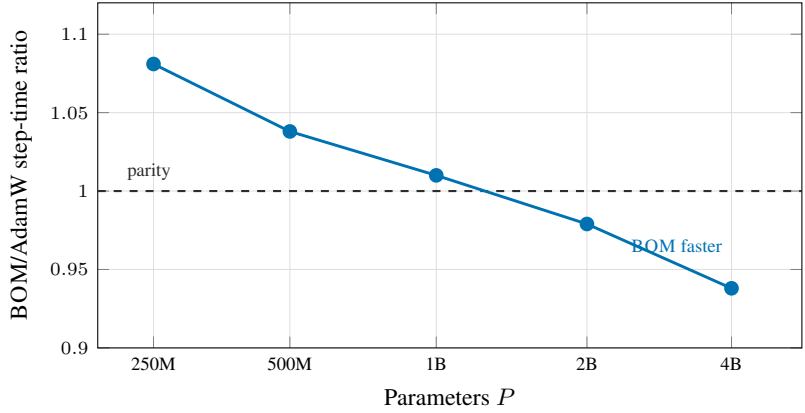

\begin{table}[!htbp]
\centering
\caption{H800 (Hopper, 80~GB) single-GPU ladder. Median step time over 200 timed steps; $\Delta$ is \bpm{} $-$ AdamW. Micro-batch (mb) is selected by the stated out-of-memory halving procedure on the \bpm{} arm; both arms share it.}
\label{tab:steptime-h800}
\setlength{\tabcolsep}{2pt}
\begin{tabular}{lccccccccc}
\toprule
Rung & $P$ & $V$ & mb & \shortstack{Tokens\\/step} & $r$ & \shortstack{AdamW\\(ms)} & \shortstack{\bpm{}\\(ms)} & \shortstack{$\Delta$\\(ms)} & Ratio \\
\midrule
250M & $2.60\times10^{8}$ & 32{,}768 & 128 & 32{,}768 & 0.24 & 504.1 & 544.9 & $+40.9$ & 1.081 \\
500M & $4.94\times10^{8}$ & 32{,}768 & 128 & 32{,}768 & 0.46 & 1041.8 & 1081.5 & $+39.7$ & 1.038 \\
1B & $9.84\times10^{8}$ & 32{,}768 & 64 & 16{,}384 & 1.83 & 889.0 & 897.9 & $+9.0$ & 1.010 \\
2B & $2.00\times10^{9}$ & 32{,}768 & 32 & 8{,}192 & 7.4 & 864.0 & 845.9 & $-18.1$ & 0.979 \\
4B & $3.95\times10^{9}$ & 32{,}768 & 16 & 4{,}096 & 29.5 & 914.5 & 858.2 & $-56.3$ & 0.938 \\
\bottomrule
\end{tabular}%

\end{table}

\paragraph{Implementation-specific measurements.}
H800 peak allocated-memory savings (AdamW minus \bpm{}) change from $-1.031$ and $-0.155$~GiB at 250M and 500M to $2.665$, $7.432$ and $14.731$~GiB at 1B, 2B and 4B (Figure~\ref{fig:pretraining-resources}). At 4B, AdamW peaks at $60.063$~GiB and \bpm{} at $45.332$~GiB; the saving matches the $14.731$~GiB FP32 first-moment buffer.

The observed parity crossing lies between the 1B and 2B workloads listed in Table~\ref{tab:steptime-h800}. This synthetic-token benchmark measures systems cost under the stated implementations; quality results are reported separately in Section~\ref{sec:cross-domain}.

\clearpage
\section{Reproducibility and Hyperparameters}

\label{app:reproducibility}

\subsection{Shared Selection, Reporting, and Environment}
\label{app:shared-protocol}

Table~\ref{tab:protocol-overview} maps the reported studies to their selection and evaluation policies. This appendix records the selected learning rates, batch size, precision, and training schedules used in the reported runs. We use separate HPO search spaces and screening horizons for fine-tuning and from-scratch pretraining because their stable step scales and affordable budgets differ. Across both domains, the default decays are $\beta_1=0.9$ and $\beta_2=0.999$: AdamW applies $\beta_1$ to its parameter-space first moment, whereas \bpm{} applies the same value to its task-space output EMA and maintains no parameter-space first-moment tensor. Other $\beta_1$ values appear only in explicitly labeled ablations.

\begin{table}[!htbp]
\centering
\footnotesize
\setlength{\tabcolsep}{4pt}
\caption{Protocol overview. LR denotes peak learning rate. HPO uses deterministic 90/10 partitions of the applicable training pool and does not access the designated validation split. After HPO selection, or after the peak rate is specified for the no-HPO 1.1B study, final runs train afresh on the complete training pool. Supervised fine-tuning reports the fixed official-validation endpoint, with logged RoBERTa intermediate evaluations shown descriptively; pretraining uses its designated validation split and reports it as validation. Each candidate uses two HPO seeds unless stated otherwise; reported quality uses five consecutive evaluation seeds disjoint from the HPO seeds, except the three-run C4-440M and C4-1.1B groups. Timing is excluded from learning-rate selection.}
\label{tab:protocol-overview}
\begin{tabularx}{\textwidth}{>{\raggedright\arraybackslash}p{0.23\textwidth}>{\raggedright\arraybackslash}X>{\raggedright\arraybackslash}X}
\toprule
Study & Selection policy & Reporting and qualifications \\
\midrule
NLP primary pairs; RoBERTa optimizer controls & Unified LR envelope $[10^{-6},3\times10^{-4}]$; 15 configurations, two-epoch screening on the internal 10\% holdout & Fresh five-epoch training on full official train; fixed epoch-5 official-validation endpoint; bf16 compute and stated storage. Muon additionally searches five update scales after LR selection. \\
FP32-state replication & Repeat the same train-only LR policy independently for each arm and task & FP32 parameters/state, bf16 compute; same five evaluation seeds and one final evaluation on the official validation split. \\
NLP own-step; full temporal kernel & Unified fine-tuning LR policy & Per-task rates in Table~\ref{tab:repro-params-roberta}; constructions in Appendix~\ref{app:roberta-ablations}. \\
Classifier-only past history & Unified 15-configuration, two-HPO-seed policy; primary score uses the protocol-selected configurations & Rescue searches on MRPC (one), QQP (one) and RTE (three) are all reported, and none replaces the primary configurations (Appendix~\ref{app:diagnostic-protocols}). \\
Objective--momentum controls & Unified fine-tuning LR policy; same 15-configuration, two-HPO-seed budget & Independently selected taskwise rates; five evaluation seeds at the epoch-5 endpoint (Table~\ref{tab:repro-params-mechanism}). \\
APM; coefficient sensitivity & Task-specific diagnostic configurations & Table~\ref{tab:repro-params-roberta}; coefficient sweeps are post-selection diagnostics. \\
Numerator/second-moment diagnostics & Baseline rates plus targeted LR searches & Appendix~\ref{app:diagnostic-protocols} specifies cell~B's CoLA/RTE searches and cell~C's CoLA stability screen; Table~\ref{tab:repro-params-mechanism} lists endpoint rates. \\
STL10 primary pair; multiclass own-step & Unified 15-configuration train-only LR policy for each backbone--method pair; two HPO seeds and two-epoch screening. Own-step independently selects $3\times10^{-5}$ on both backbones & Fresh full-train runs for 40 epochs and one final test over the five consecutive evaluation seeds (Tables~\ref{tab:repro-params-vision} and~\ref{tab:cv-ownstep}). \\
Language pretraining & Matched admissible LR range $[5\times10^{-5},10^{-2}]$ and 10-evaluation budget for both arms; internal-holdout screening at 500M/1.0B/1.5B tokens for the 55M/110M/440M scales. The first two average two HPO seeds; 440M uses one. The 1.1B study performs no HPO and fixes a shared $10^{-3}$ rate a priori & Fresh full-budget runs (3.0B/5.0B/8.5B/15.0B tokens as specified in Table~\ref{tab:repro-params-pretraining}) on the complete prepared training pool. The designated validation pool is first accessed after HPO selection, or after the peak rate is specified for 1.1B, for trajectories and endpoint loss over five consecutive-seed reporting runs except C4 440M and 1.1B ($n=3$ each). \\
ImageNet-1k pretraining & Matched pretraining LR range and 10-evaluation budget; epoch-15 top-1 on a seed-specific 10\% holdout from the official training split using two HPO seeds & Fresh 90-epoch runs on the complete training split; the official validation split is first accessed after selection for epochwise and epoch-90 validation top-1 over five consecutive-seed reporting runs disjoint from HPO. \\
\bottomrule
\end{tabularx}
\end{table}

Unless stated otherwise, every independently tuned supervised fine-tuning method receives a learning-rate search over the same admissible envelope, $[10^{-6},3\times10^{-4}]$, with a fixed budget of 15 configurations per method--dataset pair. Learning rate is the only coordinate varied during this stage. For each task and HPO seed, the official training split is partitioned deterministically into 90\% HPO training data and a 10\% internal validation holdout. Every method and candidate with the same task and seed uses the identical partition, and the official validation split is not accessed during HPO. A trial denotes one hyperparameter configuration rather than one seed-level run: each configuration is trained for two epochs on the same two HPO seeds, and its selection score is the equally weighted mean of their task-appropriate internal-holdout scores at the two-epoch endpoint. Each NLP fine-tuning HPO configuration trains for two epochs but follows the first two epochs of a five-epoch learning-rate trajectory, with linear warmup over the first $10\%$ of that five-epoch horizon and linear decay to zero only at the end of the full five-epoch trajectory. The learning-rate stage therefore comprises 15 configurations and 30 seed-level training runs; Muon's additional update-scale stage is specified below. The classifier-only historical-contribution diagnostic uses its protocol-selected two-HPO-seed configurations in the primary score. Rescue searches on MRPC, QQP and RTE are described in Appendix~\ref{app:diagnostic-protocols} and excluded from that score; all other methods use their protocol-selected HPO results.

For each independently tuned method--dataset pair, the 15-configuration learning-rate budget is allocated to ten shared coarse candidates and five additional local refinement candidates around that pair's coarse incumbent. The coarse ladder is $\{1, 1.5, 2, 3, 5, 7\}\times10^{-5}$ and $\{1, 1.5, 2, 3\}\times10^{-4}$. The coarse incumbent is the candidate with the highest mean two-epoch endpoint score on the internal holdout over the two HPO seeds. The five refinement candidates remain within the common admissible envelope: candidates lie between adjacent coarse rates, or, when the smallest coarse rate is the incumbent, below the ladder down to $10^{-6}$. This downward extension accommodates methods whose stable learning rates fall below the shared coarse ladder, including Adam-mini and its \bpm{} composition. The final rate is selected from all fifteen evaluated configurations, retaining the coarse incumbent unless a refinement candidate achieves a strictly higher selection score. Refinement is included in the fifteen-configuration budget and cannot increase it. Method-specific coefficients remain fixed throughout the learning-rate stage. Optimizer-state memory, timing, and routing diagnostics do not enter selection. After selection, every NLP run starts afresh, trains for exactly five epochs on the complete official training split, and reports the official-validation score at the fixed epoch-5 endpoint. The RoBERTa runs also log intermediate training-probe and validation evaluations, shown descriptively in Figure~\ref{fig:roberta-control-curves}; these curves do not replace the fixed reporting endpoint. The primary STL10 AdamW--\bpm{} pair and the multiclass own-step control follow the same train-only HPO split, 15-configuration budget, and final-test separation, with a 40-epoch final budget. Each pair uses the same data, precision and endpoint convention, with five common consecutive evaluation seeds disjoint from the two HPO seeds. Throughout the paper, A--E denote matched evaluation runs rather than numerical seed identifiers. The paper retains the conventional GLUE term ``validation score'' in result tables to identify the released split, although its protocol role is final test; STL10 tables use the same term for the held-out split of their one-shot final evaluation.

Muon first completes this common learning-rate search with its update scale fixed at $0.2$, then holds its selected rate fixed for an additional update-scale search over $\{0.2,0.5,1.0,1.5,2.0\}$ using the same two-epoch, two-HPO-seed internal-holdout criterion. This second stage is additional to the shared learning-rate budget; the selected scales are given in Table~\ref{tab:repro-params-method}. The historical-projection control receives the common learning-rate search. The objective--momentum controls also receive the same 15-configuration, two-HPO-seed learning-rate search, with rates selected independently for each control and task. The \apm{} diagnostic is reported at its task-specific configurations; the decay sweep holds each task's learning rate and all other coefficients fixed while varying $\beta_1$.

If coarse candidates tie in their mean selection score, the candidate with the smaller variance across the two HPO-seed endpoint scores becomes the coarse incumbent; if that variance also ties, the first evaluated candidate wins. This tie rule is applied before generating local refinement candidates.

\paragraph{Local refinement rule.}
The five-candidate local stage is specified as follows.
For the smallest coarse incumbent, $10^{-5}$, it uses
$\{1.5,2,3,5,7\}\times10^{-6}$. Otherwise, let $a$ and $b$ be the preceding and
following coarse rates; for the largest incumbent, set $b=3\times10^{-4}$.
Construct $a(b/a)^{k/8}$ for $k=1,\ldots,7$, round each to two significant digits,
remove duplicates and coarse rates, and retain the five candidates closest to the
incumbent in absolute learning-rate distance (smaller rates break ties).
The incumbent is fixed during candidate generation. All five candidates are evaluated;
selection retains the coarse incumbent on a score tie, and otherwise breaks ties
between refinement candidates by smaller learning rate.

The supervised fine-tuning and ImageNet endpoint runs use bf16 compute under their stated runners, and the fine-tuning memory and step-time figures (Tables~\ref{tab:primary-summary}, \ref{tab:composition} and~\ref{tab:memory-unified}) use the same bf16 compute; the FP32 rows of Table~\ref{tab:memory-unified} use the FP32-state configuration and learning rates described next. The FP32-state replication of Appendix~\ref{app:fp32-replication} repeats the RoBERTa-base AdamW--\bpm{} learning-rate searches with the same 15-configuration, two-HPO-seed learning-rate budget and selection rule, and the five evaluation seeds, while keeping FP32 master weights and optimizer state with bf16 autocast compute. All language-pretraining endpoint studies use the same FP32-storage/bf16-autocast combination; the separate H800 step-time ladder of Appendix~\ref{app:system-results} uses the FP32 storage protocol stated in Appendix~\ref{app:measurement-protocol}.

\paragraph{Fixed optimizer coefficients.}
Within supervised NLP fine-tuning, every coefficient of the primary AdamW--\bpm{} pair other than learning rate is fixed before HPO. Table~\ref{tab:fixed-coefficients} gives these values; Table~\ref{tab:repro-params-method} gives the method-specific settings of additional controls. Muon's update scale is the sole additional searched coefficient. Both primary methods apply decoupled weight decay, disabled for bias and normalization parameters, and use linear learning-rate warmup over the first $10\%$ of updates followed by linear decay to zero, matching the schedule shape used in language pretraining. Reported learning-rate values denote peak rates. Fine-tuning uses bf16 compute and the stated storage precision; the independently tuned FP32-state replication is detailed in Appendix~\ref{app:diagnostic-protocols}. The buffer-count ratio in Section~\ref{sec:adaptive-scaling} is independent of storage precision.

\begin{table}[!htbp]
\centering
\caption{Fixed optimizer coefficients for the primary supervised AdamW--\bpm{} comparisons. The \bpm{} objective uses a fixed $0.5/0.5$ current/history mixture, and its task-space EMA decay is set to the same value as AdamW's first-moment decay $\beta_1$, so the two \emph{stored} histories share the EMA weights $w_{t,\tau}$. After EMA bias correction becomes negligible, \bpm{}'s batch-mean projected residual has current/past total weights $0.55/0.45$, with the current covariance term weighted $0.5$ (Eq.~\ref{eq:u-decomposition}); AdamW's first moment weights current/past full gradients $0.10/0.90$.}
\label{tab:fixed-coefficients}

\begin{tabular}{lcc}
\toprule
Coefficient & AdamW & \bpm{} \\
\midrule
First-order decay $\beta_1$ (parameter EMA / task-space EMA) & 0.9 & 0.9 \\
Second-moment decay $\beta_2$ & 0.999 & 0.999 \\
Denominator constant $\epsilon$ & $10^{-8}$ & $10^{-8}$ \\
Decoupled weight decay $\omega$ & 0.01 & 0.01 \\
Current/history mixture $(1-\lambda)/\lambda$ & -- & $0.5/0.5$ \\
\bottomrule
\end{tabular}
\end{table}

\paragraph{Metrics and aggregation.}
For NLP, \emph{score} denotes Matthews correlation for CoLA, F1 for MRPC and QQP, and accuracy for RTE and SST-2. For STL10, score denotes accuracy. Reported multi-task scores are macro averages of the applicable dataset scores, computed only after averaging the five evaluation seeds within each dataset. Loss is cross-entropy. The training-side endpoint-equivalent diagnostic uses the fixed epoch-5 AdamW training-probe macro score as its target; the first \bpm{} hit is linearly interpolated between adjacent half-epoch training checkpoints. The official validation split does not enter this trajectory calculation. Optimizer-state memory, batch time, and wall-clock time are reported separately and are never combined with task scores; every timing pair covers the five tasks and five evaluation seeds, and the RoBERTa-base AdamW pair is timed both in bf16 and with FP32 master weights. Each comparison shares task, seed and GPU model (Table~\ref{tab:memory-unified}); timing does not enter HPO.

\subsubsection{Environment and Data}
Table~\ref{tab:environment} records the models, data, software, and hardware behind the reported runs; the protocol described above applies unchanged.
\begin{table}[!htbp]
\centering
\caption{Environment for the reported runs. Model identifiers are Hugging Face
model ids; every backbone is loaded through its standard sequence-classification
interface, so no custom head or pooling is introduced. Precision, batch sizes,
schedules, and the seed protocol are specified in the surrounding text of this
appendix.}
\label{tab:environment}
\setlength{\tabcolsep}{5pt}
\renewcommand{\arraystretch}{1.1}
\begin{tabularx}{\linewidth}{@{}>{\raggedright\arraybackslash}p{0.28\linewidth}>{\raggedright\arraybackslash}X@{}}
\toprule
Item & Value \\
\midrule
NLP backbones & \texttt{roberta-base}; \texttt{microsoft/deberta-v3-base}; \texttt{Qwen/Qwen3-1.7B} \\
Classification head & each backbone's \texttt{AutoModelForSequenceClassification} head, \texttt{num\_labels}$=2$ \\
Tokenization & each backbone's fast \texttt{AutoTokenizer}; the decoder-only backbone pads with its end-of-sequence (EOS) token \\
NLP data & GLUE (\texttt{nyu-mll/glue} via Hugging Face \texttt{datasets}); CoLA, MRPC, QQP, RTE, SST-2; HPO uses a deterministic 10\% holdout from official train, and the official validation split supplies the fixed epoch-5 reporting scores; logged RoBERTa intermediate evaluations are descriptive \\
Software & PyTorch $\geq$2.2, Transformers $\geq$4.53, \texttt{datasets} $\geq$2.18, \texttt{torchvision} $\geq$0.17, \texttt{timm} 1.0.15 (pinned), \texttt{adam-mini} 1.1.1 and \texttt{galore-torch} 1.0 (pinned) \\
Control implementations & Adam-mini and GaLore use the authors' released packages, \texttt{adam-mini} 1.1.1 and \texttt{galore-torch} 1.0, using the parameter grouping and projection scope described in Appendix~\ref{app:optimizer-implementation}; all other controls use the sources cited with them \\
Hardware, NLP fine-tuning endpoints & one NVIDIA RTX 4090 per run for the primary pairs and the RoBERTa-base controls; the FP32-state replication (Appendix~\ref{app:fp32-replication}), the AdamW $\beta_1=0.45$ control (Table~\ref{tab:adamw-momentum-ratio-control}) and the fixed-batch drift probe (Table~\ref{tab:drift-decomposition}) on one NVIDIA RTX 4090 per run; the own-step control and its drift telemetry (Tables~\ref{tab:stale-control} and~\ref{tab:drift}) on one NVIDIA RTX PRO 6000 per run \\
Hardware, vision endpoints & one NVIDIA RTX 6000 Ada per run (STL10 and ImageNet-1k) \\
Hardware, language pretraining & one NVIDIA RTX PRO 6000 per run (Qwen3-55M on Python code and FineWeb-Edu; C4 at 55M, 110M, 440M and 1.1B); C4-1.1B uses the Blackwell Server Edition, PyTorch 2.8.0 with CUDA 12.8 and Transformers 5.8.0 \\
Hardware, step-time measurements & RoBERTa-base, DeBERTa-v3-base and Qwen3-1.7B pairs on NVIDIA RTX 4090 cards, one job per card (Table~\ref{tab:memory-unified}); pretraining ladder on a single H800 (Appendix~\ref{app:system-results}) \\
\bottomrule
\end{tabularx}
\end{table}

\FloatBarrier

\subsection{NLP Fine-Tuning Parameters}
\label{app:hpo-repro}
Unless stated otherwise (the FP32-state replication and the fixed-batch drift probe), all supervised NLP fine-tuning runs, for every backbone and method, use batch size 32 and bf16 parameter and optimizer-state storage and compute, with the task-space EMA kept in FP32; the maximum sequence length is 128 for RoBERTa-base and DeBERTa-v3-base and 256 for Qwen3-1.7B. The AdamW--\bpm{} pairs use the fixed coefficients in Table~\ref{tab:fixed-coefficients}, and the other methods the settings in Table~\ref{tab:repro-params-method}. The FP32-state replication keeps the batch size, sequence length and fixed coefficients while independently tuning learning rates (Appendix~\ref{app:diagnostic-protocols}).
Table~\ref{tab:repro-params-roberta} lists the selected learning rates for the independently tuned RoBERTa-base comparisons and the rates used for the \apm{} diagnostic; Tables~\ref{tab:repro-params-deberta} and~\ref{tab:repro-params-qwen} cover the other two backbones, Table~\ref{tab:repro-params-vision} the cross-domain vision setting, and Table~\ref{tab:repro-params-method} the fixed method-specific coefficients and selected Muon update scales. Every reported fine-tuning rate lies inside the common $[10^{-6},3\times10^{-4}]$ envelope. On RoBERTa-base, DeBERTa-v3-base and Qwen3-1.7B, AdamW and \bpm{} independently select each task's learning rate under the same 15-configuration, two-HPO-seed protocol; the FP32-state replication repeats this protocol. 

\begin{table}[!htbp]
\centering
\setlength{\tabcolsep}{4.5pt}
\caption{Task-specific learning rates for the RoBERTa-base optimizer comparisons
and the \apm{} diagnostic, by dataset. Values are peak learning rates. All cells use max length 128,
10\% linear warmup, and subsequent linear decay to zero.}
\label{tab:repro-params-roberta}

\begin{tabular}{lccccc}
\toprule
Method & CoLA & MRPC & QQP & RTE & SST-2 \\
\midrule
AdamW                  & $3\times10^{-5}$ & $3\times10^{-5}$ & $3\times10^{-5}$ & $2.5\times10^{-5}$ & $3\times10^{-5}$ \\
\bpm{}-AdamW           & $2.5\times10^{-5}$ & $3\times10^{-5}$ & $3\times10^{-5}$ & $2\times10^{-5}$ & $3\times10^{-5}$ \\
\apm{}                 & $3\times10^{-5}$ & $3\times10^{-5}$ & $3\times10^{-5}$ & $2\times10^{-5}$ & $3\times10^{-5}$ \\
Lion                   & $3\times10^{-5}$ & $3\times10^{-5}$ & $2\times10^{-5}$ & $3\times10^{-5}$ & $1\times10^{-5}$ \\
Adam-mini              & $7\times10^{-6}$ & $1\times10^{-5}$ & $3\times10^{-6}$ & $3\times10^{-6}$ & $1\times10^{-5}$ \\
\bpm{}-Adam-mini       & $5\times10^{-6}$ & $5\times10^{-6}$ & $1\times10^{-5}$ & $3\times10^{-6}$ & $5\times10^{-6}$ \\
GaLore                 & $1.5\times10^{-5}$ & $1\times10^{-4}$ & $3\times10^{-5}$ & $1.5\times10^{-5}$ & $1\times10^{-4}$ \\
\bpm{}-GaLore          & $1.5\times10^{-5}$ & $5\times10^{-5}$ & $3\times10^{-5}$ & $1.5\times10^{-5}$ & $3\times10^{-5}$ \\
Muon                   & $3\times10^{-5}$ & $3\times10^{-5}$ & $3\times10^{-5}$ & $1.2\times10^{-4}$ & $3\times10^{-5}$ \\
SCALE                  & $1\times10^{-5}$ & $3\times10^{-5}$ & $3\times10^{-5}$ & $1\times10^{-5}$ & $3\times10^{-5}$ \\
Own-step control       & $2\times10^{-5}$ & $3\times10^{-5}$ & $2\times10^{-5}$ & $1\times10^{-5}$ & $3\times10^{-5}$ \\
\bottomrule
\end{tabular}
\end{table}

\begin{table}[!htbp]
\centering
\setlength{\tabcolsep}{3.5pt}
\caption{DeBERTa-v3-base reproducibility parameters by dataset, for the methods
reported on this backbone (AdamW and \bpm{}). Each arm independently selects its
learning rate using the common 15-configuration, two-HPO-seed protocol; the selected rates coincide.
Values are peak learning rates; all cells use 10\% linear warmup followed by linear decay to zero
and train for five epochs.}
\label{tab:repro-params-deberta}

\begin{tabular}{lcccc}
\toprule
Dataset & AdamW LR & \bpm{} LR & Epochs & Max length \\
\midrule
CoLA   & $3\times10^{-5}$ & $3\times10^{-5}$ & 5 & 128 \\
MRPC   & $3\times10^{-5}$ & $3\times10^{-5}$ & 5 & 128 \\
QQP    & $3\times10^{-5}$ & $3\times10^{-5}$ & 5 & 128 \\
RTE    & $3\times10^{-5}$ & $3\times10^{-5}$ & 5 & 128 \\
SST-2  & $3\times10^{-5}$ & $3\times10^{-5}$ & 5 & 128 \\
\bottomrule
\end{tabular}
\end{table}

\begin{table}[!htbp]
\centering
\setlength{\tabcolsep}{3.5pt}
\caption{Qwen3-1.7B reproducibility parameters by dataset, for the methods
reported on this backbone (AdamW and \bpm{}). Each arm independently selects its
learning rate using the common 15-configuration, two-HPO-seed protocol; the selected rates coincide.
Values are peak learning rates; all cells use 10\% linear warmup followed by linear decay to zero
and train for five epochs.}
\label{tab:repro-params-qwen}

\begin{tabular}{lcccc}
\toprule
Dataset & AdamW LR & \bpm{} LR & Epochs & Max length \\
\midrule
CoLA   & $2\times10^{-5}$ & $2\times10^{-5}$ & 5 & 256 \\
MRPC   & $2\times10^{-5}$ & $2\times10^{-5}$ & 5 & 256 \\
QQP    & $2\times10^{-5}$ & $2\times10^{-5}$ & 5 & 256 \\
RTE    & $1\times10^{-5}$ & $1\times10^{-5}$ & 5 & 256 \\
SST-2  & $2\times10^{-5}$ & $2\times10^{-5}$ & 5 & 256 \\
\bottomrule
\end{tabular}
\end{table}

\FloatBarrier

\subsection{Optimizer-Control and Composition Parameters}
\label{app:optimizer-params}

\begin{table}[!htbp]
\centering
\footnotesize
\caption{Settings specific to individual optimizers, beyond the learning rates
in Table~\ref{tab:repro-params-roberta}. Every value is fixed before formal HPO and is not varied during it, except Muon's update scale, fixed at $0.2$ during learning-rate screening and then given an additional five-point search
after the common learning-rate search, with the selected learning rate held fixed
(Appendix~\ref{app:reproducibility}). GaLore's projected modules and rank follow the authors' GLUE fine-tuning script, and its projection scale and refresh interval are the \texttt{galore-torch} 1.0 defaults; all are fixed for every GaLore and \bpm{}-GaLore
cell. Parameters outside the projected module list---embeddings, LayerNorms,
biases and the classification head---remain unprojected. The GaLore baseline applies the package's AdamW update to these parameters. In \bpm{}-GaLore, both branches omit the local first moment and retain their second-moment rules and decoupled weight decay; the second moments are updated from the corresponding \bpm{} numerator. SCALE applies its AdamW branch
to one-dimensional parameters at the selected matrix rate, so only that rate
is searched. SCALE uses the authors' official implementation. Its backbone matrices carry no first moment; the $0.9$ decay is the
output-layer momentum. The backbone-matrix update normalizes its gradient before applying the learning rate.}
\label{tab:repro-params-method}
\setlength{\tabcolsep}{5pt}
\renewcommand{\arraystretch}{1.05}
\begin{tabularx}{\linewidth}{@{}>{\raggedright\arraybackslash}p{0.34\linewidth}>{\raggedright\arraybackslash}X@{}}
\toprule
Setting & Value \\
\midrule
\multicolumn{2}{@{}l}{\textbf{GaLore, \bpm{}-GaLore}} \\
Implementation                 & \texttt{galore-torch} 1.0 (authors' release) \\
Projected modules              & the authors' GLUE target list: attention and feed-forward \texttt{Linear} weights \\
Projection rank                & 8 \\
Projection scale               & 1.0 \\
Projection refresh interval    & 200 steps \\
Decay coefficients $(\beta_1,\beta_2)$, $\epsilon$ & $(0.9,\,0.999)$, $10^{-8}$ \\
Decoupled weight decay         & 0.01, disabled on bias and LayerNorm parameters \\
\midrule
\multicolumn{2}{@{}l}{\textbf{Lion}} \\
Implementation                 & authors' official release \\
Decay coefficients $(\beta_1,\beta_2)$ & $(0.9,\,0.999)$ \\
Decoupled weight decay         & 0.01 \\
\midrule
\multicolumn{2}{@{}l}{\textbf{Adam-mini, \bpm{}-Adam-mini}} \\
Implementation                 & \texttt{adam-mini} 1.1.1 (authors' release), name lists re-pointed at RoBERTa \\
Second-moment sharing          & the package's own partition: per head, per output neuron, whole-tensor, or elementwise \\
Decay coefficients $(\beta_1,\beta_2)$, $\epsilon$ & $(0.9,\,0.999)$, $10^{-8}$ \\
Decoupled weight decay         & 0.01, disabled on bias and LayerNorm groups \\
\midrule
\multicolumn{2}{@{}l}{\textbf{\bpm{}-Adam-mini, \bpm{}-GaLore}} \\
Current/history mixture $(1-\lambda,\lambda)$ & $(0.5,\,0.5)$, fixed \\
Task-space EMA decay $\beta_1$ & $0.9$, fixed \\
\midrule
\multicolumn{2}{@{}l}{\textbf{SCALE}} \\
Output-layer first-moment decay & 0.9 \\
One-dimensional parameters     & AdamW branch, $(\beta_1,\beta_2)=(0.9,\,0.999)$, $\epsilon=10^{-8}$, at the selected matrix rate \\
Weight decay                   & 0 \\
\midrule
\multicolumn{2}{@{}l}{\textbf{\apm{}}} \\
Anchor                         & fixed scalar per tensor \\
Ratio clip                     & $[0.1, 3.0]$ \\
\midrule
\multicolumn{2}{@{}l}{\textbf{Muon}} \\
Implementation                 & authors' official release \\
Momentum                       & 0.95 \\
Newton--Schulz steps           & 5 \\
Update scale during LR search  & 0.2 \\
Subsequent scale candidates    & $\{0.2,\,0.5,\,1.0,\,1.5,\,2.0\}$ \\
Selected update scale          & 0.2 (2.0 on RTE) \\
\bottomrule
\end{tabularx}
\end{table}

SCALE's selected rates are frozen before the five evaluation runs; its HPO and reporting outcomes are discussed alongside the control results in Appendix~\ref{app:roberta-control-results}.

\subsubsection{Optimizer Implementations}
\label{app:optimizer-implementation}
\paragraph{GaLore: the authors' implementation.}
Both GaLore rows use the authors' released \texttt{galore-torch} 1.0 package. As with Adam-mini, the only change to the optimizer is a \texttt{use\_momentum}
flag, so the composition can drop the parameter-local first moment and its associated bias correction. The projector, reconstruction, and second-moment update and bias correction retain the package's rules;
the baseline keeps the package's first-moment update and correction. The projection scope is the one the
authors' GLUE driver uses for a BERT-family backbone: the weights of the attention and feed-forward
\texttt{Linear} modules are projected, and the embeddings, LayerNorms, biases and classification
head remain in the unprojected branch. In the GaLore baseline, this branch uses the package's AdamW update. In \bpm{}-GaLore, both the projected and unprojected branches replace the local first-moment numerator with the corresponding \bpm{} numerator and allocate no local first-moment buffer. Each branch retains its second-moment construction and bias correction, now driven by the new numerator, together with decoupled weight decay. The projected module list and rank
follow the authors' GLUE fine-tuning script and the projection scale and refresh interval are the \texttt{galore-torch} 1.0 defaults; all four are recorded in
Table~\ref{tab:repro-params-method}, fixed in advance and never searched.

\paragraph{Adam-mini: the authors' implementation.}
Both Adam-mini rows---the fixed-backbone control and the \bpm{} composition of
Table~\ref{tab:composition}---use the released \texttt{adam-mini} 1.1.1 package. The only change to the optimizer is a \texttt{use\_momentum} flag, so that the
composition can drop the parameter-local first moment exactly as it does on the other base
optimizers, disabling the associated first-moment bias correction while preserving the block partition and second-moment update and correction. Two checks
were run before training. With the flag on, the modified optimizer reproduces the released package
bit for bit---identical parameters after six steps and identical state size---so the control arm is
the authors' optimizer; with it off, no first moment is allocated. One adaptation is required for RoBERTa: the
shipped parameter-name lists are written for LLaMA-style module names, and on RoBERTa the substring
\texttt{output} also matches \texttt{attention.output.*}, which routes 48 encoder tensors to the
per-neuron branch and then indexes past the end of the one-dimensional LayerNorm weights. The lists
are re-pointed at RoBERTa's names, leaving the block scheme itself unchanged; LayerNorms then fall
through to the whole-tensor branch they were intended for. The routing printed at the start of every
run then places 24 tensors on the per-head branch (query and key of the twelve layers), 53 on the
per-neuron branch (embeddings, value, attention output, multilayer perceptron (MLP) and classifier), 25 on the
whole-tensor branch (LayerNorm weights), and 99 on the plain elementwise branch (biases). Weight
decay is disabled on the 124 bias and LayerNorm groups, matching the fine-tuning protocol.

\FloatBarrier

\subsection{Vision Fine-Tuning Parameters}
\label{app:cv-params}

In the primary STL10 comparison, the two arms select the same rate under the shared protocol (Table~\ref{tab:repro-params-vision}); the own-step control independently applies that same budget and selects $3\times10^{-5}$ for each backbone (Appendix~\ref{app:cv-ownstep-protocol}).

\begin{table}[!htbp]
\centering
\setlength{\tabcolsep}{3.5pt}
\caption{STL10 transfer parameters for AdamW and \bpm{}. Both backbones fine-tune from ImageNet-pretrained weights with batch size 32, bf16 compute, $\beta_1=0.9$, $\beta_2=0.999$, $\epsilon=10^{-8}$, weight decay $0.01$, and $\lambda=0.5$ for \bpm{}. Both arms use linear warmup over the first $10\%$ of updates followed by linear decay to zero. Independently selected learning rates coincide. Both arms report the final checkpoint under the shared five-seed convention. Own-step settings are in Appendix~\ref{app:cv-ownstep-protocol}.}
\label{tab:repro-params-vision}

\begin{tabular}{llccc}
\toprule
Backbone & Dataset & LR & Epochs & Image size \\
\midrule
ConvNeXt-Tiny  & STL10 & $3\times10^{-5}$ & 40 & 224 \\
ViT-Tiny       & STL10 & $3\times10^{-5}$ & 40 & 224 \\
\bottomrule
\end{tabular}
\end{table}

\FloatBarrier

\subsection{Pretraining Parameters}
\label{app:pretraining-params}
Tables~\ref{tab:repro-params-pretraining} and~\ref{tab:pretraining-results-appendix} list the configurations and aggregate endpoints of the from-scratch studies in Section~\ref{sec:cross-domain}. All pretraining runs use the shared default decays $\beta_1=0.9$ and $\beta_2=0.999$ for both arms: AdamW applies $\beta_1$ to its parameter-space first moment, and \bpm{} applies it to its task-space output EMA without maintaining a parameter-space first-moment tensor. The \bpm{} current/history mixture remains fixed at $0.5/0.5$ throughout training. The remaining language-modeling settings are decoupled weight decay $0.1$, gradient clipping at $1.0$, a linear learning-rate schedule with $10\%$ warmup decaying to zero, and $\epsilon=10^{-8}$ for both optimizers. The ImageNet runs use weight decay $0.05$ and a linear schedule with five warmup epochs. ImageNet uses bf16 compute under its runner; all language-modeling pairs use FP32 parameters and optimizer states with bf16 autocast compute.

\paragraph{Pretraining HPO and reporting.}
Pretraining uses the same train-only HPO separation as fine-tuning. For each HPO seed, the applicable training pool is partitioned deterministically into 90\% HPO training data and a 10\% internal validation holdout. AdamW and \bpm{} use identical seed-specific partitions, matched search spaces and matched budgets; the designated validation source is not accessed during configuration selection, and no reporting seed enters HPO. After selection, reporting runs start afresh on the complete training pool. The designated validation source is then used for reporting trajectories and endpoints, which are reported as validation measurements.
Pretraining does not reuse the fine-tuning ladder, since its stable step scales are an order of magnitude larger. For each independently tuned ImageNet-1k, Python-code, FineWeb-Edu, or C4 setting, AdamW and \bpm{} select their learning rates separately using 10 screening evaluations per setting--method study over the admissible learning-rate range $[5\times10^{-5},10^{-2}]$, organized as seven coarse evaluations on a fixed coarse ladder, $\{1, 2, 5\}\times10^{-4}$ and $\{1, 2, 4, 6\}\times10^{-3}$, followed by three local evaluations of the incumbent and its two neighboring rates, with all other optimizer, batch, schedule, and model settings fixed during screening. Except for the C4-440M study described below, one screening evaluation denotes one learning-rate setting run on the same two HPO seeds and scored by their mean, so evaluation count, HPO-seed count, screening budget, and selection criterion are matched between AdamW and \bpm{} within each setting. Language-model configurations use internal-holdout next-token loss at a scale-specific screening endpoint: the 55M settings with 3.0B-token final budgets stop at 500M tokens, C4-110M with a 5.0B-token final budget stops at 1.0B tokens, and C4-440M with an 8.5B-token final budget stops at 1.5B tokens. Each screening run retains the full horizon and 10\% warmup of its corresponding final schedule. The output-EMA decay and current/history mixture remain fixed throughout screening and final training. ImageNet-1k configurations follow the fixed 90-epoch learning-rate schedule and stop after 15 epochs for screening; selection maximizes the mean internal-holdout top-1 accuracy at the epoch-15 endpoint over the two HPO seeds. After selection, reporting runs start afresh from random initialization on the complete prepared training pool. The standard settings use five consecutive-seed reporting runs disjoint from the two HPO seeds.

C4-440M retains the same coarse grid, admissible range, 10-evaluation allocation, internal-holdout loss objective, and fixed training settings, but uses its scale-specific 1.5B-token screening endpoint and uses one HPO seed per screening evaluation instead of averaging two. AdamW and \bpm{} are screened separately and both select $4\times10^{-3}$. Its final 8.5B-token comparison then uses three consecutive-seed reporting runs rather than the standard five. Scheduled measurements and the reported endpoint use the designated validation source only after the rates are frozen.

The C4-1.1B study performs no HPO. It fixes $10^{-3}$ for both arms over a 15.0B-token schedule and uses three consecutive-seed reporting runs. The shared rate was fixed before the reporting runs. The comparison otherwise inherits the shared optimizer coefficients, precision, and schedule conventions.

\paragraph{Three-point pretraining refinement.}
For an interior coarse incumbent $g_i$ in the ordered seven-point ladder, the local stage evaluates three rates:
$g_i-0.5(g_i-g_{i-1})$, $g_i$ itself, and $g_i+0.5(g_{i+1}-g_i)$.
At the lower boundary incumbent $10^{-4}$, the local triplet is $\{5\times10^{-5},10^{-4},1.5\times10^{-4}\}$; at the upper boundary incumbent $6\times10^{-3}$, it is $\{5\times10^{-3},6\times10^{-3},10^{-2}\}$.
The incumbent is evaluated again on the same HPO seed or seeds rather than merely reusing its coarse-stage score. The coarse-stage scores determine the local triplet; final selection compares only the three new local-stage scores. Thus the seven coarse and three local evaluations use ten screening slots but cover nine distinct learning rates. The factor $0.5$ in the interior formula means half of the adjacent coarse-grid gap, not an absolute learning-rate increment; the boundary triplets are specified separately above. For example, an incumbent of $4\times10^{-3}$ gives the local triplet $\{3,4,5\}\times10^{-3}$.
For pretraining screens with two HPO seeds, candidates tied on their mean endpoint score are ordered by the variance of the two seed-level endpoint scores, with smaller variance preferred; if the variance also ties, the candidate first evaluated wins.
For the single-HPO-seed C4-440M screen, cross-seed variance is undefined, so an exact endpoint-score tie is resolved in favor of the first evaluated candidate.

\paragraph{Model and token-level update.}
Qwen3-55M denotes our randomly initialized, reduced-size implementation of the Qwen3 architecture, instantiated as \texttt{Qwen3ForCausalLM}. It has $54{,}540{,}288$ trainable parameters: 12 layers, hidden size 512, feed-forward size 1536, eight query heads and four key/value heads of dimension 64. The C4 scale study adds a 110M model with $110{,}121{,}216$ parameters (12 layers, hidden size 768, feed-forward size 2304, 12 query and six key/value heads of dimension 64) and a 440M model with $435{,}216{,}640$ parameters (20 layers, hidden size 1280, feed-forward size 3840, ten query and five key/value heads of dimension 128). The 1.1B model has $1{,}099{,}275{,}264$ parameters: 27 layers, hidden size 1792, feed-forward size 5376, 14 query heads and seven key/value heads of dimension 128. All configurations use tied input/output embeddings over a $32{,}768$-token vocabulary, SiLU, RMSNorm with $\epsilon=10^{-6}$, RoPE base $10^6$, no attention bias or dropout, and initialization standard deviation $0.02$. Python code and FineWeb-Edu use the 55M configuration; all models are trained from random initialization.

For language modeling, each supervised next-token position plays the role of an example in Section~\ref{sec:formal-update}. If $\mathcal I_t$ is the set of valid target positions and $N_t=|\mathcal I_t|$, the vocabulary-space signal is
\begin{equation}
\begin{aligned}
 s_t&=\frac{1}{N_t}\sum_{(i,j)\in\mathcal I_t}\left[\operatorname{softmax}(z_{t,i,j})-\operatorname{onehot}(y_{t,i,j})\right],\\
 \mathcal L_t^{\mathrm{mix}}&=(1-\lambda)\mathcal L_t^{\mathrm{CE}}+\frac{\lambda}{N_t}\sum_{(i,j)\in\mathcal I_t}\langle z_{t,i,j},\hat q_t\rangle.
\end{aligned}
\end{equation}
Ignored targets are excluded from both averages; $\hat q_t$ is detached. One vocabulary-sized EMA is updated per optimizer step, and no language run uses gradient accumulation. All 55M runs (Python code, FineWeb-Edu and C4) and the 110M, 440M and 1.1B C4 runs use context length 256 and respective micro-batches of 128, 24, 160 and 80, for 91{,}552, 813{,}802, 207{,}519 and 732{,}421 optimizer steps. Evaluation uses 64 held-out batches: every 4{,}096 steps at 55M and every 50M processed tokens at the larger scales. Final evaluation loss is the endpoint at every language scale; for 1.1B, this is the evaluation after 14{,}999{,}982{,}080 processed tokens. Perplexity is computed by exponentiating the final loss of each run before aggregation. The 1.1B configuration follows the shared optimizer and precision settings above, with both peak learning rates set to $10^{-3}$, $\beta_1=0.9$, $\beta_2=0.999$ and task-space EMA decay $0.9$. ImageNet instead reports the epoch-90 validation top-1, evaluated every epoch on one RTX~6000 Ada per run.

\paragraph{C4-110M micro-batch selection.}
The common micro-batch was chosen by a throughput sweep over 12 candidate sizes from 16 to 128. For each size, the qualification ran two AdamW learning rates ($3\times10^{-3}$ and $4\times10^{-3}$) and two BOM rates ($10^{-3}$ and $1.5\times10^{-3}$), and selected the size maximizing the minimum tokens/s across these four configurations. Table~\ref{tab:110m-batch-selection} reports three candidates from this sweep. Batch 24 achieved the highest observed minimum throughput; its margin over 32 was only about $0.16\%$. This was a throughput-based choice, not a maximum-memory batch limit. Both reporting arms retained batch 24. With context length 256 and no gradient accumulation, C4-110M therefore processes $6{,}144$ tokens per update, compared with $32{,}768$ for C4-55M (batch 128). This difference changes the optimization setting, including the number of updates per token budget, and may affect the training trajectory and endpoint loss.

\begin{table}[H]
\centering
\small
\caption{Selected C4-110M batch-qualification measurements. Throughput is the minimum over the four pre-run configurations described above, rounded to the nearest token/s. These measurements select a common micro-batch; they do not compare validation quality.}
\label{tab:110m-batch-selection}
\begin{tabular}{rr}
\toprule
Micro-batch & Minimum throughput (tokens/s) \\
\midrule
24 (selected) & 64{,}698 \\
32 & 64{,}598 \\
128 & 59{,}121 \\
\bottomrule
\end{tabular}
\end{table}

\paragraph{Language corpora.}
The Python corpus is prepared from \texttt{codeparrot/codeparrot-clean}, the cleaned Python-code dataset released by CodeParrot\footnote{\url{https://huggingface.co/datasets/codeparrot/codeparrot-clean}}. Its prepared training and held-out evaluation pools contain $2.10$B and $8{,}388{,}608$ tokens, respectively; the $3.0$B-token training budget counts tokens processed from the training pool. Tokenization uses Qwen3-1.7B-Base, with a corpus-frequency mapping retaining the $32{,}767$ most frequent tokens plus an unknown token ($0.981637$ of token mass retained). FineWeb-Edu uses the \texttt{sample-10BT} configuration (shards 000--005); its documents are tokenized in sorted shard order with an appended EOS, the first $3.10$B tokens form the training pool and the next $8{,}388{,}608$ tokens the held-out evaluation pool, so at most one document spans the two pools. It uses the same tokenizer and vocabulary construction, with frequencies computed from training tokens only, retaining $0.98129$ of token mass. C4 uses the English configuration of \texttt{allenai/c4} \citep{raffel2020t5}, pinned to dataset revision \texttt{1588ec454efa1a09f29cd18ddd04fe05fc8653a2}. At 55M its prepared pools contain $2.10$B training tokens and $8{,}388{,}608$ held-out tokens, and the $3.0$B budget again counts tokens processed; the 110M and 440M runs draw from an $8.50$B-token training pool with the same vocabulary construction. It uses the Qwen3-1.7B-Base tokenizer with the same top-$32{,}767$-plus-unknown construction, retaining $0.976657$ of token mass.

For the 1.1B configuration, the C4 training pool contains 15.0B tokens and uses the same frozen top-$32{,}767$ vocabulary map described above. Its batch size is 80, so the 732{,}421 updates process 14{,}999{,}982{,}080 tokens. Evaluation and checkpointing occur at each 50M-token boundary and at the final update.
\begin{table}[!htbp]
\centering
\setlength{\tabcolsep}{2.8pt}
\caption{From-scratch pretraining parameters. Reporting uses five consecutive-seed runs except for the C4 440M and 1.1B studies, which each use three. Budgets refer to final schedules. Each ImageNet HPO evaluation screens its learning rate for 15 epochs over two HPO seeds. Language HPO uses scale-specific screening endpoints: 500M tokens for the 55M/3.0B settings, 1.0B for C4-110M/5.0B, and 1.5B for C4-440M/8.5B. The 55M and 110M screens average two HPO seeds. C4-440M retains the same coarse grid, admissible range, 10-evaluation allocation, loss criterion, and selection rule but uses one HPO seed; both separate screens select $4\times10^{-3}$. C4-1.1B performs no HPO and fixes $10^{-3}$ for both arms a priori. Both arms use $\beta_1=0.9$ and $\beta_2=0.999$: AdamW applies $\beta_1$ to its parameter-space first moment and \bpm{} applies it to its task-space output EMA. The \bpm{} arm has no parameter-space first-moment tensor and uses $\lambda=0.5$.}
\label{tab:repro-params-pretraining}

\begin{tabular}{lp{90pt}cccc}
\toprule
Setting & Model & Budget & \shortstack{Schedule\\(warmup)} & AdamW LR & \bpm{} LR \\
\midrule
ImageNet-1k  & ResNet-50 (random init)  & 90 epochs      & linear (5 epochs) & $1\times10^{-3}$ & $1.5\times10^{-3}$ \\
Python code  & Qwen3-55M, 32k vocab     & 3.0B tokens    & linear (10\%)    & $5\times10^{-3}$ & $1.5\times10^{-3}$ \\
FineWeb-Edu  & Qwen3-55M, 32k vocab     & 3.0B tokens    & linear (10\%)    & $4\times10^{-3}$ & $1\times10^{-3}$ \\
C4           & Qwen3-55M, 32k vocab     & 3.0B tokens    & linear (10\%)    & $4\times10^{-3}$ & $2\times10^{-3}$ \\
C4           & Qwen3-110M, 32k vocab    & 5.0B tokens    & linear (10\%)    & $3\times10^{-3}$ & $1.5\times10^{-3}$ \\
C4           & Qwen3-440M, 32k vocab    & 8.5B tokens    & linear (10\%)    & $4\times10^{-3}$ & $4\times10^{-3}$ \\
C4           & Qwen3-1.1B, 32k vocab    & 15.0B tokens   & linear (10\%)    & $1\times10^{-3}$ & $1\times10^{-3}$ \\
\bottomrule
\end{tabular}
\end{table}

\FloatBarrier

\subsection{RoBERTa Diagnostic Training and Selection Details}
\label{app:diagnostic-protocols}

The objective--momentum controls of Table~\ref{tab:qqp-factorial-control} use independently selected learning rates under the same 15-configuration, two-HPO-seed HPO budget as the primary pair. The anchored-variant diagnostic in Appendix~\ref{app:apm-ablation-results} is reported at its task-specific configurations. The \apm{} learning rates are listed in Table~\ref{tab:repro-params-roberta}, and its anchor and ratio-clip settings in Table~\ref{tab:repro-params-method}. Table~\ref{tab:repro-params-mechanism} gives the learning rates for the two objective--momentum controls, the numerator/second-moment cross-pairings, and the fixed-batch drift probe.

\begin{table}[!htbp]
\centering
\setlength{\tabcolsep}{3.5pt}
\caption{Learning rates used for the RoBERTa-base mechanism diagnostics. The objective--momentum rows report independently selected learning rates under the same 15-configuration, two-HPO-seed HPO budget as AdamW and \bpm{} (Table~\ref{tab:qqp-factorial-control}); the cell~B/C rows correspond to Table~\ref{tab:numerator-preconditioner}; the fixed-batch drift probe corresponds to Table~\ref{tab:drift-decomposition}. Cell~B's first row uses the primary \bpm{} rates; its second row uses the rates selected by targeted CoLA/RTE searches and retains the anchor rates on the other tasks. Cell~C uses the primary AdamW rates, with a separate CoLA stability screen described in Appendix~\ref{app:diagnostic-protocols}. The two cell~B rows reuse the same MRPC, QQP and SST-2 runs, where their rates coincide.}
\label{tab:repro-params-mechanism}
\begin{tabular}{lccccc}
\toprule
Diagnostic & CoLA & MRPC & QQP & RTE & SST-2 \\
\midrule
Mixed + local $m$ & $3\times10^{-5}$ & $3\times10^{-5}$ & $3\times10^{-5}$ & $2\times10^{-5}$ & $3\times10^{-5}$ \\
CE + no local $m$ & $3\times10^{-5}$ & $3\times10^{-5}$ & $3\times10^{-5}$ & $2\times10^{-5}$ & $3\times10^{-5}$ \\
\midrule
Cell B, at \bpm{}'s rate & $2.5\times10^{-5}$ & $3\times10^{-5}$ & $3\times10^{-5}$ & $2\times10^{-5}$ & $3\times10^{-5}$ \\
Cell B, at $3\times10^{-5}$ & $3\times10^{-5}$ & $3\times10^{-5}$ & $3\times10^{-5}$ & $3\times10^{-5}$ & $3\times10^{-5}$ \\
Cell C, at AdamW's rate & $3\times10^{-5}$ & $3\times10^{-5}$ & $3\times10^{-5}$ & $2.5\times10^{-5}$ & $3\times10^{-5}$ \\
\midrule
Fixed-batch drift probe & $2.5\times10^{-5}$ & $3\times10^{-5}$ & $3\times10^{-5}$ & $2\times10^{-5}$ & $3\times10^{-5}$ \\
\bottomrule
\end{tabular}%

\end{table}

The cell~B/C runs and the fixed-batch drift probe use RoBERTa-base, training and evaluation batch size 32, maximum sequence length 128, $\lambda=0.5$, $\beta_1=0.9$, $\beta_2=0.999$, $\epsilon=10^{-8}$, and weight decay $0.01$. Their HPO runs evaluate the internal 10\% holdout every half epoch. Their final reporting runs train for exactly five epochs on the complete official training split and then evaluate the official validation split once as the final test. Cell~B/C use bf16 weights; the fixed-batch probe uses FP32 weights with bf16 autocast. The probe follows a \bpm{} trajectory without parameter-local first-moment storage and runs in evaluation mode every 20 steps on two fixed training-side batches, using $s^{\ast}=(1,-1)/\sqrt{2}$ at lags 20, 100 and 300. It is separate from both the objective--momentum controls and the own-step projection control.

\paragraph{Post-selection coefficient diagnostics.}
\label{app:coefficient-protocol}
The main configuration and HPO policy are frozen before these diagnostics, and none of their outcomes feeds back into selection. The local mixture sweep changes only $\lambda$ to $0.25$ or $0.75$ on MRPC and SST-2, holding $\beta_1=0.9$. The decay sweep changes only $\beta_1$ over $\{0,0.8,0.9,0.99\}$ on all five RoBERTa tasks, holding the $0.5/0.5$ mixture and \bpm{}'s selected task-specific learning rates fixed. Both use the five evaluation seeds and the one-shot epoch-5 final test. The AdamW current/history-ratio control changes only $\beta_1$ from $0.9$ to $0.45$ at AdamW's main-experiment settings. Results and their mechanistic interpretation are in Appendix~\ref{app:coefficient-results}.

\paragraph{Full temporal-kernel control.}
This control is reported under the unified fine-tuning HPO policy of Appendix~\ref{app:reproducibility}: the common learning-rate envelope, configuration budget, and two-seed internal-holdout selection rule, followed by five evaluation seeds. The selected learning rates are $2\times10^{-5}$ (CoLA), $3\times10^{-5}$ (MRPC), $2\times10^{-5}$ (QQP), $1\times10^{-5}$ (RTE), and $3\times10^{-5}$ (SST-2). Runs use the five evaluation seeds, batch size 32, maximum length 128, $\beta_2=0.999$, $\epsilon=10^{-8}$ and weight decay $0.01$. Parameters and second moments use bf16 storage; the added full-gradient history buffer is FP32. Each final run trains for five epochs on the complete official training split and is then evaluated once on the official validation split used as the final test set.

\paragraph{Classifier-only history and rescue searches.}
Runs use RoBERTa-base, five epochs, the five evaluation seeds, batch size 32, maximum length 128, $\lambda=0.5$, $\beta_1=0.9$, $\beta_2=0.999$, $\epsilon=10^{-8}$, weight decay $0.01$, bf16 parameter, gradient and second-moment storage, and an FP32 residual EMA.

Primary selection for this diagnostic follows the unified learning-rate envelope, 15-configuration budget, two-epoch screening endpoint, and two-HPO-seed internal-holdout mean used by the other independently tuned fine-tuning controls; it selects $2\times10^{-5}$ on MRPC, $3\times10^{-5}$ on QQP and $2\times10^{-5}$ on RTE, and these five-seed groups form the primary result. To test whether the control is under-tuned, rescue searches on MRPC, QQP and RTE, triggered by reporting outcomes, use the same candidate envelope, configuration budget, screening endpoint and internal-holdout criterion with a three-HPO-seed mean as the selection statistic. MRPC and QQP each receive one rescue search, selecting $3\times10^{-5}$ and $2\times10^{-5}$; the three RTE rescue searches yield two distinct alternative rates, $1.5\times10^{-5}$ and $1\times10^{-5}$. Accordingly, none enters the primary classifier-only score. Every five-seed group is reported in Table~\ref{tab:classifier-history-results}.

\paragraph{Numerator/second-moment cross-pairings.}
The cross-pairings are mechanism diagnostics with baseline-rate checks and targeted
learning-rate searches. Cell~B is first evaluated at \bpm{}'s task-specific rate.
Additional searches on CoLA and RTE use two-epoch internal-holdout scores averaged over
the two HPO seeds, selecting $3\times10^{-5}$ on both tasks. The CoLA candidates are
$\{1,1.5,2,2.5,3,5,7\}\times10^{-5}$; RTE uses the same set without $2.5\times10^{-5}$.
MRPC, QQP and SST-2 retain their $3\times10^{-5}$ anchor rates, so the second cell~B
column combines the targeted-search CoLA/RTE endpoints with the unchanged runs on those three tasks.
Cell~C is reported at AdamW's task-specific rate, supplemented by a CoLA stability
screen over $\{1,2,3,5\}\times10^{-6}$ and $\{1,2,3,5\}\times10^{-5}$ using the same
two-epoch, two-HPO-seed criterion. These targeted searches are separate from the
15-configuration protocol for the independently tuned optimizer comparisons. Changing the source of
$v_{\ell,t}$ changes the per-coordinate effective step sizes, so a shared nominal
rate does not by itself hold update magnitudes fixed.

\paragraph{Output-only second moment.}
The OutRMS diagnostic uses two paired seeds on the two reported datasets at the fixed epoch-5 endpoint, under the primary fine-tuning schedule. It is a fixed-configuration diagnostic rather than an independently tuned optimizer comparison (Table~\ref{tab:out-rms-ablation}).

\paragraph{FP32-state replication.}
Both arms use FP32 master weights and FP32 optimizer state with bf16 autocast compute, batch size 32, maximum length 128, the same fixed optimizer coefficients, and the same five-epoch budget as the primary study. AdamW and \bpm{} are independently tuned for every task with the same 15-configuration, two-HPO-seed learning-rate budget and selection rule used by the bf16 comparison, then evaluated on the five evaluation seeds. The selected learning-rate pairs (AdamW, \bpm{}) are $(3,1.5)\times10^{-5}$ on CoLA, $(2,3)\times10^{-5}$ on MRPC, $(3,2)\times10^{-5}$ on QQP, $(2,2)\times10^{-5}$ on RTE, and $(3,2)\times10^{-5}$ on SST-2.

\FloatBarrier

\subsection{Multiclass Own-Step Training and Selection Details}
\label{app:cv-ownstep-protocol}

The STL10 own-step control independently uses the shared 15-configuration, two-HPO-seed, two-epoch screening policy and selects $3\times10^{-5}$ on both backbones. It uses pretrained initialization, batch size 32, image size 224, bf16 compute, $\beta_1=0.9$, $\beta_2=0.999$, $\epsilon=10^{-8}$, weight decay $0.01$ and $\lambda=0.5$. Final runs train on the complete training pool for 40 epochs and evaluate once using the five consecutive evaluation seeds disjoint from the two HPO seeds. Current-reprojection references reuse the primary STL10 endpoints.

\FloatBarrier

\subsection{Memory and Timing Protocols}
\label{app:measurement-protocol}
For RoBERTa-base, optimizer state counts the parameter-shaped buffers allocated by each run, excluding the task-space vector. Memory is measured in MiB ($2^{20}$ bytes); peak allocated memory is recorded with \texttt{max\_memory\_allocated} at the end of each run and includes parameters, gradients, activations and workspace. Paired ratios and differences are formed within task and seed and then averaged.

All timing pairs run on RTX 4090 cards with one job per card and cover five tasks and five evaluation seeds. The RoBERTa-base AdamW pairs train for five epochs under torch 2.5.1, in bf16 and with FP32 master weights, and the two runs of a pair execute back to back; the DeBERTa-v3-base and Qwen3-1.7B AdamW pairs are timed in bf16. In every AdamW pair, both arms use the same AdamW implementation, and the \bpm{} arm omits the first moment and its bias correction. The Adam-mini and GaLore pairs use torch 2.8.0; in each pair, the base arm runs the update of the authors' released package and the \bpm{} arm retains the corresponding update structure while omitting the local first moment. Every pair shares task, seed and GPU model. Each row's absolute step time is the mean per-step time over its runs, while composition comparisons use the mean paired ratio; absolute timings are compared within each pair.

For the training-probe comparison in Table~\ref{tab:aggregate-effective-convergence}, we compute each dataset's task-appropriate training-probe score and then average over the five tasks and five evaluation seeds at every half-epoch checkpoint. The epoch-5 AdamW macro score of $0.8832$ defines the reference. Linear interpolation between adjacent checkpoints gives the epoch at which \bpm{} first reaches this reference. This is a training-score comparison; the reported validation comparison uses the fixed epoch-5 endpoint. Separately, the step-time measurements in Table~\ref{tab:memory-unified} time the forward pass, backward pass and optimizer update, excluding evaluation and checkpointing.

The separate H800 timing protocol behind Section~\ref{sec:steptime} uses FP32 parameters, gradients, and optimizer state; bf16 autocast compute; context length 256; weight decay $0.1$; $\beta_2=0.999$; gradient clipping $1.0$; \bpm{} at the $0.5/0.5$ current/history mixture with output-EMA decay $\beta_1=0.9$; gradient accumulation 1. Batches are synthetic tokens pre-generated on device, so no data pipeline contributes. For each rung the \bpm{} arm runs first and determines the feasible micro-batch (halved automatically on out-of-memory), and the AdamW arm then runs at the same micro-batch on the same GPU back to back; each arm takes 50 warmup and 200 CUDA-event-timed steps, and we report the median. The 250M--4B rungs use the benchmark builder's size-specific Qwen3 configurations. Table~\ref{tab:system-architectures} gives their layer configurations and exact parameter counts. Table~\ref{tab:steptime-h800} lists the workloads and measurements, and Figure~\ref{fig:steptime} plots the resulting ratio curve.

\begin{table}[!htbp]
\centering
\setlength{\tabcolsep}{3.5pt}
\caption{Architectures of the single-GPU systems benchmark in Table~\ref{tab:steptime-h800}, taken from its model builder; $P$ is the exact parameter count, including query/key normalization weights. All use \texttt{Qwen3ForCausalLM}, tied input/output embeddings and disabled KV caching; the timed context length is 256. $L$, $d$, and $d_{\mathrm{ff}}$ denote decoder layers, hidden width and intermediate width; $H_q/H_{kv}$ are query/key--value head counts, and $d_h$ is head width. Pos. is the configured maximum position count, and GC denotes activation checkpointing. Remaining fields in constructed configurations use the benchmark's \texttt{Qwen3Config} defaults.}
\label{tab:system-architectures}
\begin{tabular}{lrrrrcrrrc}
\toprule
Rung & $P$ & $L$ & $d$ & $d_{\mathrm{ff}}$ & $H_q/H_{kv}$ & $d_h$ & Vocab. & Pos. & GC \\
\midrule
250M & 260{,}089{,}344 & 18 & 1024 & 3072 & 8/4  & 128 & 32{,}768 & 4096 & No \\
500M & 494{,}207{,}488 & 23 & 1280 & 3840 & 10/5 & 128 & 32{,}768 & 4096 & Yes \\
1B   & 983{,}658{,}240 & 24 & 1792 & 5376 & 14/7 & 128 & 32{,}768 & 4096 & Yes \\
2B   & 1{,}995{,}527{,}680 & 31 & 2304 & 6912 & 18/6 & 128 & 32{,}768 & 4096 & Yes \\
4B   & 3{,}954{,}407{,}168 & 35 & 3072 & 9216 & 24/8 & 128 & 32{,}768 & 4096 & Yes \\
\bottomrule
\end{tabular}%

\end{table}

The C4 resource study uses process-wall durations, including data I/O, evaluation, checkpoint writes and other overhead within the measured interval. The 55M and 110M summaries each use five complete paired reporting runs. For resumed runs at those scales, total duration is computed by dividing the total processed tokens by the logged full-budget process-wall throughput. At 440M, two \bpm{} reporting runs initially held a redundant tensor copy that does not enter the update computation; it was dropped at the checkpoints at steps 95,215 and 92,774, and the third run never held it. The first evaluation points without the copy are steps 96,436 and 93,995, so all three pairs use the common window from step 96,436 to 207,519, with identical AdamW endpoints: 3,950,018,560 to 8,499,978,240 tokens, or 4,549,959,680 measured tokens. Duration is the difference of the two recorded process-wall times, with no timer reset inside the window. This retains in-window I/O, evaluation and checkpoint writes, but excludes pre-window initialization, resume loading and all earlier training; it is neither a full-run 440M duration nor an isolated-step timer. For each arm, throughput is measured tokens divided by mean duration. The time-ratio statistic is $\bar r=n_t^{-1}\sum_{i=1}^{n_t}T_{\bpm,i}/T_{\mathrm{AdamW},i}$, with sample standard deviation across $n_t=5,5,3$ pairs. Peak memory at 440M is also measured without the redundant copy. Because the copy does not affect the update, quality uses the complete 8.5B-token trajectories of the three reporting runs.

\FloatBarrier

For the 1.1B timing comparison, one pair uses steps 388,184--732,421 (7,049,973,760 tokens), the interval after its \bpm{} run dropped the redundant copy; the other two pairs use their full 14,999,982,080-token budgets. Token endpoints are identical within each AdamW--\bpm{} pair. At an interval boundary, let $N$ be cumulative processed tokens and $\rho$ the logged cumulative process-wall throughput. Cumulative elapsed time is $T=N/\rho$, so the partial-interval duration is $N_{\mathrm{end}}/\rho_{\mathrm{end}}-N_{\mathrm{start}}/\rho_{\mathrm{start}}$; full-run durations use the final $N/\rho$. These process-wall measurements include training, evaluation, checkpointing and other elapsed overhead. Throughput across the unequal intervals is $\sum_i N_i/\sum_i T_i$. The mean duration describes the measured intervals, not a full 15B-token run. The time ratio is the arithmetic mean of the three matched duration ratios, with sample-standard-deviation bars. Peak allocation and checkpoint sizes are measured without the redundant copy.

\subsubsection{First-order cost accounting}
\label{sec:cost-model}
\label{app:cost-model}
Here $B$ is the number of examples or sequences, $C=d_{\mathrm{out}}$ the number of classes, $T$ the language-model context length, and $V=d_{\mathrm{out}}$ the vocabulary size.
For a dense first moment on one unsharded device without gradient accumulation, the relocation changes two elementwise workloads: it removes maintenance of $m_{\ell,t}$ over $P$ elements and adds task-space bookkeeping over $N_{\mathrm{out}}$ supervised output coordinates per optimizer step---$B\,C$ for classification and $B\,T\,V$ for autoregressive language modeling. Treating the shared model forward and backward work as common gives the first-order approximation
\begin{equation}
\Delta t_{\mathrm{step}} \;\approx\; \alpha\,N_{\mathrm{out}} - \beta\,P,
\qquad r \;\coloneqq\; \frac{P}{N_{\mathrm{out}}},
\label{eq:cost-model}
\end{equation}
where $\alpha,\beta$ are time costs per element that depend on hardware, precision, and implementation. Within this approximation, shorter steps require $r>r^{\ast}=\alpha/\beta$. Table~\ref{tab:cost-model} records the element counts; converting them into timing predictions requires calibration for the implementation being measured. Section~\ref{sec:steptime} reports paired single-device timings. Permanent optimizer-state savings instead compare the removed $P$-element buffer with the added $d_{\mathrm{out}}$-element EMA; peak memory additionally depends on temporary output-space allocations.

\paragraph{H800 calibration and checkpoint accounting.}
For Figure~\ref{fig:pretraining-resources}(a), we fit Eq.~(\ref{eq:cost-model}) by unweighted least squares without an intercept to the five single-GPU, $V=32{,}768$ H800 rungs from 250M to 4B, using their exact parameter counts and $N_{\mathrm{out}}=BTV$. With time in milliseconds, the fitted relation is
\begin{equation}
\widehat{\Delta t}
=43.2758\,\frac{N_{\mathrm{out}}}{10^9}
-15.4769\,\frac{P}{10^9}.
\label{eq:h800-calibrated-cost}
\end{equation}
The plotted ratio at rung $i$ is $1+\widehat{\Delta t}_i/t_{\mathrm{AdamW},i}$, where the denominator is the measured AdamW step time. This curve is the in-sample cost-model fit to those five workloads. Both arms use activation checkpointing from the 500M rung onward, and tokens per step decrease from 1B onward. Figure~\ref{fig:pretraining-resources}(c) instead uses unfitted tensor counts: FP32 model weights plus two parameter-shaped moment buffers give $S_{\mathrm{AdamW}}\simeq12P$ bytes, whereas removing the first moment gives $S_{\mathrm{BOM}}\simeq8P$ bytes. These leading terms omit small serialization metadata and task-space buffers. The theoretical curves extend to 4B; checkpoint measurements are at 110M, 440M and 1.1B.

\begin{table}[!htbp]
\centering
\caption{The dimensionless element-count ratio $r=P/N_{\mathrm{out}}$ of Eq.~(\ref{eq:cost-model}) across the training configurations of the main suite. These are structural element counts. Timing predictions additionally require the implementation-specific coefficients in Eq.~(\ref{eq:cost-model}).}
\label{tab:cost-model}
\setlength{\tabcolsep}{3pt}
\begin{tabularx}{\textwidth}{@{}Xcccc@{}}
\toprule
Setting & \shortstack{Output\\coordinates} & $P$ & $N_{\mathrm{out}}$ & $r$ \\
\midrule
NLP fine-tuning, RoBERTa-base ($B{=}32$, $C{=}2$) & $B\cdot C$ & $1.25\times10^{8}$ & $64$ & ${\sim}2\times10^{6}$ \\
STL10 fine-tuning, ViT-Tiny / ConvNeXt-Tiny ($B{=}32$, $C{=}10$) & $B\cdot C$ & $5.7\times10^{6}$--$2.8\times10^{7}$ & $3.2\times10^{2}$ & ${\sim}10^{4}$--$10^{5}$ \\
ImageNet-1k pretraining, ResNet-50 ($B{=}256$, $C{=}1000$) & $B\cdot C$ & $2.56\times10^{7}$ & $2.6\times10^{5}$ & ${\sim}10^{2}$ \\
LM pretraining, Qwen3-55M ($B\,T{=}32{,}768$, $V{=}32{,}768$) & $B\cdot T\cdot V$ & $5.45\times10^{7}$ & $1.1\times10^{9}$ & ${\sim}0.05$ \\
\bottomrule
\end{tabularx}%

\end{table}

\FloatBarrier

\subsubsection{Sharding and Gradient Accumulation}
\label{app:distributed-accounting}
The measured step-time crossover uses one unsharded device without gradient accumulation. Extending the accounting requires $P_{\mathrm{state,local}}$, the first-moment elements maintained per device, and $N_{\mathrm{out,local}}$, the output coordinates processed on that device across an optimizer step. At fixed local output workload, state sharding reduces $P_{\mathrm{state,local}}$; at fixed micro-batch size, accumulation increases $N_{\mathrm{out,local}}$. Both can reduce the expected saving from first-moment maintenance, while distributed communication requires separate measurement. These extensions have not been benchmarked here.

\FloatBarrier
\clearpage
\section{Conditional Descent and Stationarity}
\label{app:convergence}
This appendix gives a sufficient condition for descent of the original loss and a finite-horizon stationarity bound. The detached surrogate in Eq.~(\ref{eq:mix-objective}) is a device for constructing an update, not a fixed objective whose minimization implies convergence of the supervised loss. We therefore analyze the actual update as a perturbed, diagonally scaled gradient step. The assumptions below are explicit; they are not established by the empirical comparisons.

\paragraph{Update and assumptions.}
Stack all parameter tensors into $\theta_t$. Let $F$ be the differentiable objective of interest, bounded below by $F_\star$, with $L$-Lipschitz gradient on a region containing the iterates and their connecting segments. Write $G_t=\nabla F(\theta_t)$ and let $\widetilde u_t$ denote the numerator actually supplied to the adaptive update, including clipping when used. Define
\begin{equation}
 D_t=\operatorname{diag}\!\left((\sqrt{\hat v_t}+\epsilon)^{-1}\right),\qquad
 e_t=\widetilde u_t-G_t+D_t^{-1}W\theta_t,
 \label{eq:convergence-error}
\end{equation}
where $W$ is the diagonal matrix of decoupled weight-decay coefficients. The implemented parameter update is exactly
\begin{equation}
 \theta_{t+1}=\theta_t-\eta_t D_t(G_t+e_t).
 \label{eq:convergence-update}
\end{equation}
Assume $0<a\le b<\infty$ with $aI\preceq D_t\preceq bI$ along the trajectory. For the stated second-moment recursion, $\|\widetilde u_t\|_\infty\le U_\infty$ and $\epsilon>0$ suffice: bias correction makes each coordinate of $\hat v_t$ a convex combination of squared past numerators, so one can take $a=(U_\infty+\epsilon)^{-1}$ and $b=\epsilon^{-1}$. These bounds may be conservative. No independence between $D_t$ and the current gradient is assumed.

\paragraph{Proposition 1 (finite-horizon bound).}
If $0\le\eta_t\le a/(4Lb^2)$ and $S_T=\sum_{t=1}^T\eta_t>0$, then
\begin{equation}
 \frac{\sum_{t=1}^T\eta_t\|G_t\|^2}{S_T}
 \le \frac{4\bigl(F(\theta_1)-F_\star\bigr)}{aS_T}
 +\frac{3b^2}{a^2}\,
 \frac{\sum_{t=1}^T\eta_t\|e_t\|^2}{S_T}.
 \label{eq:convergence-bound}
\end{equation}
Thus the weighted mean squared gradient tends to zero if $S_T\to\infty$ and the weighted mean squared perturbation tends to zero. This is a stationarity guarantee, not convergence to a global minimum or convergence of the parameter sequence. A persistent perturbation gives a residual bound instead.

\paragraph{Proof.}
Smoothness and the spectral bounds give
\begin{align}
 F(\theta_{t+1})
 &\le F(\theta_t)-\eta_t\langle G_t,D_t(G_t+e_t)\rangle
       +\frac{L\eta_t^2}{2}\|D_t(G_t+e_t)\|^2 \notag\\
 &\le F(\theta_t)-a\eta_t\|G_t\|^2
       +b\eta_t\|G_t\|\|e_t\|
       +Lb^2\eta_t^2(\|G_t\|^2+\|e_t\|^2) \notag\\
 &\le F(\theta_t)-\frac{a\eta_t}{4}\|G_t\|^2
       +\frac{3b^2\eta_t}{4a}\|e_t\|^2.
 \label{eq:convergence-descent}
\end{align}
The last line uses $b\|G_t\|\|e_t\|\le (a/2)\|G_t\|^2+(b^2/(2a))\|e_t\|^2$, the step-size restriction, and $a\le b$. Summing, using $F(\theta_{T+1})\ge F_\star$, and dividing by $aS_T/4$ proves Eq.~(\ref{eq:convergence-bound}). In particular, a nonstationary step strictly decreases $F$ whenever $\eta_t>0$ and $\|e_t\|<a\|G_t\|/(\sqrt{3}b)$. The argument holds for every realized batch sequence; expectations can also be taken when the terms are integrable. It does not treat an adaptive preconditioner as independent of gradient noise.\hfill$\square$

\paragraph{The BOM perturbation.}
Let $g_t$ be the current batch gradient and $u_t$ the unclipped BOM numerator. The exact batch decomposition in Eq.~(\ref{eq:batch-decomposition}) yields
\begin{equation}
 e_t=(g_t-G_t)
 +\lambda\bar J_t^\top(\hat q_t-s_t)-\lambda c_t
 +(\widetilde u_t-u_t)+D_t^{-1}W\theta_t.
 \label{eq:convergence-bom-error}
\end{equation}
This separates sampling error, temporal tracking error, the covariance contribution removed from the history branch, clipping, and decoupled decay. In particular, a small tracking error alone does not imply that $u_t$ is a descent direction: the covariance and other terms also matter. Decoupled decay is included in the perturbation for the chosen $F$; it is not silently identified with the gradient of an ordinary $\ell_2$-regularized objective under a varying $D_t$.

\paragraph{Lemma 2 (bias-corrected EMA tracking).}
For $q_0=0$ and $0\le\beta_1<1$, define
$\alpha_{t,\tau}=(1-\beta_1)\beta_1^{t-\tau}/(1-\beta_1^t)$.
Then $\sum_{\tau=1}^t\alpha_{t,\tau}=1$ and
\begin{equation}
 \|\hat q_t-s_t\|
 \le \sum_{\tau=1}^t\alpha_{t,\tau}\|s_\tau-s_t\|.
 \label{eq:ema-tracking-general}
\end{equation}
For a fixed full batch, suppose $s(\theta)$ is $L_s$-Lipschitz and
$\|\theta_{j+1}-\theta_j\|\le\eta_j U$ with $\eta_j\le\bar\eta$. Then
\begin{equation}
 \|\hat q_t-s_t\|
 \le L_s U\bar\eta\,\frac{\beta_1}{1-\beta_1}.
 \label{eq:ema-tracking-fixed}
\end{equation}
\emph{Proof.} Expand the bias-corrected EMA and apply the triangle inequality. For the second claim, telescope parameter displacements to bound $\|s_\tau-s_t\|$ by $L_s U\bar\eta(t-\tau)$. The mean age under the normalized truncated geometric weights is at most $\beta_1/(1-\beta_1)$. For $\beta_1=0$ the tracking error is zero.\hfill$\square$

\paragraph{Corollary 3 (a sufficient convergence regime).}
Consider the fixed-full-batch case $g_t=G_t$, without clipping or weight decay, under the assumptions above. Suppose additionally $\|\bar J_t\|_{\mathrm{op}}\le K$ and $\|c_t\|\le C$. Equations~(\ref{eq:convergence-bom-error}) and~(\ref{eq:ema-tracking-fixed}) imply
\begin{equation}
 \frac{\sum_{t=1}^T\eta_t\|G_t\|^2}{S_T}
 \le \frac{4\bigl(F(\theta_1)-F_\star\bigr)}{aS_T}
 +\frac{3b^2\lambda^2}{a^2}
 \left(C+K L_s U\bar\eta\frac{\beta_1}{1-\beta_1}\right)^2.
 \label{eq:bom-fullbatch-rate}
\end{equation}
If $C=0$, a horizon-dependent schedule with $\bar\eta=O(T^{-1/2})$ and $S_T=\Omega(\sqrt T)$ gives an $O(T^{-1/2})$ stationarity rate, provided the stated constants are uniform in $T$. A fixed-fraction linear warmup followed by linear decay, with peak rate proportional to $T^{-1/2}$ and satisfying the step-size restriction, meets these schedule conditions. Vanishing covariance holds, for example, when the examples have identical Jacobians; it is not assumed for the reported neural-network runs. For nonzero $C$, Eq.~(\ref{eq:bom-fullbatch-rate}) retains an explicit covariance-dependent residual term.

\paragraph{Scope of the guarantee.}
The bound applies to the AdamW-style BOM update through its actual diagonal preconditioner. The full-batch corollary demonstrates a sufficient convergence regime; it does not prove that fixed-batch stochastic training with fixed $\lambda=0.5$, clipping, nonzero decay, and the reported peak learning rates satisfies it. With changing batches, Eq.~(\ref{eq:ema-tracking-general}) also contains sample replacement effects, and Eq.~(\ref{eq:ema-tracking-fixed}) cannot be invoked from parameter smoothness alone. Establishing a sharper stochastic guarantee for those settings, or for the Adam-mini and GaLore compositions, remains open. The result neither predicts faster convergence than AdamW nor guarantees a particular validation loss.

\end{document}